\documentclass[pdflatex,sn-mathphys-num]{sn-jnl}

\usepackage{graphicx}%
\usepackage{multirow}%
\usepackage{amsmath,amssymb,amsfonts}%
\usepackage{amsthm}%
\usepackage{mathrsfs}%
\usepackage[title]{appendix}%
\usepackage{xcolor}%
\usepackage{textcomp}%
\usepackage{manyfoot}%
\usepackage{booktabs}%
\usepackage{algorithm}%
\usepackage{algorithmicx}%
\usepackage{algpseudocode}%
\usepackage{listings}%
\usepackage{romannum}
\usepackage{subcaption}
\AtBeginDocument{\pagenumbering{arabic}}

\theoremstyle{thmstyleone}%
\theoremstyle{thmstyletwo}%

\theoremstyle{thmstylethree}%

\begin{document}

\title[Article Title]{Eyettention \Romannum{2}: A Dual-Sequence Architecture for Modeling Fixation Location, Within-Word Landing Position, and Fixation Duration in Reading}


\author[1,$^*$]{\fnm{Shuwen} \sur{Deng}}

\author[2,$^*$]{\fnm{Cui} \sur{Ding}}

\author[1,2]{\fnm{David} \sur{R. Reich}}

\author[1]{\fnm{Paul} \sur{Prasse}}

\author[2,3]{\fnm{Lena} \sur{A. J\"{a}ger}}

\affil[1]{\orgdiv{Department of Computer Science}, \orgname{University of Potsdam}, \orgaddress{\country{Germany}}}

\affil[2]{\orgdiv{Department of Computational Linguistics}, \orgname{University of Zurich}, \country{Switzerland}}

\affil[3]{\orgdiv{Department of Informatics}, \orgname{University of Zurich}, \country{Switzerland}}

\abstract{
The way our eyes move while reading provides valuable insights into both the reader's cognitive processes and the properties of the text. In particular, eye-tracking-while-reading data has shown to be highly beneficial in various technological applications, such as enhancing and interpreting language models and inferring a reader's characteristics. However, these applications often rely on large-scale, data-driven models, which demand extensive eye-tracking datasets that are challenging to obtain due to the resource-intensive nature of data collection. To address the challenge of data scarcity, we develop Eyettention~\Romannum{2}, an end-to-end trained deep-learning model capable of generating realistic scanpaths consisting of a complete set of fixation attributes in chronological order, including fixation location, within-word landing position, and fixation duration. Our model is lightweight, efficiently trainable on limited GPU resources, and closely aligned with cognitive theories. We demonstrate that Eyettention~\Romannum{2} surpasses state-of-the-art models in scanpath prediction and mirrors human-like gaze behavior by capturing key psycholinguistic phenomena. With its robust performance, Eyettention~\Romannum{2} holds the potential to drive advancements in natural language processing, facilitate piloting the materials of psycholinguistic experiments, and uncover new insights beyond what is explicitly encoded in theoretical cognitive models.
}



\renewcommand\thefootnote{$^*$}
\footnotetext{Both authors contributed equally to this research.}




\maketitle

\section{Introduction}
\label{sec_intro}




The way our eyes move while reading reveals important information about both the reader's cognitive processes~\cite{Rayner1998} and the properties of the stimulus text~\cite{rayner2009}. Over the past decades, extensive research in cognitive psychology~\cite{reichle2003ezreader, engbert2002dynamical} and psycholinguistics~\cite{engelmann2013framework} has been dedicated to analyzing eye movements during reading and developing computational cognitive models to simulate human reading behavior. These models can be understood as computational implementations of theories about human reading behavior and its underlying cognitive mechanisms. The primary goal of this research line is to deepen our understanding of the cognitive mechanisms involved in reading and, more generally, human language processing. Most existing approaches remain heuristic, rule-based, or hybrid, as cognitive modeling prioritizes interpretability. Consequently, these models typically feature only a small number of free parameters to adhere to Ockham’s razor and to ensure falsifiability.  While this enhances transparency, it often constrains their ability to capture complex patterns and adapt to variations across different populations, languages, stimulus layouts, lab setups, and reading tasks.

Recent machine learning research has increasingly leveraged eye-tracking-while-reading data for various technological applications. This includes enhancing natural language processing (NLP) models by incorporating eye movement features to enrich text features~\cite{barrett-etal-2016-weakly, mishra-etal-2016-leveraging, hollenstein-zhang-2019-entity} or to regularize neural attention mechanisms to make their inductive bias more human-like~\cite{barrett-etal-2018-sequence, Sood2020ImprovingAttention, sood2023multimodal}. Eye movement data has also been employed to gain insights into the interpretability of neural language models (LMs), providing a means to better understand the differences between human and machine language processing and to evaluate the cognitive plausibility of LMs, which underpin state-of-the-art NLP systems~\cite{beinborn2024cognitive, sood2020interpreting, hollenstein2021multilingual,hollenstein2022patterns, merkx-frank-2021-human}. Additionally, eye movement data is utilized to infer a reader’s characteristics, such as cognitive conditions like ADHD~\cite{deng2022detection} or dyslexia~\cite{Raatikainen2021DetectionData, haller2022eye-tracking}, and linguistic skills, including reading comprehension capacity or whether the reader is a native speaker of the stimulus’ language~\cite{reich2022inferring, Ahn2020TowardsBehavior, Berzak2018Assessing}. For a recent benchmark on predictive modeling from eye movements that includes most of these tasks, see \url{https://eyebench.github.io/}~\cite{shubieyebench}. 

Data scarcity has implications not only for training models, but also for technological applications that rely on real-time gaze input for arbitrary stimuli at deployment.  For example, the above-mentioned approaches for enhancing LMs with gaze-derived features assume access to eye-movement data at application time. In principle, such gaze-enhanced models could leverage human gaze patterns to identify relevant passages, for example, when generating a summary of an input text. However, the assumption that gaze data is available for arbitrary input texts is unrealistic for many real-world applications. For example, when a user wants a language model to summarize a given text, it cannot be assumed that eye-movement data for that specific text has been collected beforehand. To overcome this issue, researchers have started using simulated gaze data for enhancing LMs, resulting in advantages across various NLP tasks~\cite{Sood2020ImprovingAttention,khurana-etal-2023-synthesizing,deng-etal-2023-pre,deng-etal-2024-fine,reich-etal-2024-reading}.

 Finally, one persistent challenge in psycholinguistic experimentation is the limited size of eye-tracking datasets, which has contributed to the replication crisis in the field~\cite{amrhein2017earth, open2015estimating, jager2020interference}. Small samples often result in low statistical power, increasing the likelihood of misleading effect estimates and reducing the reliability of research findings~\cite{vasishthReplicability}. By using simulated data, we can perform statistical power analyses that are closely tailored to the specific experimental materials planned for use with human participants. Such simulations provide more realistic expectations of effect sizes and enable more informed and robust study planning. 
Along the same lines, generative models of eye movements could also be used to pilot the materials for eye-tracking experiments. By simulating readers’ eye-movement patterns in advance, such models could help identify anomalies or potential confounds in the stimuli, prompting revisions to the materials before data collection.

In sum, there is a broad range of potential usages for simulated gaze data:

\begin{itemize}

\item In technological applications, including gaze-augmented NLP, simulated eye-tracking data can be employed to (pre-)train complex machine-learning models that require large amounts of data. For example,  simulated eye-tracking data can support the pre-training of models designed to infer a reader’s characteristics~\cite{prasse2024improving}. In this approach, a model is first pre-trained on simulated eye-tracking data (or on a mixture of simulated and real data) to learn generalizable eye-movement patterns, and then fine-tuned on a smaller set of real human eye-tracking recordings for a specific task. For instance, a model predicting reading comprehension could be pre-trained on synthetic gaze data in a self-supervised way and subsequently fine-tuned using real gaze recordings paired with readers’ comprehension scores obtained through comprehension questions.

While current approaches use synthetic data primarily to learn general eye-movement patterns~\cite{prasse2024improving}, generating scanpaths that are conditioned on specific stimulus characteristics (e.g., text type or layout), on the reader’s goals (e.g., preparing for comprehension questions or summarizing the text), or tailored to specific populations (e.g., L2 readers) or even individual readers could offer additional advantages. Using such targeted synthetic data to augment pre-training or training datasets may further improve model performance. Evidence that increased diversity in training data can be beneficial has already been shown for real eye-tracking data~\cite{reich-etal-2024-reading}, suggesting that similarly diverse simulated data could provide additional gains.

\item As pointed out above, some modeling approaches---such as gaze-enhanced LMs---use gaze data as additional, auxiliary input. For example, when automatically generating a text summary, eye movements on the input text could help a LM identify the most relevant parts. In such scenarios, however, gaze data is typically not available for the specific text at application time, nor is it necessary that the gaze data come from a concrete, real individual. Simulated eye-tracking data can therefore be particularly useful in settings where real-time gaze information is required as input at application time, such as in gaze-augmented NLP.

\item Simulated eye-tracking data can be valuable for preparing psycholinguistics experiments, such as piloting stimulus materials, or estimating expected effect size on a given set of stimuli, which, in turn, can be used for a statistical power analysis.

\item High-performing deep learning models, trained on large-scale data, can capture patterns not explicitly encoded in theoretical frameworks, offering novel insights into factors that influence eye movement behavior, even if such models remain not fully interpretable. For example, they can serve as a tool for investigating which linguistic or visual features are most informative for guiding human reading behavior.

\end{itemize}

Earlier work on eye movement prediction during reading has predominantly focused on either spatial aspects of eye movements, such as fixation location~\cite{deng2023eyettention, bolliger-etal-2023-scandl} and word skipping probability~\cite{hahn2016modeling, wang2019new}, or temporal aspects, like aggregated word-level total fixation duration~\cite{hollenstein-etal-2021-multilingual, hollenstein-etal-2021-cmcl, Sood2020ImprovingAttention, hollenstein-etal-2022-cmcl}. However, there is a growing need for developing models that can predict complete scanpaths--capturing both spatial and temporal dimensions of sequential eye movements during reading--to enable versatile application across various fields.
To this end, we present Eyettention~\Romannum{2}, a lightweight generative deep-learning model designed to generate human-like scanpaths on text. Eyettention~\Romannum{2} extends our previous model, Eyettention~\cite{deng2023eyettention}, an attention-based dual-sequence model that established a new state-of-the-art in fixation location prediction. The new model is capable of generating full scanpath attributes, including fixation location, within-word landing position, and fixation duration,
significantly broadening its applicability 
to a wide range of use cases and research questions. In particular, the contribution of our work is as follows:
\begin{itemize}
    \item We develop an end-to-end trained dual-sequence architecture for scanpath prediction, which can predict comprehensive scanpath attributes, including fixation location, landing position within a word, and fixation duration, achieving state-of-the-art performance. Additionally, we develop a specialized version that can generate scanpaths tailored to specific readers by conditioning the model on reader identity so that it learns and reproduces characteristic reading patterns of individual readers observed during training.
    
    \item We present extensive experiments to evaluate our proposed model across a range of application scenarios, both within and across datasets, and for two typologically different languages with different scripts (alphabetic vs logographic).

    \item We provide qualitative and quantitative inspections of our model’s behavior, including an in-depth analysis of the model's capability to reproduce key psycholinguistic phenomena observed in human reading behavior, namely effects of lexical frequency, word length, and surprisal, as well as an analysis of the impact of different decoding strategies for generating scanpath. 

    \item We perform a power analysis as a demonstration of how the model can be applied to prepare an eye-tracking-while-reading study.
    
    \item We further investigate which key properties of the LM used in Eyettention as the stimulus encoder influence its ability to predict human reading behavior. In particular, we examine how differences in model architecture (i.e., model family) and the number of parameters affect predictive performance.
\end{itemize}

The remainder of this paper is structured as follows. In Section~\ref{sec_relatedwork}, we provide a summary of the related work. In Section~\ref{sec_model}, we introduce the Eyettention~\Romannum{2} model for scanpath prediction. Section~\ref{sec_exp} describes our experiments and evaluation results, followed by a detailed model analysis. Section~\ref{sec_discussion} discusses the implications of the results, and Section~\ref{sec_conclusion} concludes the paper.

\section{Related Work}
\label{sec_relatedwork}
\bmhead*{Cognitive Models of Eye Movement Control in Reading} Cognitive models of eye movement control in reading aim to explain when and where readers move their eyes through the interplay of visual perception, text properties, attention, and oculomotor control. Over the past three decades, two models have been particularly influential in cognitive psychology: the E-Z Reader model~\cite{reichle1998toward, reichle1999eye, reichle2003ezreader} and the SWIFT model~\cite{engbert2005swift}. Both models account for how word-level variables, such as lexical frequency and predictability, affect fixation durations, word skipping, and regressions. They differ, however, in how attention is allocated across words and in how linguistic processing influences the timing and targeting of saccades during reading.

E-Z Reader and SWIFT can be compared by considering several processing stages: (i) visual and perceptual encoding, (ii) familiarity checking, (iii) saccade timing and target selection, (iv) attention allocation, and (v) post-lexical or sentence-level processing. Both models assume that perceptual input quality decreases with eccentricity. For instance, when a fixated word is long, the following word falls farther into the parafovea, reducing preview benefit due to visual acuity drop-off. However, there are more differences than similarities. E-Z Reader adopts a serial, stage-based account of attention~\cite{rayner1976guides, posner1980attention}, in which attention is focused on one word at a time, and lexical processing must be completed before attention shifts to the next word. In contrast, SWIFT assumes a gradient of attention distributed across multiple words~\cite{bouma1974control, morrison1984manipulation}, allowing parallel lexical processing with the highest processing rate at the fovea and decreasing rates for parafoveal words to the left and right of the fixated word, and extending, though to a lesser extent, to multiple coming words. The models also differ in how linguistic properties influence eye movement control. In E-Z Reader, early-stage familiarity check is modulated by word predictability and frequency. Completing the familiarity check directly triggers saccadic programming, thereby tightly linking progress in linguistic processing to fixation durations. In SWIFT, saccade timing is governed by an internal, autonomous mechanism that is influenced, rather than directly triggered, by linguistic processing~\cite{engbert2001mathematical}, and saccade target selection arises from probabilistic competition across a dynamic activation field. At the post-lexical level, E-Z Reader includes a separate integration stage that can delay or halt eye movements under processing difficulty, whereas SWIFT lacks an explicit integration stage; instead, processing difficulty, functioning like a ``brake'' (foveal inhibition), slows eye movements by naturally interfering with the autonomous saccade timer.

More recent work extends these classic accounts by integrating eye-movement control with broader theories of language processing. For instance, \cite{engelmann2013framework} developed a framework that embeds syntactic processing difficulty into the EMMA eye movement control model~\cite{salvucci2001integrated} inside the cognitive architecture ACT-R~\cite{anderson2004integrated} by linking slow post-lexical processing to regressions, enabling predictions of re-reading behavior based on parsing difficulty. The \"{U}ber-Reader model~\cite{veldre2020towards} aims to provide a more comprehensive account of reading by embedding eye movement control within an architecture that explicitly models word identification across sentence reading, lexical decision, and word naming tasks, thereby linking eye movements directly to word identification processes. Similarly, SEAM~\cite{rabe2024seam} combines the SWIFT model with a cue-based memory retrieval framework~\cite{lewis2005activation} to account for how post-lexical processes, such as syntactic dependency completion and similarity-based interference, influence fixation durations and regressive eye movements.

While these models are designed to explain average eye movement phenomena at the group level using cognitively plausible mechanisms, they are typically developed and tested using a single parameter set fit to aggregated data from a small number of participants. 
Importantly, this does not imply that such models are incapable of capturing individual differences. Indeed, previous work has shown that systematic adjustment of model parameters can account for meaningful individual differences in reading skill, task demands, language experience, and so on (e.g., \cite{mancheva04072015, reichle2013using}). However, in practice, large-scale and systematic modeling of individual differences is relatively uncommon and often takes a secondary role to explaining average effects. Additionally, the intentional use of a small number of parameters, while ensuring psychological plausibility, can limit the models' ability to generalize to diverse unseen readers or text types. This reduced generalizability limits their effectiveness in broader technological applications.

\bmhead*{Machine-Learning Models of Eye Movement Control in Reading}
Early machine-learning models for eye movement control rely on hand-crafted features---explicitly coded, conceptually meaningful variables---extracted from eye-tracking data, such as word length, frequency, and surprisal~\cite{nilsson2009learning, nilsson2010towards, nilsson2011entropy, matthies-sogaard-2013-blinkers, hara-etal-2012-predicting}. While these features are easily interpretable, 
a fixed set of hand-crafted features may fail to capture all relevant information and may require manual adjustments to adapt to different reading tasks or datasets. Additionally, the limited number of parameters in these models further constrains their predictive power.
More recent approaches have shifted toward neural networks, which exhibit significantly higher predictive performance. Research has focused on predicting aggregated reading measures for each word of the linguistic stimulus, averaged across readers. These measures include word skipping probability~\cite{hahn2016modeling, wang2019new} and gaze fixation duration~\cite{hollenstein-etal-2021-multilingual, hollenstein-etal-2021-cmcl, Sood2020ImprovingAttention, hollenstein-etal-2022-cmcl}. 
This aggregation process, however, can lead to the loss of potentially relevant sequential information, including important aspects of eye movement behavior such as regressions and re-fixations, as well as individual differences in scanpaths. Moreover, related work in scene viewing shows that deep saliency models can help refine mechanistic scanpath models~\cite{dmoves}, highlighting the importance of modeling full fixation sequences. Recent advancements have addressed this limitation by developing neural networks that predict the complete sequential order of fixations in a scanpath~\cite{deng2023eyettention, bolliger-etal-2023-scandl, khurana-etal-2023-synthesizing}, significantly broadening the applicability of these models across a wider range of use cases. In this context, a pre-trained LM is integrated as a core component of the proposed architectures to process linguistic information essential for accurate scanpath prediction.

\section{Eyettention~\Romannum{2}}
\label{sec_model}
All existing generative models of eye movements---both in cognitive psychology and machine learning—--share a key conceptual and technical limitation: they model eye movements during reading as a standard (i.e., single axis) sequence problem. However, one of the key properties of eye-tracking-while-reading data is its dual-sequence nature: 
The words are ordered following the grammatical rules of the language (\textit{linguistic sequence axis}), whereas the fixations on these words are chronologically ordered (\textit{temporal sequence axis}). As humans do not strictly read from left-to-right, but rather skip or re-fixate words and regress to previous words, the alignment of the linguistic and the temporal sequence axes poses a major architectural challenge. Neither standard sequence models (i.e., with a single input sequence axis), nor encoder-decoder architectures,  that align an input with an output sequence, can handle the \textit{dual-sequence input} structure of eye movements in reading. 
The Eyettention model~\cite{deng2023eyettention} addresses this limitation as the first dual-sequence model that simultaneously processes the sequence of words and the chronological sequence of fixations. The alignment between the two sequences is achieved through a cross-sequence attention mechanism. Eyettention~\Romannum{2} builds upon this framework, extending the original model---which focuses on predicting the fixation location in a scanpath---by also predicting within-word landing positions and fixation durations. These extensions allow Eyettention~\Romannum{2} to capture both the spatial and temporal dynamics of eye movements during reading, offering a more comprehensive picture of human scanpath patterns. 

In brief, Eyettention~\Romannum{2} generates scanpaths for a given text stimulus in a causal, auto-regressive manner, processing each fixation sequentially. This means that each generated fixation, along with all preceding fixations, is used as input for the neural network to predict the next fixation. Consequently, the prediction of each fixation is based on both the text stimulus and the preceding fixations in the scanpath. We formulate the task of scanpath prediction as a ``next fixation prediction'' problem which we detail in
~\textsection~\ref{sec_problemsetting}.
Following this, we introduce the model architecture in~\textsection~\ref{sec_architecture} and~\textsection~\ref{sec_reader_specific}, 
along with training procedures.
outlined in~\textsection~\ref{sec_optimiatzion}.

\subsection{Problem Formulation}
\label{sec_problemsetting}
Given a sentence~$\mathbf{w}=\langle w_1,\dots,w_m\rangle$ and an initial part $\langle \mathbf{f}_1,\dots,\mathbf{f}_{i-1}\rangle$ of the complete scanpath $\langle \mathbf{f}_1,\dots,\mathbf{f}_n\rangle$ for the input sentence, the goal is to predict the next fixation~$\mathbf{f}_i$. Each fixation $\mathbf{f}_i = (k_i, l_i, d_i)$ is a vector of a word index $k_i$ with $1\leq k_i \leq m$, a landing position $l_i$ within the word $w_{k_{i}}$, and a fixation duration $d_i$. For English texts, each $w_{k_{i}}$ represents a word, whereas for Chinese texts, we use $w_{k_{i}}$ to represent characters. The landing position $l_i$ is defined as the normalized horizontal offset in pixel coordinates of a fixation relative to the left boundary of the currently fixed word (for English) or character (for Chinese), yielding values between 0 and 1. This definition offers maximal flexibility to researchers or practitioners using the data generated by Eyettention~\Romannum{2}, as it is language-independent and allows any desired mapping to linguistically or graphically meaningful units, including letters, morphemes, syllables, or even sub-character units such as Chinese radicals or diacritics in alphabetic languages. We treat the problem of word index prediction as a multi-class classification task, with each class representing a possible word index in a sentence. Specifically, we estimate a likelihood function $P(k_i|\mathbf{w},\mathbf{f}_1,\dots,\mathbf{f}_{i-1})$ for predicting the word index. In addition, we treat the problems of predicting landing position and fixation duration as regression tasks, producing point estimates for the predicted landing position $\hat{l}_i$ and duration $\hat{d}_i$. This process is iteratively applied for each fixation in the scanpath until the entire scanpath is predicted.

\subsection{Model Architecture}
\label{sec_architecture}

\begin{figure*}[!ht]
	\centering
	  \includegraphics[width=\textwidth,keepaspectratio,trim={0cm 0 15cm 0}, clip]
   {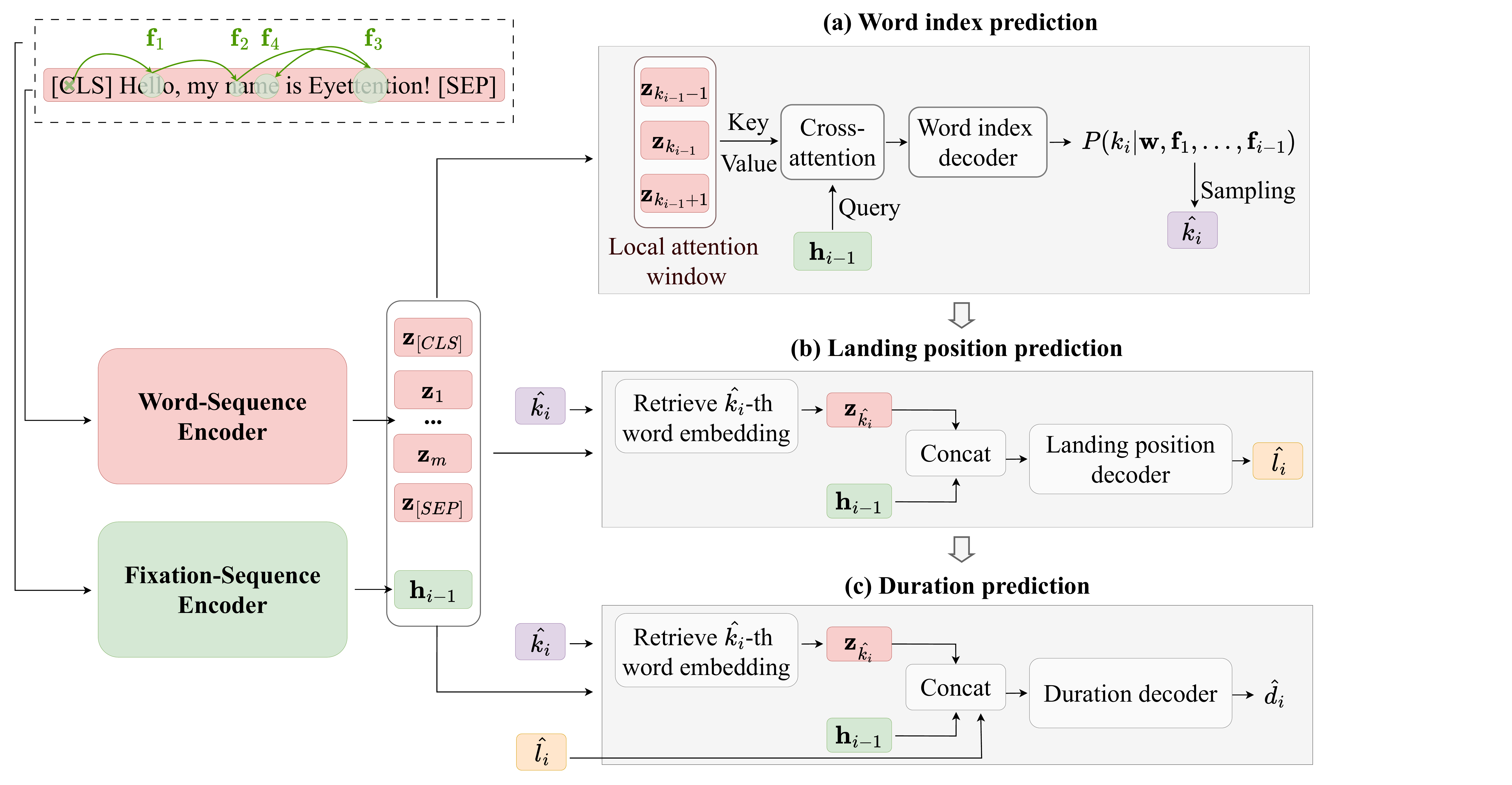}
	  \caption{Overview of the Eyettention~\Romannum{2} model for next fixation prediction. The model predicts three attributes of the next fixation sequentially, with each attribute associated with its own dedicated prediction module: (a) word index $k$ prediction, (b) landing position $l$ prediction, and (c) fixation duration $d$ prediction. The arrow \raisebox{0.1ex}{\includegraphics[trim=20 5 20 5, clip]{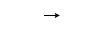}} indicates the flow of information between model components, including both inputs and outputs, while the arrow \raisebox{-1.2ex}{\includegraphics[height=3.3ex, trim=16 16 16 16, clip]{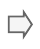}} indicates the sequential order in which fixation attributes are predicted.
	  }
	\label{fig:Eyettention}
\end{figure*}

The architecture of Eyettention~\Romannum{2} closely follows that of the original Eyettention model~\cite{deng2023eyettention}, with 
two key enhancements: Eyettention~\Romannum{2} is designed to predict both the spatial and temporal dynamics of scanpath. This includes not only the word index but also the within-word landing position and the duration of each fixation in a scanpath, expanding beyond the original model's focus on word index prediction alone.To achieve this, we introduce two additional prediction modules, namely a module for the prediction of within-word landing position described in (b) below and a module for the prediction of fixation duration described in (c), while keeping the other components of the model identical to those of the original Eyettention model.
Details regarding hyperparameter tuning for the new modules, along with implementation details, can be found in Section~\ref{sec_implementation}. 
The overall model architecture is depicted in Figure~\ref{fig:Eyettention}. 

\subsubsection{Overview}
The Eyettention~\Romannum{2} model takes a stimulus sentence or text (a \textit{word sequence}) as input and iteratively generates a sequence of fixations on this stimulus. Each fixation is characterized by its \textit{word index} (the location within the word sequence), its \textit{landing position} within that word, and its \textit{duration}.

To generate the next fixation, the model is provided with the initial part of the scanpath, that is, all fixations previously generated for this stimulus. For each fixation, the model not only uses the three fixation attributes (location, landing position, and duration) but also incorporates semantic information about the fixated word in the form of its word embedding. The first fixation is generated based on a special token that marks the beginning of the scanpath.

A Fixation-Sequence Encoder processes this initial scanpath into a hidden representation, which then ``looks at'' the stimulus sentence through cross-attention. Crucially, the model attends only to the words around the currently fixated word, thereby mimicking the limited span of the human visual field. The attended word representations are contextualized embeddings computed by the Word-Sequence Encoder, which uses a pretrained language model as its core component.

In sum, the model generates fixations on a given stimulus in an autoregressive manner, that is, it produces them one by one in chronological order, relying only on past fixations, the words that have already been viewed, and the words surrounding the current fixation. In this way, the model simulates human reading behavior, as it does not consult future fixations or words that have not yet been fixated when generating the next fixation.

\subsubsection{Technical Details}
In the following, we describe the technical details of all model components.
As mentioned above, following \cite{deng2023eyettention}, Eyettention~\Romannum{2} employs a dual-sequence bi-encoder structure,
where the Word-Sequence Encoder processes a sequence of words $\langle w_1,\dots,w_m\rangle$ and extracts word embeddings $\langle \mathbf{z}_1,\dots,\mathbf{z}_m\rangle$,  and the Fixation-Sequence Encoder encodes the initial fixation sequence $\langle \mathbf{f}_1,\dots,\mathbf{f}_{i-1}\rangle$ (consisting of the three fixation attributes location (\textit{word index}, \textit{landing position} and \textit{duration}) together with an embedding of the fixated word) into the representation vector $\mathbf{h}_{i-1}$ at time $i-1$. We first describe how the model uses the output of these two encoders to predict the three fixation attributes via the \textit{Word Index Prediction} module (see (a) in Figure~\ref{fig:Eyettention}), the \textit{Landing Position Prediction} module (see (b) in Figure~\ref{fig:Eyettention}), and the \textit{Duration Prediction} module (see (c) in Figure~\ref{fig:Eyettention}). Subsequently, we will describe the technical details of the two encoders. 

\paragraph{(a) Word Index Prediction}
\label{sec_wordidx}

For predicting a fixation, the model first predicts its location, expressed in terms of the index of the fixated word (\textit{word index}). To this end, the outputs of the two encoders are aligned using a windowed Gaussian version of cross-attention.
 Specifically, we use a local attention window $[\mathbf{z}_{k_{i-1}-D}, \mathbf{z}_{k_{i-1}+D}]$, centered on the embedding of the currently fixated word $\mathbf{z}_{k_{i-1}}$, to simulate the human foveal and parafoveal vision, which, in a typical experimental set-up, extends a few words to the left and right of the fixation location. 
We compare the query $\mathbf{h}_{i-1}$ with the word embeddings $\mathbf{z}_n$ within the window ($k_{i-1}-D\leq n \leq k_{i-1}+D$), which serve as keys, to compute their attention weights $\mathbf{a}_{i}(n)$, reflecting their relevance for deciding where to fixate next. This is achieved using dot-product attention followed by Gaussian smoothing.
The Gaussian smoothing is applied to account for the parafoveal information intake around the currently fixated word.

\begin{equation*}
\displaystyle
\mathbf{a}_{i}(n) = \frac{\exp\ (\mathbf{h}_{i-1}^T W_{a}\mathbf{z}_{n})}{\sum_{j=k_{i-1}-D}^{k_{i-1}+D}\exp\ (\mathbf{h}_{i-1}^T W_{a}\mathbf{z}_{j})} \underbrace{\exp(-\frac{(n-k_{i-1})^{2}}{2\sigma^{2}})}_\text{Gaussian kernel}\ ,
\end{equation*}
where $W_{a}$ represents the learnable weights to align $\mathbf{h}_{i-1}$ and $\mathbf{z}_n$. 
We compute a weighted average of the word embeddings within the window according to their calculated attention weights. $$\mathbf{c}_{i}=\sum_{n=k_{i-1}-D}^{k_{i-1}+D}\mathbf{a}_{i}(n)\mathbf{z}_n.$$

 We concatenate $\mathbf{c}_{i}$ with $\mathbf{h}_{i-1}$ to form the final output of the cross-attention mechanism. This combined vector is then fed into the word index decoder, which consists of feed-forward layers with rectified linear unit (ReLU) activation. We consider the saccade target selection as an inherently stochastic process, similar to SWIFT~\cite{seelig2020bayesian} and DeepGaze~\Romannum{3}~\cite{kummerer2022deepgaze}. Consequently, a softmax layer is added as the last layer, which outputs a probability distribution over candidate word indices $P(k_i|\mathbf{w},\mathbf{f}_1,\dots,\mathbf{f}_{i-1})$\footnote{The model's output is actually a likelihood function over all possible saccade ranges; an intermediate step converts this to the likelihood function over word index by summing the saccade range to the current word index.}. To compute a point estimate $\hat{k_i}$, we sample from this distribution.

\paragraph{(b) Within-Word Landing Position Prediction}
\label{sec_landpos}
Eyettention~\Romannum{2} incorporates a new module specifically for predicting the within-word landing position after determining the next fixated word index $\hat{k_i}$. Once $\hat{k_i}$ is predicted, we retrieve the corresponding word embedding $\mathbf{z}_{\hat{k_i}}$ from the output of the Word-Sequence encoder. This embedding $\mathbf{z}_{\hat{k_i}}$ is then concatenated with the fixation sequence representation $\mathbf{h}_{i-1}$, and fed into a landing position decoder, which consists of feed-forward layers with ReLU activation. In brief, the next landing position is predicted based on the word to be fixated next and the scanpath history.

\paragraph{(c) Fixation Duration Prediction}
\label{sec_dur}
Eyettention~\Romannum{2} includes a new module for predicting fixation duration after determining the next fixated word index $\hat{k_i}$ and landing position $\hat{l_i}$. We concatenate the retrieved word embedding $\mathbf{z}_{\hat{k_i}}$, the fixation sequence representation $\mathbf{h}_{i-1}$, and the predicted landing position $\hat{l_{i}}$. This combined representation is subsequently fed into a duration decoder, consisting of feed-forward layers with ReLU activation. In brief, the prediction of fixation duration is based on the next word to be fixated, the next landing position, and the scanpath history.

\paragraph{Encoders}
\label{sec_encoder}
\begin{figure*}[!ht]

 \subfloat[Word-Sequence Encoder\label{subfig:word_encoder}]
	{   \includegraphics[width=0.5\textwidth,keepaspectratio]{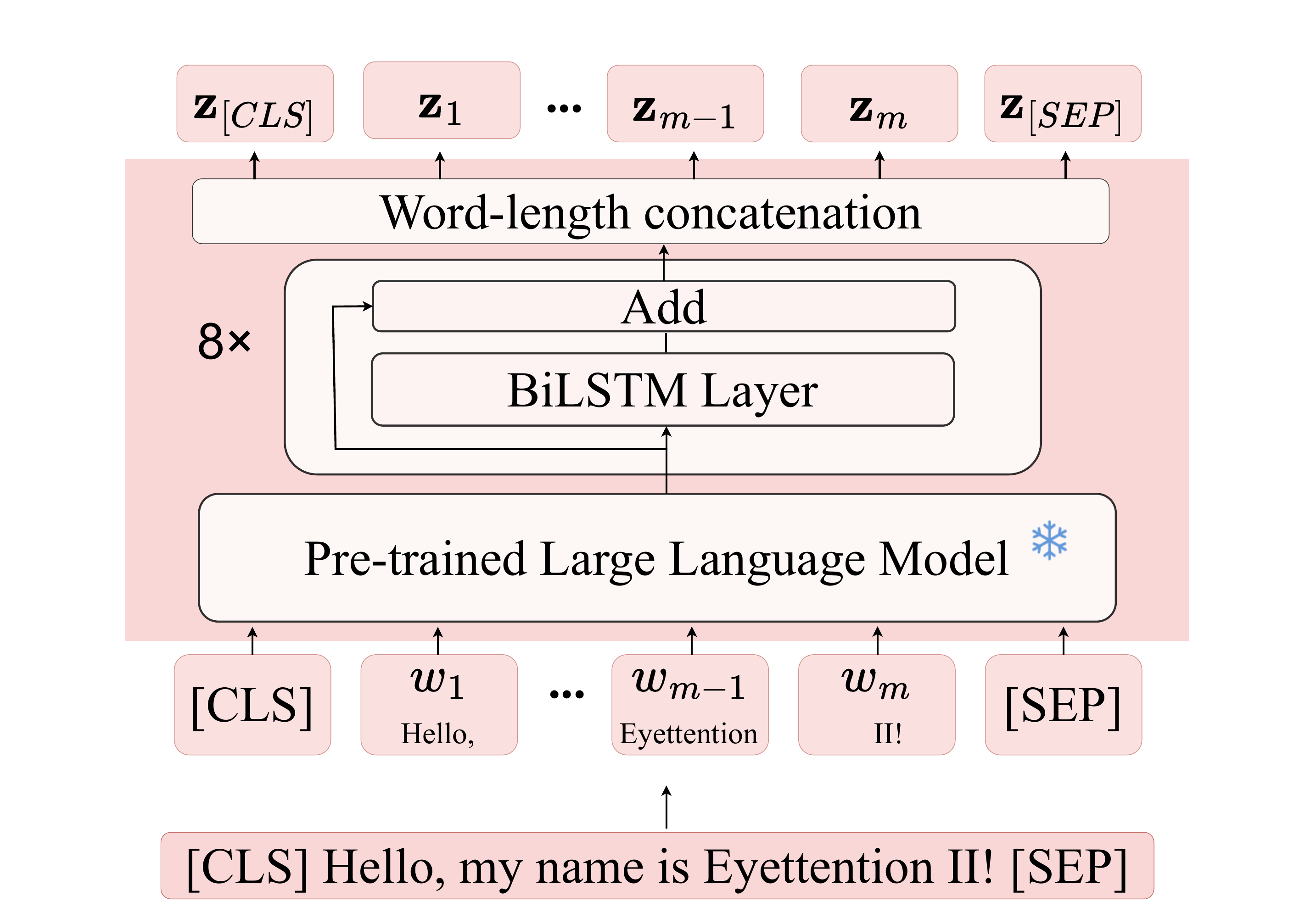}
    }
   \subfloat[Fixation-Sequence Encoder\label{subfig:fix_encoder}]
	{   \includegraphics[width=0.5\textwidth,keepaspectratio]{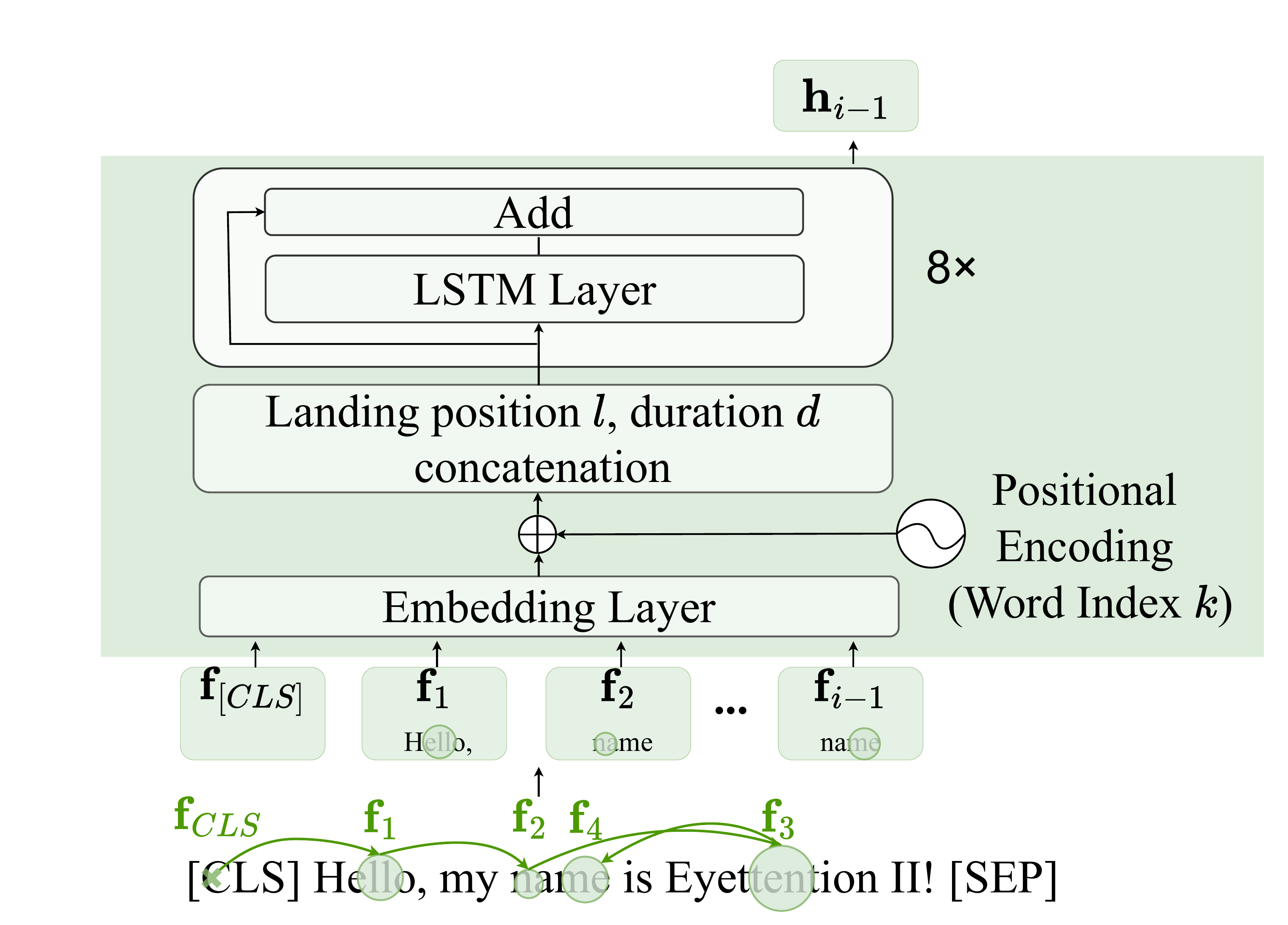}
    }
	  \caption{Bi-encoders for processing dual-sequence eye-tracking-while-reading data: (a) Word-Sequence Encoder for embedding the linguistic features, and (b) Fixation-Sequence Encoder for embedding the temporal sequence of fixations. The pretrained LLM is frozen during the training of Eyettention~\Romannum{2}.
	  }
	\label{fig:encoder}
\end{figure*}

We build upon the design of the encoders from the original Eyettention model, incorporating a bidirectional Word-Sequence Encoder for embedding the linguistic stimulus sequence and a unidirectional Fixation-Sequence Encoder for embedding the temporal sequence. The bidirectional encoder over the linguistic stimulus allows the model to incorporate both left and right contextual information when constructing contextualized token representations. In contrast, the unidirectional Fixation-Sequence Encoder ensures that predictions of future fixations cannot be observed or used, preventing the model from attending to information from fixations that have not yet been generated. Importantly, while the model has access to the entire sentence to build contextualized word representations, it does not have access to future fixations, nor the embeddings of words that have neither been fixated yet nor are contained in the local window around the currently fixated word.  This design preserves autoregressivity in the temporal prediction dimension while still enabling the use of rich linguistic information.

The Word-Sequence Encoder, depicted in Figure~\ref{subfig:word_encoder}, transforms a sequence of words into a corresponding list of word embeddings. To achieve this, we employ a pre-trained LLM, such as BERT$_{\text{base}}$~\cite{devlin2018bert} or RoBERTa$_{\text{base}}$~\cite{liu2019roberta}, 
that tokenizes the input sentence into sub-word units and returns their embeddings. The embedding for each word is computed by summing the embeddings of its constituting sub-word tokens. As the LLM remains frozen during training, we subsequently feed these word embeddings into a deep stacked bidirectional long short-term memory (BiLSTM)~\cite{hochreiter1997long, graves2005framewise} network to encode the relevant linguistic features for the task of scanpath prediction. We newly introduce residual connections among the BiLSTM layers\footnote{Due to the differing dimensions between the LLM embeddings and the LSTM hidden states, we have chosen to omit residual connections for the first LSTM layer to avoid introducing many additional parameters by applying linear projections to align the dimensions.} to facilitate the effective training of the network and enhance model performance. Finally, the output of the BiLSTM network is concatenated with the word length feature to produce the final word embedding $\mathbf{z}$.

The Fixation-Sequence Encoder, depicted in~\ref{subfig:fix_encoder}, generates a hidden vector that encodes the context of preceding fixations. We prepend a special ``fixation" token $\mathbf{f}_0$ for [CLS] to the fixation sequence to denote the start. The input representation for each fixation is constructed by summing the embedding of the fixated word and a trainable position embedding that encodes the word index $k$. The position embeddings are defined as absolute, trainable embeddings learned jointly with the fixation encoder. This follows the standard approach introduced in the Transformer architecture~\cite{vaswani2017attention}. These input representations are then concatenated with the fixation duration and landing position. To encode the fixation sequence features for the task, we utilize a deep-stacked LSTM network with residual connections among the layers.  The output of the Fixation-Sequence Encoder is given by the final hidden state $\mathbf{h}_{i-1}$ of the last LSTM layer.

\subsection{Reader-Specific Eyettention~\Romannum{2}$_\text{reader}$}
\label{sec_reader_specific}
Eyettention~\Romannum{2} is designed to predict generic scanpaths that are not specific to any individual reader. However, eye movements during reading exhibit strong idiosyncratic patterns~\cite{bargary2017individual,JaegerECML2019,makowski2021deepeyedentificationlive}, and depending on the intended application, simulating reader- or group-specific scanpaths can be crucial—for instance, for systems tailored to a particular user or for piloting psycholinguistic experiments involving different populations such as L1 and L2 readers.
To address this need, we develop Eyettention~\Romannum{2}$_\text{reader}$, a variant of Eyettention~\Romannum{2} that simulates the eye-movement behavior of a specific reader.

For training Eyettention~\Romannum{2}$_\text{reader}$, we are given a set of sentences with associated scanpaths. For each scanpath, 
the data includes a reader ID, allowing us to build user-specific models. To achieve this, we introduce a trainable reader identifier embedding with an embedding size of $d_{\text{emb}}$ into our
model. This embedding is concatenated
with the input representation of each fixation in the Fixation-Sequence Encoder.

Importantly, this reader-specific architecture can also be used to simulate group-specific behavior. In practice, this simply requires replacing the reader ID with a group ID. In the computational experiments presented in this paper, we focus on generic and reader-specific scanpaths and leave the modeling of group-specific scanpaths to future work.

\subsection{Model Optimization}
\label{sec_optimiatzion}
We train the model to generate scanpaths that resemble human eye-movement patterns. To do so, we first need to define a loss function that operationalizes what it means for two scanpaths to “resemble” each other.  For simplicity, we define separate loss functions for each of the three fixation attributes and minimize an objective function that combines three loss terms. Specifically, we use negative log-likelihood (NLL) of the predicted \textit{word index} $k$, the mean-squared-error (MSE) between the estimated landing position $\hat{l}$ and the ground-truth landing position $l$ in the training data, and the MSE between the estimated fixation duration $\hat{d}$ and the ground-truth fixation duration $d$ provided by the training data. The loss function for a scanpath consisting of $I$ fixations is defined as:
\begin{align*}
\mathcal{L} &= \alpha\mathcal{L}_{\text{word index}} + \beta\mathcal{L}_{\text{landing position}} + \gamma\mathcal{L}_{\text{duration}} \\
&= \alpha\frac{1}{I}\sum_{i=1}^{I}-logP(k_{i}|\mathbf{w},\mathbf{f}_{0},\dots,\mathbf{f}_{i-1})
+ \beta\frac{1}{I}\sum_{i=1}^{I}(\hat{l_{i}} - l_{i})^{2}
+ \gamma\frac{1}{I}\sum_{i=1}^{I}(\hat{d_{i}} - d_{i})^{2}
\end{align*}
 The coefficients $\alpha$, $\beta$, and $\gamma$ are chosen to balance the three loss terms (see Table~\ref{tab: hp} for the optimized values after hyperparameter tuning), ensuring they have similar magnitudes in the fully trained model. We use teacher forcing~\cite{williams1989learning} instead of auto-regressive training to enable faster and more stable convergence.

\section{Computational Experiments}
\label{sec_exp}
We evaluate Eyettention~\Romannum{2} on two typologically different languages with different types of scripts, English and Chinese, using a range of publicly available eye-tracking-while-reading datasets, and compare its performance against state-of-the-art models. Our evaluation includes both within-~(\textsection~\ref{sec_withindata_eval}) and across-dataset evaluation~(\textsection~\ref{sec_crossdata_eval}) to assess the model's generalization capabilities across various use cases. Lastly, we conduct a series of analyses~(\textsection~\ref{sec_model_behavior}) to gain deeper insights into the model's effectiveness.




\subsection{Datasets}
\label{sec_datasets}

We use four eye-tracking-while-reading corpora to train, tune, and/or evaluate our proposed model and the reference methods. These datasets differ in terms of stimulus language and script (English vs Chinese), as well as stimulus layout and eye-tracking hardware used for recording. Descriptive statistics for each of the datasets are presented in Table~\ref{tab:descriptive-corpora-stats}.

\bmhead*{BSC}
The \emph{Beijing Sentence Corpus} (BSC)~\cite{pan2021bsc} comprises eye-tracking data collected from native Chinese speakers as they read sentences from Chinese newspapers, with each sentence read by multiple readers.

\bmhead*{CELER}
The \emph{Corpus of Eye Movements in L1 and L2 English Reading} (CELER)~\cite{berzak2022celer} contains eye-tracking data from both native (L1) and non-native (L2) English speakers that read English newswire sentences. For the purpose of our study, we use only the data from native speakers (CELER~L1). In this dataset, half of the sentences are shared across readers while the other half are unique to each reader. 

\bmhead*{ZuCo}
The \emph{Zurich Cognitive Language Processing Corpus} (ZuCo~\cite{hollenstein2018zuco} and ZuCo~2.0~\cite{hollenstein2019zuco2})  contain eye-tracking data from native English speakers that read  English movie reviews and Wikipedia articles. These corpora include two distinct reading paradigms: ``task-specific'' and ``natural reading''. We use the ``natural reading'' subset (ZuCo/ZuCo~2.0~NR) in our study. Each sentence is read by multiple readers.

\begin{table}[t!]
\begin{footnotesize}
  \caption{Descriptive statistics of the four eye-tracking-while-reading corpora used for model training and evaluation. The number of words per sentence is reported using the mean $\pm$ standard deviation.}
    \centering
    \begin{tabular}{l|l|c|c|c|l}
    \toprule
           & &\# Unique & \# Words per &  &  \\
         Dataset & Eye-tracker & sentences & sentence & \# Readers  & Language \\ \hline
         BSC~\cite{pan2021bsc} & Eyelink~II (500~Hz)& $150$ &$11.2\pm1.6$ & $60$ & Chinese  \\
         CELER~L1~\cite{berzak2022celer} & Eyelink~1000 (1000~Hz) & 5456 & 11.2 $\pm$ 3.6 & $69$ & English \\
         ZuCo~NR~\cite{hollenstein2018zuco} & EyeLink~1000~Plus (500~Hz)& $700$& 19.6 $\pm$ 9.8& $12$&English\\
         ZuCo~2.0~NR~\cite{hollenstein2019zuco2} &EyeLink~1000~Plus (500~Hz) & $349$&$19.6\pm8.8$& $18$&English\\
         \bottomrule
    \end{tabular}
    \label{tab:descriptive-corpora-stats}
\end{footnotesize}
\end{table}

\subsection{Reference Methods}
\label{sec_ref}

We include several baseline models in our evaluation to benchmark our model against both established computational cognitive models of eye-movement control in reading and machine learning-based state-of-the-art approaches.  

\bmhead*{Eyettention~\cite{deng2023eyettention} \& Mean Baseline}Since the original Eyettention model only predicts fixation locations, but neither landing position within a word nor fixation duration, we benchmark our model against a slightly modified version of Eyettention where we combine the original model with a mean baseline model which predicts landing position and fixation duration as the mean values derived from the training data for these variables.

\bmhead*{\citet{hollenstein-etal-2021-multilingual}} The pre-trained BERT model with a linear dense layer is used to predict fixation duration. The model is fine-tuned on the task data\footnote{\label{note1}To compare with our model, we 
fit and evaluate the model based on individual fixation durations in a scanpath, rather than aggregated word-level reading measures such as first fixation duration or total reading time on a given word as in the original study.}.

\bmhead*{Last~\cite{bestgen-2021-last}} This approach feeds a number of hand-crafted features, 
such as word frequency, word length, POS-tags, lexical characteristics, and behavioral measures into a LightGBM model for fixation duration prediction\footref{note1}. It achieved top performance in the 2021 CMCL Shared Task on Eye-Tracking Data Prediction~\cite{hollenstein-etal-2021-cmcl}.

\bmhead*{E-Z reader~\cite{rayner2007chineseezreader, reichle2003ezreader}} The E-Z Reader model is a serial, stage-based cognitive model of eye movements in reading, in which attention and linguistic processing are assumed to shift sequentially across words. Lexical processing is divided into two stages: an early familiarity check stage that triggers saccadic programming, and a later lexical access stage, after which attention shifts to the next word. The model predicts fixation locations, within-word landing positions, and fixation durations by linking progress in lexical processing to the timing and execution of saccades, and by modulating lexical processing with factors such as word frequency and predictability. Higher-level processing difficulty can delay or change planned eye movements, accounting for regressions and increased fixation times under comprehension difficulty.

\bmhead*{SWIFT~\cite{rabe2021bayes}} The SWIFT model is a cognitive model of eye movement control in reading that proposes parallel processing of multiple words. It predicts fixation positions, with-word landing positions, and fixation durations by incorporating linguistic features such as word length and lexical frequency.

\bmhead*{Human}  We compute inter-reader scanpath similarity by randomly selecting scanpaths from two different individuals reading the same sentence and reporting the mean similarity across all pairs of scanpaths.

\subsection{Evaluation Metrics}
\label{sec_metric}
There is not one established metric standardly used for evaluating generative models of eye movements. The different metrics highlight different aspects of scanpath generation. 
Hence, we use two types of evaluation metrics to assess the models' performance from different perspectives.

\bmhead*{Evaluating Conditional Fixation Predictions} We follow the work of~\cite{deng2023eyettention, kummerer2022deepgaze} to assess each fixation in a scanpath individually, evaluating the model's ability to predict each fixation when given information about the reader's previous fixations in a scanpath. This approach examines whether the model and a human reader make similar fixation decisions under identical conditions---that is, when conditioned on the same sequence of prior fixations made by the human reader. Importantly, this evaluation method is parameter-free. We evaluate each fixation attribute separately: negative log-likelihood (NLL) for word index prediction and mean absolute error (MAE) for landing position and fixation duration. The NLL measures the average negative log-probability that the model assigns to the true fixation locations in the scanpath observed in the human training data, reflecting how accurately the model predicts each fixation. Lower NLL values indicate that the model assigns higher probabilities to the correct fixation targets, corresponding to better predictive performance. The MAE measures the average absolute deviation between the predicted and observed fixation attributes, indicating how closely the model’s continuous predictions (landing position and fixation duration) align with human data.
For each attribute, the final score for a model is computed by first averaging the scores across fixations within each scanpath, followed by averaging across all scanpaths.

\bmhead*{Evaluating Scanpath Similarity}  
We evaluate the quality of the model's scanpath generation by comparing actual human scanpaths with model-generated scanpaths. For this, we employ the widely adopted MultiMatch metric~\cite{jarodzka2010vector}, which assesses scanpath similarity 
by computing distances across five dimensions: saccade vector, saccade length, saccade direction, fixation position, and fixation duration. We apply the default parameters provided in the respective implementations~\cite{wagner2019multimatch} for consistency.

\subsection{Hyperparameter Tuning and Implementation Details}
\label{sec_implementation}

\begin{table*}[t!]
\caption{Hyper-parameter grid for the Eyettention~\Romannum{2} model.}
\label{tab: hp}
\begin{center}
\begin{footnotesize}
    \begin{tabular}{l|l|l}
    \toprule
    Hyper-parameter       & Search space  & Best value           \\\midrule
    
    Landing position decoder: \# dense layers & \{0, 2, 4, 6\} & 4\\
    Landing position decoder: \# dense units & \{64, 128, 256, 512, 1024\} & {128, 128, 128, 128}\\
    Duration decoder: \# dense layers & \{0, 2, 4, 6\} & 2\\
    Duration decoder: \# dense units & \{64, 128, 256, 512, 1024\} & {1024, 64}\\
    $\alpha$ & \{0.005, 0.01, 0.02, 0.03, 0.05, 0.1\}& 0.02\\
    $\beta$& \{0.1, 0.2, 0.3, 0.4, 0.5, 0.7, 1\}&0.3\\
    $\gamma$&\{0.1, 0.5, 0.8, 1, 3, 5\} &1\\
    $\sigma$ &\{0.5, 0.75, 1, 1.25, 1.5, 2, 3\} &0.75\\
    
    \bottomrule
    \end{tabular}
    \end{footnotesize} 
\end{center}

\end{table*}

We perform hyperparameter tuning to optimize the newly added prediction modules for landing position and duration. Hyperparameters are model settings that are chosen before training---such as the number of layers or the size of each layer---and can strongly affect model performance. To select them, we use a random grid search. We train the model on the CELER L1 dataset and validate the model on the held-out dataset ZuCo 2.0 NR. The tuning set is discarded after the hyperparameter optimization and not used in any further experiments. Table~\ref{tab: hp} shows the hyperparameter grid used for tuning and indicates the 
configuration found to be optimal. For the remaining model configuration, we maintain the same settings as the previous version of Eyettention~\cite{deng2023eyettention}, which can be found in Table~\ref{tab: Eyettention_configure} of the Appendix.

We report results using two LLMs in the text encoder: BERT$_{\text{base}}$~\cite{devlin2018bert} and RoBERTa$_{\text{base}}$~\cite{liu2019roberta}. For each token, we extract its contextualized representation from the final layer of the full pretrained LLM, which is then used as input to the BiLSTM layer. The Fixation-Sequence Encoder uses absolute, trainable position embeddings to represent the word index of each fixation. These embeddings have the same dimensionality as the LLM embeddings, enabling their element-wise summation. These embeddings are initialized randomly and optimized jointly with the model parameters. The entire Eyettention~\Romannum{2} model comprises 3.7 million trainable parameters.
For scanpath generation, we use nucleus sampling with $p = 0.7$ for the BSC data, and temperature sampling with $t = 0.04$ for the CELER data. A detailed analysis of the effect of different decoding strategies can be found in Section~\ref{sec_decode_strategy}. The network weights are optimized using the Adam optimizer~\cite{kingma2014adam} with a learning rate of 1e-3. We train the models for 1000 epochs with early stopping (patience of 20 on a validation set sampled from the training data) and a batch size of 256. All neural networks are trained using the PyTorch~\cite{pytorch2019paszke} library on an NVIDIA A100-SXM4-40GB GPU via the NVIDIA CUDA platform.

\subsection{Within-Dataset Evaluation}
\label{sec_withindata_eval}

\begin{figure}[t!]
	\centering
	  \includegraphics[width=\textwidth,keepaspectratio,trim={0cm 0 0cm 0}, clip]
   {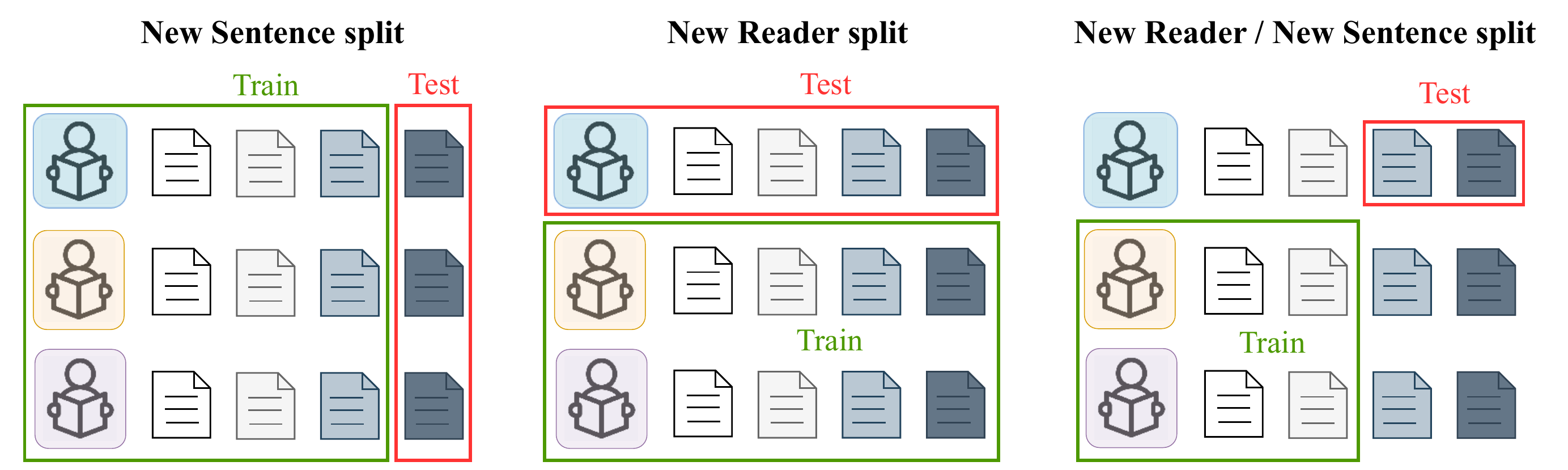}
	  \caption{Overview of the three different train/test splits applied in the within-dataset evaluation. \textbf{New Sentence split}: Models are tested on scanpaths on sentences not included in the training data, but generated by known readers from whom scanpaths on other sentences are included in the training data.  \textbf{New Reader split}: Models are tested on scanpaths from new readers who are not included in the training data,  on sentences that are included in the training data but read by different readers. \textbf{New Sentence\,/\,New Reader split}: Models are tested on scanpaths generated by new readers not contained in the training data reading new sentences not contained in the training data neither. 
	  }
	\label{fig:three_splits}
\end{figure}

In the within-dataset evaluation, we train and test the models on the same dataset, dividing the data into separate training and test subsets to assess the models' in-domain generalization capabilities. The evaluation is performed on two datasets: BSC~\cite{pan2021bsc}, which uses Chinese stimuli, and CELER~\cite{berzak2022celer}, which uses English stimuli. We explore three distinct train/test splits, each representing different use case scenarios, as illustrated in Figure~\ref{fig:three_splits}. We apply identical data splits across all models to ensure a fair comparison.

(\romannum{1}) \textbf{New Sentence split}: We evaluate how the models perform
when predicting scanpaths of a known sample of readers on novel sentences. We conduct 5-fold hold-out cross-validation, where the train/test split is based on sentence IDs. In each fold, we use 80\% of the sentences for training, and the remaining 20\% are reserved for testing, ensuring that all models are tested on sentences not seen during training. 

(\romannum{2}) \textbf{New Reader split}: We evaluate the models’ capacity to generalize to novel readers when generating scanpaths for known sentences. The model is trained on scanpaths from a subset of readers and tested on a separate set of readers, ensuring that no eye-movement data from the test readers is seen during training. We perform 5-fold cross-validation, with an 80/20\% train/test split based on reader IDs, ensuring the models are tested on readers not encountered during training.  

(\romannum{3}) \textbf{New Sentence\,/\,New Reader split}: This split examines the models' ability to generalize to both novel readers and novel sentences at the same time. To maintain an 80/20\% train/test split based on both reader IDs and sentence IDs, we perform random re-sampling, repeated 5 times. Consequently, the models are tested on sentences and readers that are not seen during training.

\bmhead*{Results}
We present an overview of the results for Eyettention~\Romannum{2} and the reference methods in Table~\ref{tab: res_BSC} for the BSC (Chinese) dataset, and in Table~\ref{tab: res_CELER} for the CELER (English) dataset.  
Overall, Eyettention~\Romannum{2} consistently outperforms all reference methods. Whether BERT or RoBERTa is used in the Word Sequence Encoder, Eyettention~\Romannum{2} achieves similarly strong performance.
In evaluating \emph{conditional fixation prediction}, Eyettention~\Romannum{2} improves upon the previous state-of-the-art. In terms of the prediction of fixation location (word index) prediction it outperforms the previous Eyettention model~\cite{deng2023eyettention}, by 2-3\% boost on the BSC dataset and by 6\% on CELER. 
In terms of fixation duration prediction, Eyettention~\Romannum{2} demonstrates a significant advantage over state-of-the-art methods, outperforming~\cite{hollenstein-etal-2021-multilingual} or/and Last~\cite{bestgen-2021-last} by 5-6\% on BSC and 4-5\% on CELER. Unlike these methods, which focus on predicting aggregated word-level fixation durations, Eyettention~\Romannum{2} incorporates scanpath history, leading to superior performance in predicting sequential fixations. For prediction of landing position, Eyettention~\Romannum{2} surpasses the Mean baseline by 9-11\% on BSC and 17\% on CELER. Furthermore, in evaluating \emph{scanpath similarity}, Eyettention~\Romannum{2} significantly outperforms the cognitive models E-Z reader~\cite{rayner2007chineseezreader, reichle2003ezreader} and SWIFT~\cite{rabe2021bayes} across all dimensions of the MultiMatch metrics. The Human baseline yields relatively lower similarity scores than our model, reflecting the considerable individual variability in human reading behaviors.

\bmhead*{Results Eyettention~\Romannum{2}$_\text{reader}$}

We evaluate the performance of Eyettention~\Romannum{2}$_\text{reader}$ (\textsection~\ref{sec_reader_specific}) 
in predicting scanpaths of known individuals on novel sentences by incorporating reader-specific embeddings that encode reader identity. We compare this version to the base Eyettention~\Romannum{2} model without these identity embeddings, as shown in Table~\ref{tab: reader_specific}. We observe that individualizing the scanpath prediction to specific readers significantly improves the model’s performance across all fixation attributes, highlighting the effectiveness of the Eyettention~\Romannum{2}$_\text{reader}$ model. The improvements are consistent across various reader embedding sizes ranging from 16 to 128, though using an embedding size of 128 can occasionally lead to worse results.

\begin{table*}[]
\caption{\textbf{Within-Dataset Evaluation on BSC (Chinese) dataset.} The standard error is indicated as the subscript. The dagger $\dagger$ indicates that a model is significantly worse than the best model. Statistical tests were performed using a two-tailed t-test, with $p < 0.05$.}
\label{tab: res_BSC}
\begin{center}
\fontsize{5}{8}\selectfont

\begin{tabular}{lp{1cm}p{1cm}p{1cm}p{1cm}p{1cm}p{1cm}p{1cm}}
    \toprule
    & Word Idx  & Duration    & Land Pos &\multicolumn{4}{c}{MultiMatch}\\\cmidrule{5-8}
   Model                   & NLL$\downarrow$  & MAE$\downarrow$  & MAE$\downarrow$ &Vector$\uparrow$  &Length$\uparrow$ &Position$\uparrow$  &Duration$\uparrow$\\\midrule\midrule

     \multicolumn{8}{c}{\emph{New Sentence Split}}    \\\midrule
    
    Eyettention~\cite{deng2023eyettention}\&Mean                   &  1.854\textsubscript{0.01}$\dagger$               & 66.856\textsubscript{0.28}$\dagger$                                         & 0.235\textsubscript{0.00}$\dagger$  & 0.990\textsubscript{0.00}$\dagger$ & 0.981\textsubscript{0.01}$\dagger$ & \textbf{0.868}\textsubscript{0.09} & 0.726\textsubscript{0.19}$\dagger$\\
    
     \citet{hollenstein-etal-2021-multilingual}    &--         & 65.311\textsubscript{0.27}$\dagger$        &--  & --  & -- & --&--\\


   E-Z Reader~\cite{rayner2007chineseezreader} & --& --& -- &0.956\textsubscript{0.03}$\dagger$ &0.935\textsubscript{0.04}$\dagger$ &0.855\textsubscript{0.07}$\dagger$ &0.636\textsubscript{0.11}$\dagger$\\

   Eyettention~\Romannum{2}-BERT
   & \textbf{1.793}\textsubscript{0.01}                  & 61.429\textsubscript{0.29}   & \textbf{0.208}\textsubscript{0.00} &\textbf{0.993}\textsubscript{0.00} &\textbf{0.987}\textsubscript{0.01} &0.864\textsubscript{0.09}$\dagger$
&\textbf{0.743}\textsubscript{0.18}\\
  Eyettention~\Romannum{2}-RoBERTa
  & \textbf{1.793}\textsubscript{0.01}                   & \textbf{61.384}\textsubscript{0.28}   & 0.209\textsubscript{0.00}
  &\textbf{0.993}\textsubscript{0.00} &\textbf{0.987}\textsubscript{0.01} &0.857\textsubscript{0.09}$\dagger$
&0.731\textsubscript{0.20}$\dagger$\\

 Human & --& --& -- &0.991\textsubscript{0.01}$\dagger$ &0.984\textsubscript{0.01}$\dagger$ &0.856\textsubscript{0.09}$\dagger$ &0.668\textsubscript{0.20}$\dagger$
  \\\midrule

 \multicolumn{8}{c}{\emph{New Reader Split}}    \\\midrule
    
   Eyettention~\cite{deng2023eyettention}\&Mean                   &  1.872\textsubscript{0.01}$\dagger$              & 65.798\textsubscript{0.31}$\dagger$                                         & 0.235\textsubscript{0.00}$\dagger$ &0.990\textsubscript{0.00}$\dagger$  &0.981\textsubscript{0.01}$\dagger$  &\textbf{0.866}\textsubscript{0.09}
&0.726\textsubscript{0.18}$\dagger$  \\
    
    \citet{hollenstein-etal-2021-multilingual}    &--         & 66.715\textsubscript{0.26}$\dagger$             &--& --& --& --& --\\


   E-Z Reader~\cite{rayner2007chineseezreader} & --& --& -- &0.956\textsubscript{0.03}$\dagger$  &0.935\textsubscript{0.04}$\dagger$  &0.856\textsubscript{0.07}$\dagger$  &0.633\textsubscript{0.11}$\dagger$ \\

   Eyettention~\Romannum{2}-BERT
   & \textbf{1.804}\textsubscript{0.01}                  & 62.588\textsubscript{0.28}  &\textbf{0.211}\textsubscript{0.00} &\textbf{0.993}\textsubscript{0.00} &\textbf{0.987}\textsubscript{0.01} &0.863\textsubscript{0.09}
&0.730\textsubscript{0.19}\\

     Eyettention~\Romannum{2}-RoBERTa
     & 1.809\textsubscript{0.01}                  & \textbf{62.391}\textsubscript{0.29} & \textbf{0.211}\textsubscript{0.00} &\textbf{0.993}\textsubscript{0.00} &\textbf{0.987}\textsubscript{0.01} &0.862\textsubscript{0.09}$\dagger$ 
&\textbf{0.733}\textsubscript{0.19}\\

 Human & --& --& -- &0.991\textsubscript{0.01}$\dagger$  &0.984\textsubscript{0.01}$\dagger$  &0.853\textsubscript{0.09}$\dagger$  &0.663\textsubscript{0.20}$\dagger$ \\
\midrule

   \multicolumn{8}{c}{\emph{New Sentence / New Reader Split}}    \\\midrule
    Eyettention~\cite{deng2023eyettention}\&Mean                   &  1.834\textsubscript{0.02}               & 67.288\textsubscript{0.62}$\dagger$                                         & 0.237\textsubscript{0.00}$\dagger$  &0.990\textsubscript{0.00}$\dagger$ &0.982\textsubscript{0.01}$\dagger$ &\textbf{0.869}\textsubscript{0.08}
&\textbf{0.746}\textsubscript{0.16}  \\
    
   \citet{hollenstein-etal-2021-multilingual}    &--         & 65.563\textsubscript{0.62}$\dagger$            &--& --& --& --& --\\


    E-Z Reader~\cite{rayner2007chineseezreader} & --& --& -- &0.957\textsubscript{0.03}$\dagger$ &0.936\textsubscript{0.04}$\dagger$ &0.858\textsubscript{0.07}$\dagger$ &0.640\textsubscript{0.11}$\dagger$\\

  Eyettention~\Romannum{2}-BERT
  & \textbf{1.789}\textsubscript{0.02}                 & 62.621\textsubscript{0.63}  & \textbf{0.215}\textsubscript{0.00} 
  &\textbf{0.993}\textsubscript{0.00} &\textbf{0.987}\textsubscript{0.01} & 0.864\textsubscript{0.09}
&0.738\textsubscript{0.18}
\\

     Eyettention~\Romannum{2}-RoBERTa
     & 1.793\textsubscript{0.02}                  & \textbf{62.584}\textsubscript{0.64}  & 0.217\textsubscript{0.00}
     &\textbf{0.993}\textsubscript{0.00} &\textbf{0.987}\textsubscript{0.01} &0.860\textsubscript{0.09}$\dagger$
&0.733\textsubscript{0.20}$\dagger$\\

 Human & --& --& -- &0.991\textsubscript{0.01}$\dagger$ &0.984\textsubscript{0.01}$\dagger$ &0.857\textsubscript{0.09}$\dagger$ &0.663\textsubscript{0.19}$\dagger$\\

    \botrule

    \end{tabular}

\end{center}
\end{table*}

\begin{table*}[]
\caption{\textbf{Within-Dataset Evaluation on CELER (English) dataset.} The standard error is indicated as the subscript. The dagger $\dagger$ indicates that a model is significantly worse than the best model. Statistical tests were performed using a two-tailed t-test, with $p < 0.05$.}
\label{tab: res_CELER}
\begin{center}
\fontsize{5}{8}\selectfont

\begin{tabular}{lp{1cm}p{1cm}p{1cm}p{1cm}p{1cm}p{1cm}p{1cm}}
    \toprule
    & Word Idx  & Duration    & Land Pos &\multicolumn{4}{c}{MultiMatch}\\\cmidrule{5-8}
   Model                   & NLL$\downarrow$  & MAE$\downarrow$  & MAE$\downarrow$ &Vector$\uparrow$  &Length$\uparrow$ &Position$\uparrow$  &Duration$\uparrow$\\\midrule\midrule

     \multicolumn{8}{c}{\emph{New Sentence Split}}    \\\midrule
    
    Eyettention~\cite{deng2023eyettention}\&Mean                   & 2.226\textsubscript{0.01}$\dagger$                & 66.678\textsubscript{0.40}$\dagger$                                         & 0.241\textsubscript{0.00}$\dagger$  &0.975\textsubscript{0.02}$\dagger$ &0.962\textsubscript{0.03}$\dagger$ &0.853\textsubscript{0.12}$\dagger$
&0.719\textsubscript{0.15}$\dagger$\\
    
     \citet{hollenstein-etal-2021-multilingual}    &--         & 66.124\textsubscript{0.40}$\dagger$       &-- & --  & -- & --&--\\
    
 Last~\cite{bestgen-2021-last}  &--                              &  65.952\textsubscript{0.40}$\dagger$     &-- & --  & -- & --&--\\


   SWIFT~\cite{rabe2021bayes} & --& --& -- &0.947\textsubscript{0.03}$\dagger$ &0.918\textsubscript{0.04}$\dagger$ &0.757\textsubscript{0.11}$\dagger$ &0.549\textsubscript{0.27}$\dagger$\\

   E-Z Reader~\cite{ reichle2003ezreader} & --& --& -- &0.971\textsubscript{0.02}$\dagger$ &0.956\textsubscript{0.03}$\dagger$ &0.841\textsubscript{0.09}$\dagger$ &0.609\textsubscript{0.15}$\dagger$\\

 Eyettention~\Romannum{2}-BERT
 & 2.107\textsubscript{0.01}$\dagger$                  & 63.200\textsubscript{0.40} & 0.200\textsubscript{0.00} 
 &\textbf{0.978}\textsubscript{0.02} &\textbf{0.967}\textsubscript{0.02} &0.870\textsubscript{0.11}$\dagger$
&0.764\textsubscript{0.14}\\

     Eyettention~\Romannum{2}-RoBERTa
     & \textbf{2.087}\textsubscript{0.01}            & \textbf{62.828}\textsubscript{0.41} & \textbf{0.199}\textsubscript{0.00} 
     &\textbf{0.978}\textsubscript{0.02} &\textbf{0.967}\textsubscript{0.02} &\textbf{0.873}\textsubscript{0.10}
&\textbf{0.766}\textsubscript{0.14}\\

Human & --& --& -- &0.973\textsubscript{0.02}$\dagger$ &0.958\textsubscript{0.03}$\dagger$ &0.849\textsubscript{0.09}$\dagger$ &0.674\textsubscript{0.17}$\dagger$\\\midrule

 \multicolumn{8}{c}{\emph{New Reader Split}}    \\\midrule
    
   Eyettention~\cite{deng2023eyettention}\&Mean                   & 2.227\textsubscript{0.01}$\dagger$              & 66.841\textsubscript{0.40}$\dagger$                                         & 0.241\textsubscript{0.00}$\dagger$  &0.975\textsubscript{0.02}$\dagger$  &0.963\textsubscript{0.02}$\dagger$  &0.860\textsubscript{0.11}$\dagger$ 
&0.716\textsubscript{0.15}$\dagger$ \\
   \citet{hollenstein-etal-2021-multilingual}    &--         &  65.912\textsubscript{0.40}$\dagger$             &--& --& --& --& --\\
   Last~\cite{bestgen-2021-last}  &--                              &  66.029\textsubscript{0.40}$\dagger$           &--& --& --& --& --\\

 SWIFT~\cite{rabe2021bayes} & --& --& --&0.947\textsubscript{0.03}$\dagger$  &0.918\textsubscript{0.04}$\dagger$  &0.759\textsubscript{0.11}$\dagger$  &0.552\textsubscript{0.27}$\dagger$ 
 \\
 E-Z Reader~\cite{ reichle2003ezreader} & --& --& --&0.971\textsubscript{0.02}$\dagger$  &0.956\textsubscript{0.03}$\dagger$  &0.840\textsubscript{0.09}$\dagger$  &0.609\textsubscript{0.14}$\dagger$ \\

   Eyettention~\Romannum{2}-BERT
   &2.099\textsubscript{0.01}                  & 63.621\textsubscript{0.41}  & \textbf{0.199}\textsubscript{0.00} 
   &\textbf{0.979}\textsubscript{0.02} &\textbf{0.967}\textsubscript{0.02} &\textbf{0.866}\textsubscript{0.12}
&\textbf{0.760}\textsubscript{0.15}\\

  Eyettention~\Romannum{2}-RoBERTa
  & \textbf{2.091}\textsubscript{0.01}                 & \textbf{63.182}\textsubscript{0.41}& \textbf{0.199}\textsubscript{0.00} 
  &0.978\textsubscript{0.02} &\textbf{0.967}\textsubscript{0.02} &0.865\textsubscript{0.12}
&0.757\textsubscript{0.15}\\

Human & --& --& -- &0.973\textsubscript{0.02}$\dagger$  &0.958\textsubscript{0.03}$\dagger$  &0.850\textsubscript{0.09}$\dagger$  &0.676\textsubscript{0.17}$\dagger$ \\
\midrule

   \multicolumn{8}{c}{\emph{New Sentence / New Reader Split}}    \\\midrule
   Eyettention~\cite{deng2023eyettention}\&Mean                   & 2.198\textsubscript{0.01}$\dagger$              & 67.891\textsubscript{0.97}$\dagger$                                        & 0.241\textsubscript{0.00}$\dagger$  &0.977\textsubscript{0.02}$\dagger$ &0.963\textsubscript{0.02}$\dagger$ &0.845\textsubscript{0.14}$\dagger$
&0.723\textsubscript{0.15}$\dagger$\\
   \citet{hollenstein-etal-2021-multilingual}    &--         & 68.151\textsubscript{0.98}$\dagger$             &--& --& --& --& --\\
   Last~\cite{bestgen-2021-last}  &--                              &  67.413\textsubscript{0.98}           &--& --& --& --& --\\


    SWIFT~\cite{rabe2021bayes} & --& --& --&0.949\textsubscript{0.03}$\dagger$ &0.921\textsubscript{0.04}$\dagger$ &0.758\textsubscript{0.11}$\dagger$ &0.554\textsubscript{0.28}$\dagger$
 \\

   E-Z Reader~\cite{ reichle2003ezreader} & --& --& --&0.973\textsubscript{0.02}$\dagger$ &0.957\textsubscript{0.03}$\dagger$ &0.840\textsubscript{0.10}$\dagger$ &0.612\textsubscript{0.15}$\dagger$\\

    Eyettention~\Romannum{2}-BERT
    & 2.073\textsubscript{0.01}                 & \textbf{64.936}\textsubscript{0.99}   & 0.201\textsubscript{0.00}
    &\textbf{0.980}\textsubscript{0.02} &\textbf{0.968}\textsubscript{0.02} &0.873\textsubscript{0.10}
&\textbf{0.769}\textsubscript{0.13}\\

       Eyettention~\Romannum{2}-RoBERTa
       
       & \textbf{2.056}\textsubscript{0.01}                  & 65.055\textsubscript{1.00}    & \textbf{0.200}\textsubscript{0.00}
       &\textbf{0.980}\textsubscript{0.02} &\textbf{0.968}\textsubscript{0.02} &\textbf{0.879}\textsubscript{0.10}
&0.762\textsubscript{0.14}\\

Human & --& --& -- &0.974\textsubscript{0.02}$\dagger$ &0.959\textsubscript{0.02}$\dagger$ &0.855\textsubscript{0.09}$\dagger$ &0.686\textsubscript{0.15}$\dagger$\\

    \botrule

    \end{tabular}

\end{center}
\end{table*}

\begin{table*}[]
\caption{\textbf{New Sentence split for the reader-specific Eyettention~\Romannum{2}$_\text{reader}$-BERT.} $d_{emb}$ denotes the size of the reader-specific embedding. The standard error is indicated as the subscript. The percentages in brackets indicate the percentage improvement over the model without reader embedding. The dagger $\dagger$ indicates models significantly better or worse than the models without reader embeddings. Statistical tests were performed using a two-tailed t-test, with $p < 0.05$.}

\label{tab: reader_specific}
\begin{center}
\begin{footnotesize}
\begin{tabular}{lllccc}
    \toprule
    & & & Word Idx  & Duration    & Land Pos \\
    Dataset      &Model  &$d_{emb}$                 & \multicolumn{1}{c}{NLL$\downarrow$}  & \multicolumn{1}{c}{MAE$\downarrow$}  & \multicolumn{1}{c}{MAE$\downarrow$}  \\\hline

  BSC &  Eyettention~\Romannum{2} &-   & 1.793\textsubscript{0.01}                  & 61.429\textsubscript{0.29}  & 0.208\textsubscript{0.00}\\
 (Chinese) & Eyettention~\Romannum{2}$_\text{reader}$  &16 & 1.674\textsubscript{0.01}$\dagger$ (6.6\%)                 & 58.739\textsubscript{0.28}$\dagger$ (4.4\%)   & \textbf{0.204}\textsubscript{0.00}$\dagger$ (1.9\%) \\
 &  Eyettention~\Romannum{2}$_\text{reader}$  &32 & \textbf{1.667}\textsubscript{0.01}$\dagger$ (7.0\%)                 & 58.257\textsubscript{0.28}$\dagger$ (5.2\%)   & \textbf{0.204}\textsubscript{0.00}$\dagger$ (1.9\%) \\
 &  Eyettention~\Romannum{2}$_\text{reader}$  &64 & 1.677\textsubscript{0.01}$\dagger$ (6.5\%)                 & \textbf{57.780}\textsubscript{0.28}$\dagger$ (5.9\%)   & 0.207\textsubscript{0.00}$\dagger$  (0.5\%) \\
 &  Eyettention~\Romannum{2}$_\text{reader}$  &128 & 1.683\textsubscript{0.01}$\dagger$ (6.1\%)                 & 58.112\textsubscript{0.28}$\dagger$ (5.4\%)   & 0.210\textsubscript{0.00}$\dagger$ (-1.0\%) \\

\hline
  
  CELER L1 &  Eyettention~\Romannum{2}  &- & 2.107\textsubscript{0.01}                 & 63.200\textsubscript{0.40}  & 0.200\textsubscript{0.00} \\
  
   (English) &  Eyettention~\Romannum{2}$_\text{reader}$  &16 &  1.998\textsubscript{0.01}$\dagger$ (5.2\%)                 &  60.894\textsubscript{0.38}$\dagger$  (3.6\%)  & \textbf{0.195}\textsubscript{0.00}$\dagger$ (2.5\%)\\

   &  Eyettention~\Romannum{2}$_\text{reader}$  &32 &  \textbf{1.989}\textsubscript{0.01}$\dagger$ (5.6\%)                 &  60.839\textsubscript{0.37}$\dagger$ (3.7\%)  & \textbf{0.195}\textsubscript{0.00}$\dagger$ (2.5\%)\\

    &  Eyettention~\Romannum{2}$_\text{reader}$  &64 &  
 1.991\textsubscript{0.01}$\dagger$ (5.5\%)                 &  \textbf{60.578}\textsubscript{0.37}$\dagger$ (4.1\%)   & 0.196\textsubscript{0.00}$\dagger$ (2.0\%)\\
     &  Eyettention~\Romannum{2}$_\text{reader}$  &128 &  2.004\textsubscript{0.01}$\dagger$ (4.9\%)                 &  60.745\textsubscript{0.38}$\dagger$ (3.9\%)   & 0.196\textsubscript{0.00}$\dagger$ (2.0\%)\\

    \botrule

    \end{tabular}
 \end{footnotesize}

\end{center}
\end{table*}

\subsection{Cross-Dataset Evaluation}
\label{sec_crossdata_eval}
In the cross-dataset evaluation, we assess Eyettention~\Romannum{2}’s ability to generalize to entirely different datasets that it was not exposed to during training.
Here, we train the model on the CELER~L1~\cite{berzak2022celer} dataset and test it on the ZuCo~NR~\cite{hollenstein2018zuco} dataset. These datasets differ substantially in experimental setups (e.g., eye tracking hardware, sampling rate, presentation style) and stimulus properties, with data collected in different labs across different countries.
Additionally, we explore whether a pretraining/fine-tuning approach could enhance model performance on small datasets, as is often the case in eye-tracking-while-reading applications.

\bmhead*{Results}
In Table~\ref{tab: crossdataset}, we compare the model trained on the CELER dataset and tested on the completely different dataset ZuCo with the model trained and tested both on ZuCo. We use 5-fold cross-validation, where 80\% of the ZuCo data is used for training and 20\% for testing in each fold. We ensure that the same ZuCo test data is used in each fold to evaluate both models for consistency. We observe a significant performance drop across all three fixation attributes. This indicates the challenge of generalizing to a different dataset when experimental setups and stimulus properties differ substantially. 
To address this challenge, we explore a \emph{pre-training} and \emph{fine-tuning} strategy where the model is first trained on 
the CELER dataset, then fine-tuned on 
80\% of the ZuCo data in each fold, followed by testing on the remaining 20\% of ZuCo test data. This strategy yields the best performance, suggesting that the pre-training and fine-tuning strategy can effectively enhance the model's generalization, especially when working with relatively small eye-tracking datasets. 
This strategy is particularly valuable for researchers aiming to simulate eye movements for a specific lab setup and stimulus design, but who can only collect a small amount of pilot data due to time or resource constraints. By leveraging available datasets recorded in different labs---even those with differing experimental setups and stimulus properties---researchers can pre-train the model on these external datasets, and then fine-tune it on a small sample recorded in their own target lab. Despite the limited size of the pilot data, our pre-training and fine-tuning strategy enables more accurate and setup-specific eye movement modeling than training on the target data alone.

\begin{figure}[t!]
  \begin{center}
  \includegraphics[width=\textwidth,keepaspectratio,trim={0.0cm 1cm 0.0cm 0.0cm}, clip]{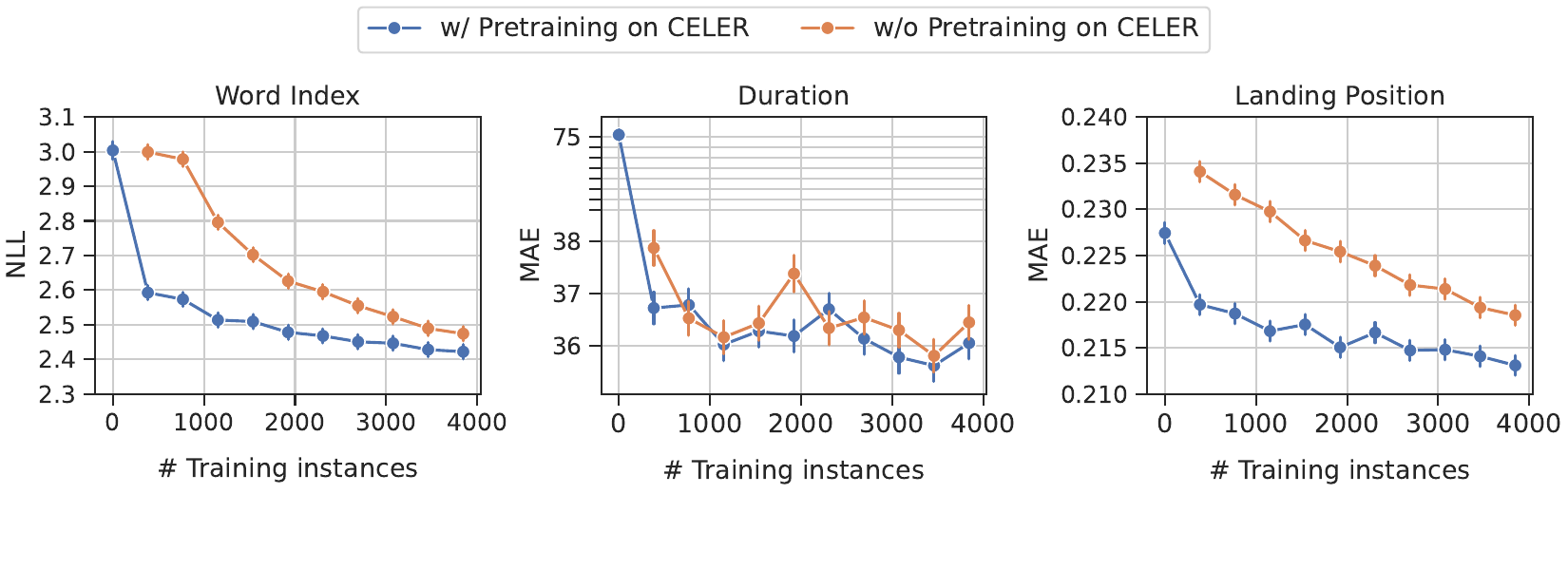}

    \caption{ Impact of the amount of data used for training (orange line, no pre-training) or fine-tuning (blue line, pre-training on CELER) on ZuCo, when testing on held-out data from ZuCo.}
    \label{fig: cross_dataset}
    \end{center}
  \end{figure}

To further illustrate this, in Figure~\ref{fig: cross_dataset} we present model performance as a function of the number of training instances from the ZuCo dataset when training it from scratch compared to the number of fine-tuning instances from ZuCo when the model has been pre-trained on CELER. Notably, with pre-training on CELER, the model achieves significant performance improvements with just a few hundred fine-tuning instances from ZuCo. In contrast, the model trained solely on ZuCo without pre-training requires thousands of instances to reach comparable performance, especially in word index and landing position prediction. This emphasizes the considerable benefits of employing a pre-training approach, particularly in the context of small datasets.


\begin{table*}[]
\caption{\textbf{Cross-Dataset Evaluation}. Exploring the possibility to generalize to completely different datasets. The standard error is indicated as the subscript.}
\label{tab: crossdataset}
\begin{center}
\begin{footnotesize}
\begin{tabular}{llllccc}
    \toprule
    &Training &Fine-tune &Testing &Word Idx  & Duration    & Land Pos \\
    Model &Data &Data &Data                   & \multicolumn{1}{c}{NLL$\downarrow$}  & \multicolumn{1}{c}{MAE$\downarrow$}  & \multicolumn{1}{c}{MAE$\downarrow$}  \\\midrule
    
    Eyettention~\Romannum{2} & ZuCo NR & -- & ZuCo NR & 2.474\textsubscript{0.02} & 36.441\textsubscript{0.33} & 0.219\textsubscript{0.00}\\
    

    Eyettention~\Romannum{2} & CELER L1 & -- & ZuCo NR &3.003\textsubscript{0.03}                 & 75.139\textsubscript{0.46}  & 0.227\textsubscript{0.00} \\


    Eyettention~\Romannum{2} & CELER L1 & ZuCo NR & ZuCo NR  & \textbf{2.422}\textsubscript{0.02}  &\textbf{36.046}\textsubscript{0.31}   & \textbf{ 0.213}\textsubscript{0.00} \\

    \botrule

    \end{tabular}
 \end{footnotesize}

\end{center}
\end{table*}

\subsection{Investigation of Model Behavior}
\label{sec_model_behavior}

\subsubsection{Psycholinguistic Analysis}
\label{sec_Psycho_analysis}
In this section, we assess Eyettention~\Romannum{2}'s capability to mirror human-like gaze behavior by investigating whether the model exhibits well-established psycholinguistic phenomena that characterize human reading behavior. Specifically, we will investigate whether the model exhibits effects of \textit{word length, lexical frequency, and surprisal}. 
From the fixation sequences, we therefore first compute reading measures commonly used in psycholinguistic studies. Definitions of these reading measures are presented in Table~\ref{tab:rm}.

\begin{table}[t!]
\caption{Reading measures used for psycholinguistic analyses. We follow the terminology of terminology of~\cite{wilcox2024mouse} and~\cite{jakobi2024potec}.}
\centering
\begin{footnotesize}
\begin{tabular}{p{0.3\textwidth}p{0.63\textwidth}}

\toprule
\textbf{Reading Measures} & \textbf{Description} \\ 
\midrule
First Fixation & Duration of the first fixation on a word. \\ 
Gaze Duration & The total time spent fixating on a word during the first pass,  before the reader either moves forward or makes a regression to an earlier word. \\ 
Go Past Time & Sum of all fixation durations from the first pass on a word until the reader progresses to a word to the right. \\ 
Total Duration & Total time spent fixating on a word, including re-reading. \\
\midrule
First Landing Position & The relative position within a word where the eye first lands during the initial fixation, normalized by the word's length (e.g., 0.0 and 1.0 represent the first and last character, respectively, and 0.5 represents the center of the word). \\ \midrule
First-pass Fixation Rate & Proportion of trials in which a word is fixated during the first pass. \\ 
First-pass Regression Rate & Proportion of trials in which a regression is initiated to earlier words during the first pass. \\ \midrule
Number of Fixations & Total number of fixations on a word. \\ 
\bottomrule
\end{tabular} 
\end{footnotesize}
\label{tab:rm}
\end{table}

For each reading measure, we then implement a Bayesian (generalized) linear mixed-effects model~\cite{bates2005fitting} with this reading measure as target variable and z-score normalized \textit{word length, lexical frequency, and surprisal} as predictors.  
 The models are defined as:

\begin{equation*}
\hat{y_{ij}} =
{\mathcal{D}}\left(
g^{-1}\left(
\beta_0 + b_{0i} + c_{0j} +
\sum_{m=1}^{M}\left(\beta_m + b_{mi} + c_{mj}\right)x_m
\right),
{\boldsymbol{\theta}_{\mathcal D}}
\right)
\end{equation*}

where $y_{ij}$ is the measured response variable and $\hat{y_{ij}}$ the corresponding predicted estimate for participant $i$ on item $j$. For continuous reading times, $y_{ij} \in \mathbb{R}$; for fixation and regression rates, $y_{ij} \in \{0, 1\}$; for First Landing Position, $y_{ij}$ is a value in $[0, 1]$; and for Number of Fixations, $y_{ij}$ is a count variable. $\beta_0$ is the global intercept, $b_{0i}$ and $c_{0j}$ represents random intercepts for participants and item, $x_m$ are the $M=3$ psycholinguistic predictor variables, with coefficients $\beta_{m}$ and random slopes  for participants $b_{mi}$ and random slopes  for items $c_{mj}$ 
\footnote{By-subject Random effects are only included when modeling human data, as participant information is not available for computational models.}. 
Focusing on the likelihood family distribution $\mathcal{D}(\cdot)$, continuous reading times are modeled using a lognormal distribution, binary fixation and regression rates using a Bernoulli distribution, first landing position using a Beta distribution, and fixation counts using a zero-inflated Poisson distribution. The link function $g(\cdot)$ is the identity function for the lognormal model (on the log scale), the logit function for Bernoulli and Beta models, and the log function for the Poisson component of the zero-inflated Poisson model. The vector $\boldsymbol{\theta}_{\mathcal D}$ denotes family-specific parameters: a scale parameter $\sigma$ for the lognormal distribution, a precision parameter $\phi$ for the Beta distribution, no additional parameter for the Bernoulli distribution, and an additional zero-inflation probability for the zero-inflated Poisson model.

The models are fit in \textit{R}~\cite{rteam} \cite{rteam} using the \textit{brms} package~\cite{brms2021}. Each model is fit using four parallel chains of 4,000 iterations, with the first 2,000 iterations used for warm-up. 
Mildly informative priors, inspired by~\cite{nicenboim2021introduction}, 
are applied to ensure estimation stability. Details of the priors are provided in Table~\ref{tab:priors}.

\begin{table}[t!]
\caption{\textbf{Priors for model parameters:} $\beta_0$ and $\beta_m$ represent the coefficients of the fixed effects (intercept and psycholinguistic predictors, respectively), $b$ denotes random effects, and $sd(b)$, $sd(c)$ denotes the standard deviation of random effects. $\Omega$ is the correlation matrix for random effects, $\sigma$ is the residual standard deviation for continuous outcomes, and $\phi$ is the precision parameter for Beta-distributed proportions.}
\centering
\renewcommand{\arraystretch}{1.1} 
\setlength{\tabcolsep}{3pt} 
\begin{footnotesize}
\begin{tabular}{llll}
\toprule
\textbf{RTs} & \textbf{First Landing Pos.} & \textbf{Fix./Regr. Rate} & \textbf{Num. Fixations} \\ 
\midrule
$\beta_0 \sim \mathrm{Normal}(6, 1)$ & $\beta_0 \sim \mathrm{Normal}(0, 1)$ & $\beta_0 \sim \mathrm{Normal}(0, 1)$ & $\beta_0 \sim \mathrm{Normal}(0, 1)$ \\

$\beta_m \sim \mathrm{Normal}(0, 1)$ & $\beta_m \sim \mathrm{Normal}(0, 1)$ & $\beta_m \sim \mathrm{Normal}(0, 1)$ & $\beta_m \sim \mathrm{Normal}(0, 1)$ \\

$sd(b)\sim \mathrm{Exponential}(2)$ & $sd(b) \sim \mathrm{Exponential}(2)$ & $sd(b)\sim \mathrm{Exponential}(2)$ & $sd(b) \sim \mathrm{Exponential}(2)$\\

$sd(c)\sim \mathrm{Exponential}(2)$ & $sd(c) \sim \mathrm{Exponential}(2)$ & $sd(c)\sim \mathrm{Exponential}(2)$ & $sd(c) \sim \mathrm{Exponential}(2)$\\

$\Omega \sim \mathrm{LKJ}(2)$ & $\Omega \sim \mathrm{LKJ}(2)$ & $\Omega \sim \mathrm{LKJ}(2)$ & $\Omega \sim \mathrm{LKJ}(2)$\\

$\sigma \sim \mathrm{Exponential}(2)$ & $\phi \sim \mathrm{gamma}(2, 1)$ & --- & --- \\
\bottomrule
\end{tabular}
\end{footnotesize}
\label{tab:priors} 
\end{table}

\paragraph{Results}
\label{sec_Psycho_analysis_results}

\begin{figure}[t!]
    \centering
    \begin{minipage}[c]{\textwidth}
    \includegraphics[width=0.93\textwidth]{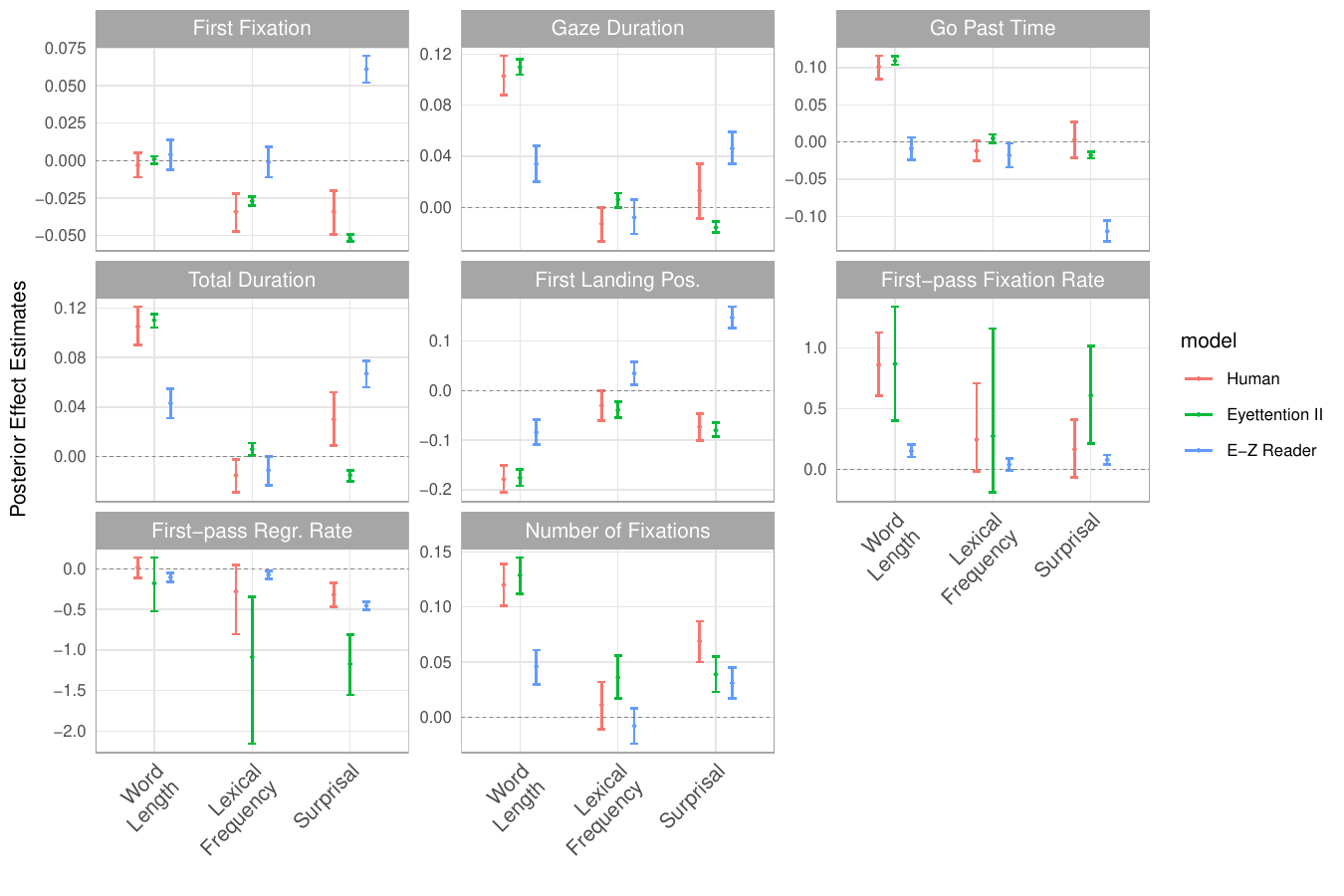}
    \vspace{-10pt}
        \subcaption{BSC dataset (Chinese)} \label{fig:zh_psy}
    \end{minipage}
    
    \vspace{-5pt} 
    \begin{minipage}[c]{\textwidth}
        \includegraphics[width=0.93\textwidth]{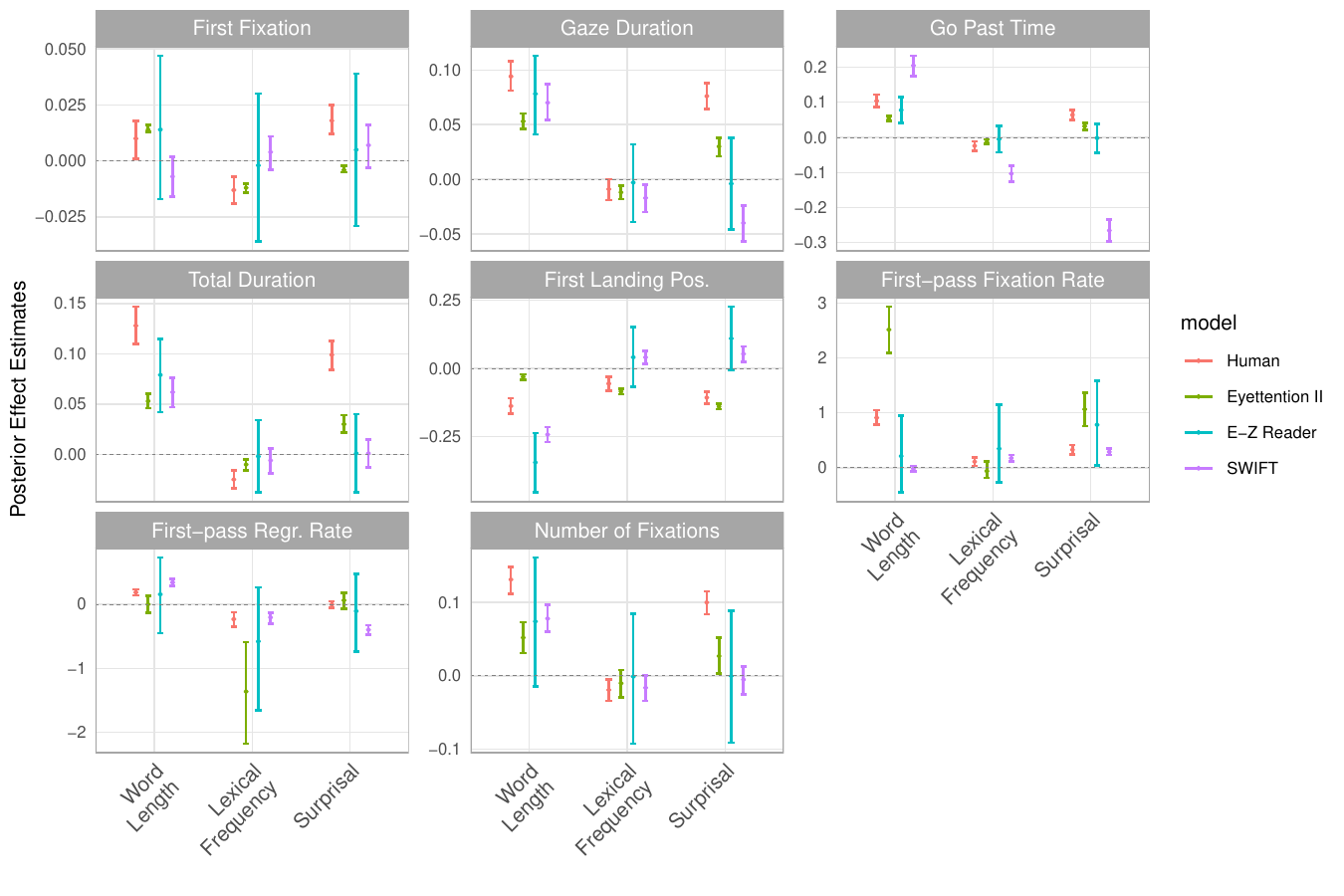}
        \vspace{-10pt}
        \subcaption{CELER L1 dataset (English)} \label{fig:en_psy}
    \end{minipage}
    \caption{\small \textbf{Comparison of posterior effect estimates across human readers, models and languages:} Posterior means and $95$\% credible intervals for psycholinguistic predictors in human eye-tracking data and model-predicted scanpaths. Horizontal dashed lines indicate zero effect size.}
    \label{fig:psy-effects}
    \setlength{\belowcaptionskip}{-30pt}
\end{figure}

 Figure~\ref{fig:psy-effects} displays posterior coefficient estimates for \textit{word length, lexical frequency, and surprisal} across eight eye-tracking reading measures from human participants and computational models (Eyettention~\Romannum{2}, E-Z Reader, SWIFT), separated by language: (a) for the Chinese BSC dataset, and (b) for the English CELER dataset.
In both the Chinese BSC and English CELER datasets, human reading behavior follows established psycholinguistic effects. Longer words lead to prolonged fixations and more regressions, especially in Total Duration, Gaze Duration, and Go Past Time. More frequent words receive shorter fixations and fewer regressions, with the strongest effects in reading time measures. Higher surprisal words increase fixation durations and regressions, particularly in later reading time measures, such as Go Past Time and Total Duration. First Fixation Duration and First-pass Regression Rate show weaker surprisal effects, suggesting surprisal disrupts processing later.

For the computational models, in the Chinese BSC dataset (Figure~\ref{fig:zh_psy}), Eyettention~\Romannum{2} demonstrates strong alignment with human patterns for all three psycholinguistic predictors across most reading measures. However, it struggles with the frequency effect in First-pass Regression Rate and the surprisal effect in First-pass Fixation Rate,  where its estimates show high uncertainty, reflected in a much wider 95\% Credible Interval. This increased uncertainty may stem from the fact that there are fewer training examples with regressions or skips compared to the reading time measures. The cognitive model E-Z Reader tends to underestimate the word length effect across most measures
and fails to capture the surprisal effect in reading time measures.

For the English CELER dataset (Figure~\ref{fig:en_psy}), Eyettention~\Romannum{2} closely replicates human patterns across most measures. 
Notably, it outperforms most established cognitive models such as SWIFT and E-Z Reader in most, but not all, measures.  While E-Z Reader exhibits greater uncertainty in its estimates (wider 95\% Credible Interval), SWIFT deviates more from human patterns in most of its estimated effect sizes.


Overall, these findings highlight Eyettention~\Romannum{2}'s strength in aligning with the psycholinguistic phenomena observed in humans across both English and Chinese datasets compared to traditional models. It consistently produced effect size estimates and uncertainty levels that are more closely aligned with human data. For the English dataset, Eyettention~\Romannum{2} aligns closely with human reading behavior and effectively exhibits phenomena that are key to psycholinguistic theories of human language processing, although---unlike cognitive models---it is not explicitly designed to account for these phenomena. However, the results also suggest areas for improvement, particularly in refining Eyettention~\Romannum{2}'s performance on binary measures like regression probabilities, to further enhance its generalizability. Its performance on the Chinese dataset could be further refined to match the stronger performance observed in English.

\subsubsection{Decoding Strategies for Scanpath Generation}
\label{sec_decode_strategy}

\begin{figure}[t!]

    \subfloat[BSC dataset.\label{subfig:sampling_strategy_BSC}]{%
  \includegraphics[width=0.45\textwidth,keepaspectratio,trim={0.2cm 0 0.2cm 0}, clip]{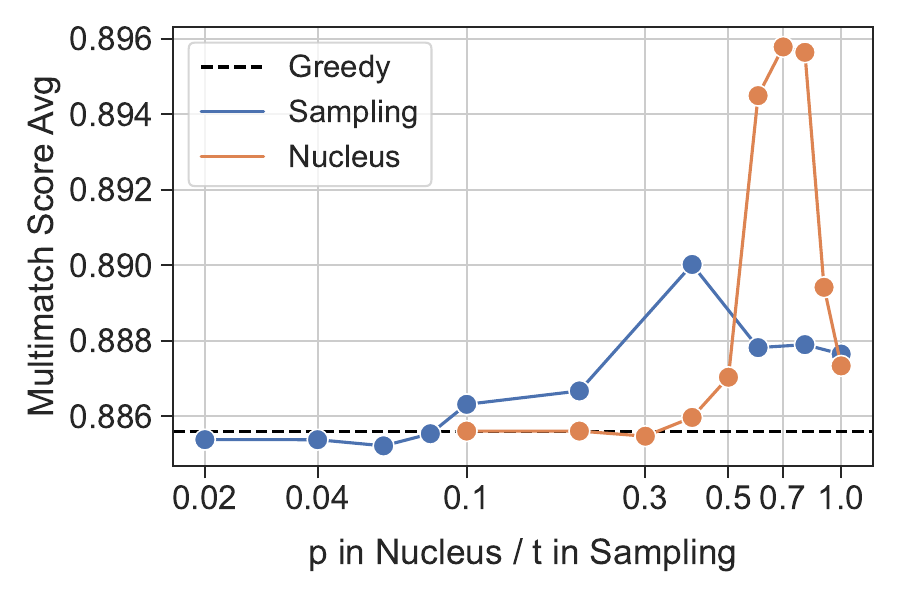}
    }
    \hfill
    \subfloat[CELER L1 dataset.\label{subfig:sampling_strategy_CELER}]{%
      \includegraphics[width=0.45\textwidth,keepaspectratio,trim={0.2cm 0 0.2cm 0}, clip]{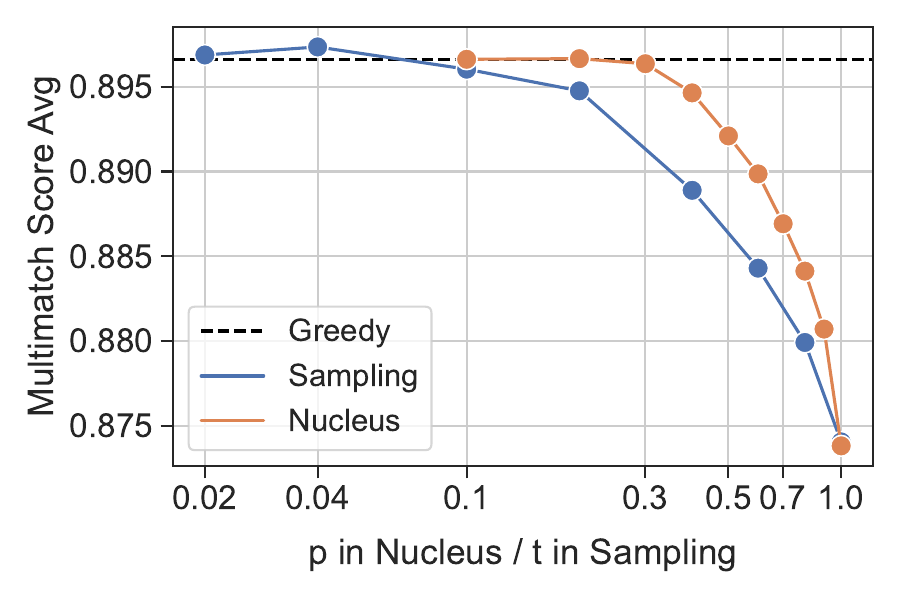}
    }
    
    \caption{Comparison of scanpaths generated by Eyettention~\Romannum{2} using different decoding strategies. Higher Multimatch scores indicate greater similarity between the generated scanpaths and observed human scanpaths.}
    \label{fig: sampling_strategy}

  \end{figure}

The Eyettention~\Romannum{2} model generates fixation locations in a probabilistic manner. Specifically, the Word Index Decoder (see Figure \ref{fig:Eyettention}) produces a conditional probability distribution over all possible fixation locations, given the stimulus sentence and the preceding fixations. To determine the next fixation, the model does not simply select the location with the highest probability; instead, it samples from the distribution generated by the Word Index Decoder, resulting in inherently non-deterministic behavior. The procedure used to perform this sampling is referred to as \textit{decoding strategy} and can take different forms, each reflecting a particular trade-off between selecting highly probable fixation locations and allowing for greater variability in the generated scanpaths.

Technically, the word index for each fixation is sampled from a probabilistic distribution using a pre-defined so-called \textit{decoding strategy}~(\textsection~\ref{sec_wordidx}). 
To determine the optimal decoding strategy for Eyettention~\Romannum{2}, we compare three approaches: (\romannum{1}) \textbf{Greedy decoding} selects the next fixated word with the highest likelihood, yielding a deterministic scanpath for a given text. (\romannum{2}) \textbf{Temperature sampling}~\cite{ackley1985learning} adjusts the probability distribution based on a `temperature' hyperparameter $t$. Lower temperatures emphasize high-probability events, producing more likely scanpaths, while higher temperatures introduce variability. (\romannum{3}) \textbf{Nucleus sampling}~\cite{holtzmancurious} samples from the smallest set of top-ranked events that cover a cumulative probability threshold $p$. 

Results for the BSC dataset are shown in Figure~\ref{subfig:sampling_strategy_BSC}. Nucleus sampling with $p=0.7$ achieves the best performance, effectively improving scanpath quality by truncating the less reliable tail of the probability distribution.  
For the CELER dataset, as shown in Figure~\ref{subfig:sampling_strategy_CELER}, we observe comparable good performance across the greedy method, low-temperature sampling, and nucleus sampling with small $p$ values, with temperature sampling at $t=0.04$ performing best. Due to the variability in CELER’s training data, which includes longer scanpaths with frequent refixations, regressions, and long-distance saccades, the model learns a more uniform probability distribution for word index prediction. Here, lowering the temperature or nucleus threshold helps reduce randomness, prioritize high-probability events, and ultimately produce high-quality scanpaths.

\subsection{Power Analysis}
\label{sec_power_analysis}
We want to ask whether Eyettention~\Romannum{2} could serve as a reliable tool for experiment planning, particularly in scenarios where human pilot data are scarce or difficult to collect, or no prior literature exists to obtain effect estimates for a certain psycholinguistic manipulation on a given set of stimulus materials. Given its demonstrated ability to mirror human-like gaze behavior---closely imitating the way humans move their eyes while reading---and its alignment with human eye-tracking data with respect to well-established psycholinguistic phenomena, we explore whether Eyettention~\Romannum{2} can be used to simulate data to conduct a statistical power analysis for a given set of stimulus materials. Specifically, we present results for the New Sentence\,/\,New Reader split. This scenario involves conducting a power analysis on new materials, read by a new sample readers that the model has never encountered before. We assume that the lab setups (e.g., eye tracking hardware, sampling rate, presentation style) are similar to the one used to record the model's training data.

\begin{figure}[t!]
    \centering
    \begin{minipage}[c]{\textwidth}
    \centering
    \includegraphics[width=0.85\textwidth]{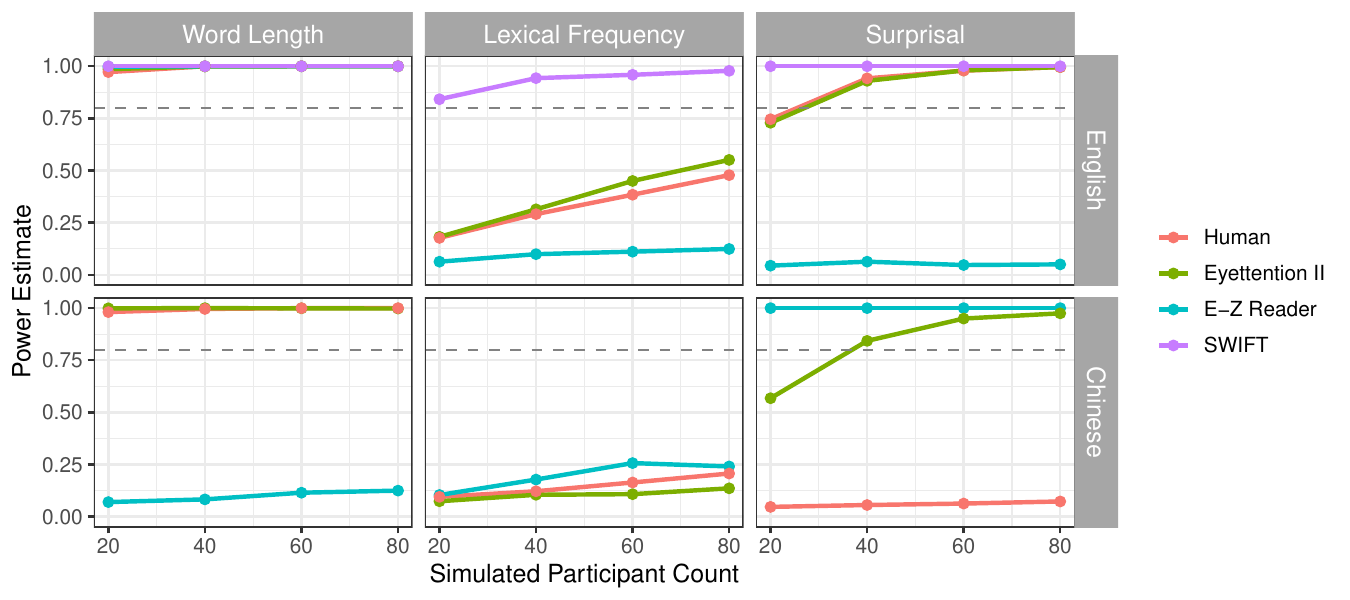}
    \vspace{-7pt}
        \subcaption{Go Past Time.} \label{fig:go_past_time}
    \end{minipage}
    
    \vspace{-5pt} 
    \begin{minipage}[c]{\textwidth}
    \centering
        \includegraphics[width=0.85\textwidth]{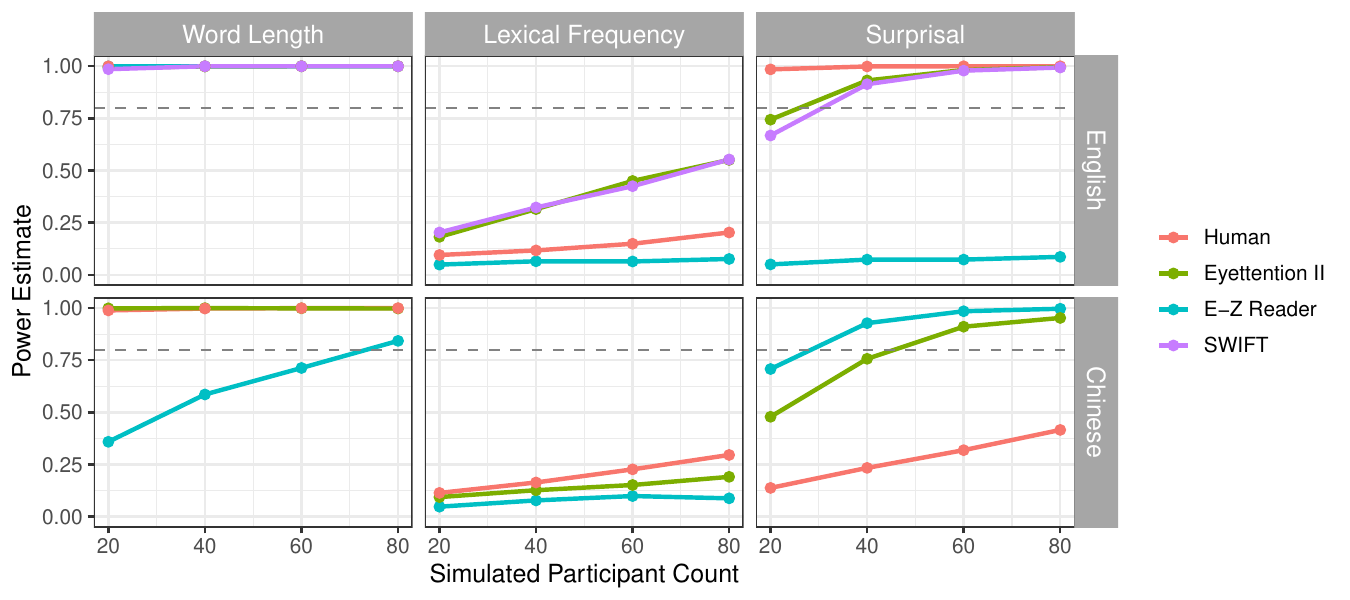}
        \vspace{-7pt}
        \subcaption{Gaze Duration.} \label{fig:gaze_duration}
    \end{minipage}

    \vspace{-5pt} 
    \begin{minipage}[c]{\textwidth}
    \centering
        \includegraphics[width=0.85\textwidth]{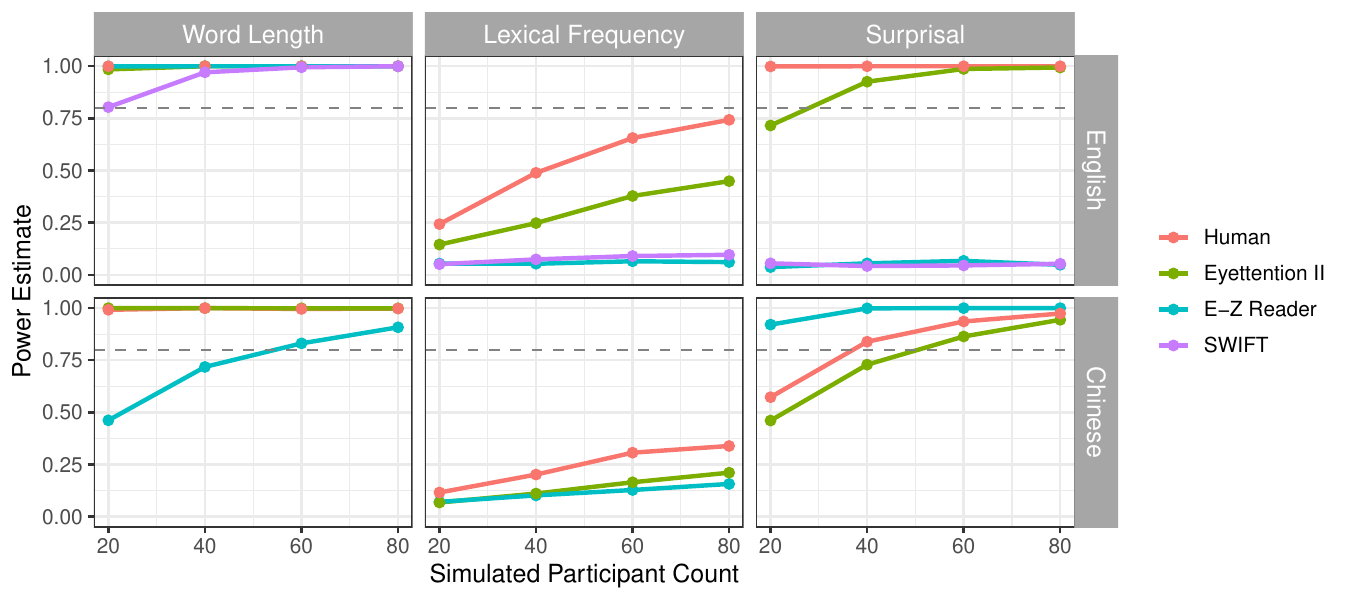}
        \vspace{-7pt}
        \subcaption{Total Duration.} \label{fig:total_duration}
    \end{minipage}

    \caption{\small{\textbf{Comparison of simulated power across models and predictors:} the predicted statistical power for various participant counts (20, 40, 60, 80) across three predictors: word length, lexical frequency, and surprisal. Results are simulated based on human data and data generated by computational models (Eyettention~\Romannum{2}, E-Z Reader, and SWIFT) for (a) Go Past Time, (b) Gaze Duration, and (c) Total Duration. 
    The dashed line indicates a power of 0.8.\looseness=-1}}
    \label{fig:power-analysis}
    \setlength{\belowcaptionskip}{-30pt}
\end{figure}

We conduct an exemplary power analysis for three independent variables of interest: \textit{word length, lexical frequency, and surprisal}, and three dependent variables: \textit{Go Past Time}, \textit{Gaze Duration}, \textit{Total Duration}. We compare the power estimates obtained with human data against the results obtained from data simulated with Eyettention~\Romannum{2} and two reference methods,  E-Z Reade and SWIFT, and includes data from English CELER and Chinese BSC datasets. 
For each language, model, and predictor, we simulate 1,000 datasets with a fixed item count of 50 and varying participant counts (20, 40, 60, and 80). The simulations are based on parameter estimates (intercepts, predictor effect sizes, and residual standard deviations) obtained from models. 
Simulated responses are generated using a linear model with fixed effects for predictors and a residual term. A linear regression model is then applied to each simulated dataset, and statistical power is calculated as the proportion of simulations where the p-value for a given predictor is below 0.05. The final power estimates, illustrating the sensitivity of different models to experimental setups and psycholinguistic phenomena, are presented in Figure~\ref{fig:power-analysis}.

\paragraph{Results}
\label{power_analysis_results}

Eyettention~\Romannum{2} closely aligns with human power estimates across all predictors in English, making it an effective tool for power estimation and experimental planning. It follows a power change pattern similar to human data, reaching over 0.8 power with 80 participants for word length and surprisal effects and lower estimated power for lexical frequency effect, between 0.4 and 0.7 for later reading time measures. In contrast, SWIFT consistently overestimates power for lexical frequency and surprisal, reporting near-perfect power even at small sample sizes. Conversely, E-Z Reader underestimates power for these predictors, often below 0.2 regardless of sample size, indicating its limitations in experimental planning.

In the Chinese BSC dataset, Eyettention~\Romannum{2} closely matches human power estimates for word length and lexical frequency but overestimates power for the surprisal effect in Go Past Time and Gaze Duration, reaching nearly 1 with 80 participants, whereas human data remain around 0.5. E-Z Reader exhibits even greater overestimation for surprisal, particularly at smaller sample sizes, and it consistently underestimates the power for word length and lexical frequency across the simulated participant counts. These findings indicate that while Eyettention~\Romannum{2} is highly effective for English experiment planning and outperforms other models for Chinese experiment planning, further refinements are necessary to enhance its accuracy for Chinese psycholinguistic phenomena.

\subsection{Inspection of LLMs in Scanpath Prediction}
\label{sec_LLM}

\subsubsection{Effects of Large-Scale LMs on Scanpath Prediction}
\label{sec_larger_LMs}

\begin{figure}[t!]
    \begin{center}
        
  \subfloat[BSC dataset.\label{subfig:diff_LM_BSC}]{
  \includegraphics[width=\textwidth,keepaspectratio,trim={0.0cm 1.4cm 0.0cm 0}, clip]{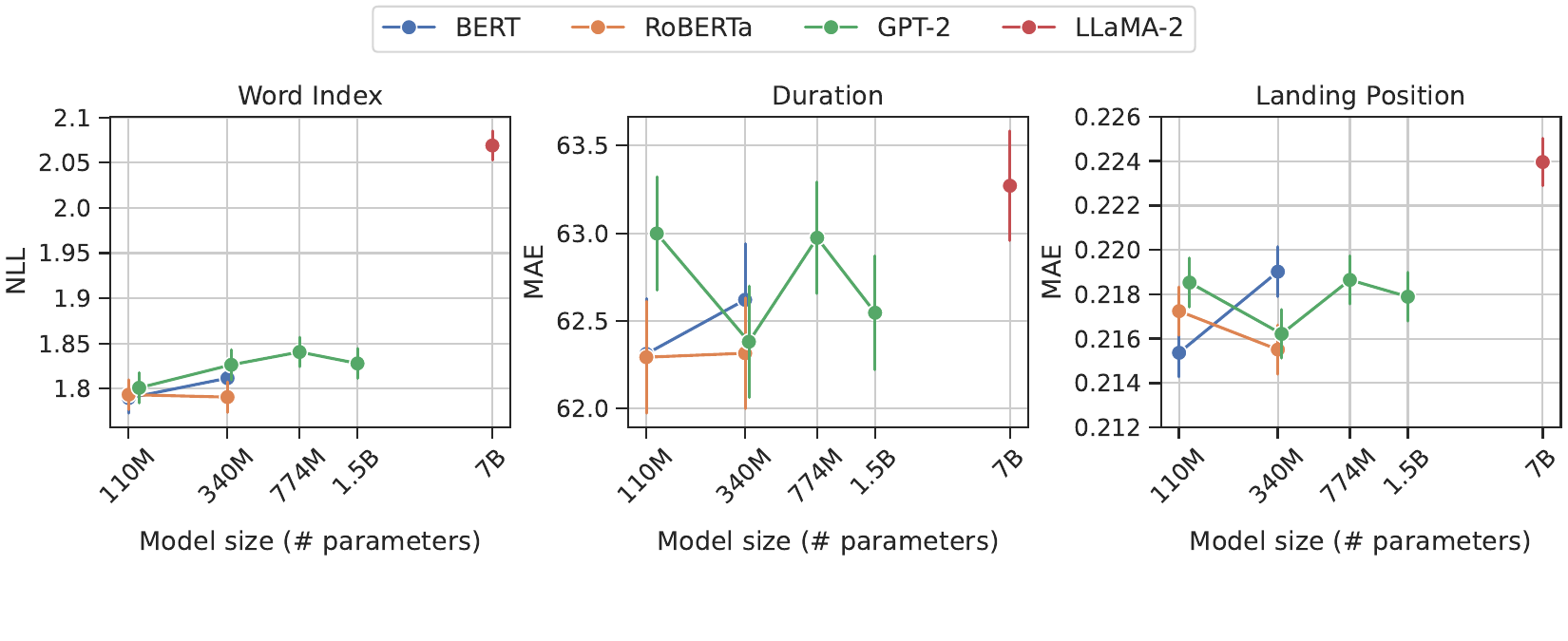}
  }\\
  \subfloat[CELER L1 dataset.\label{subfig:diff_LM_CELER}]{\includegraphics[width=\textwidth,keepaspectratio,trim={0.0cm 1.4cm 0.0cm 1.4cm}, clip]{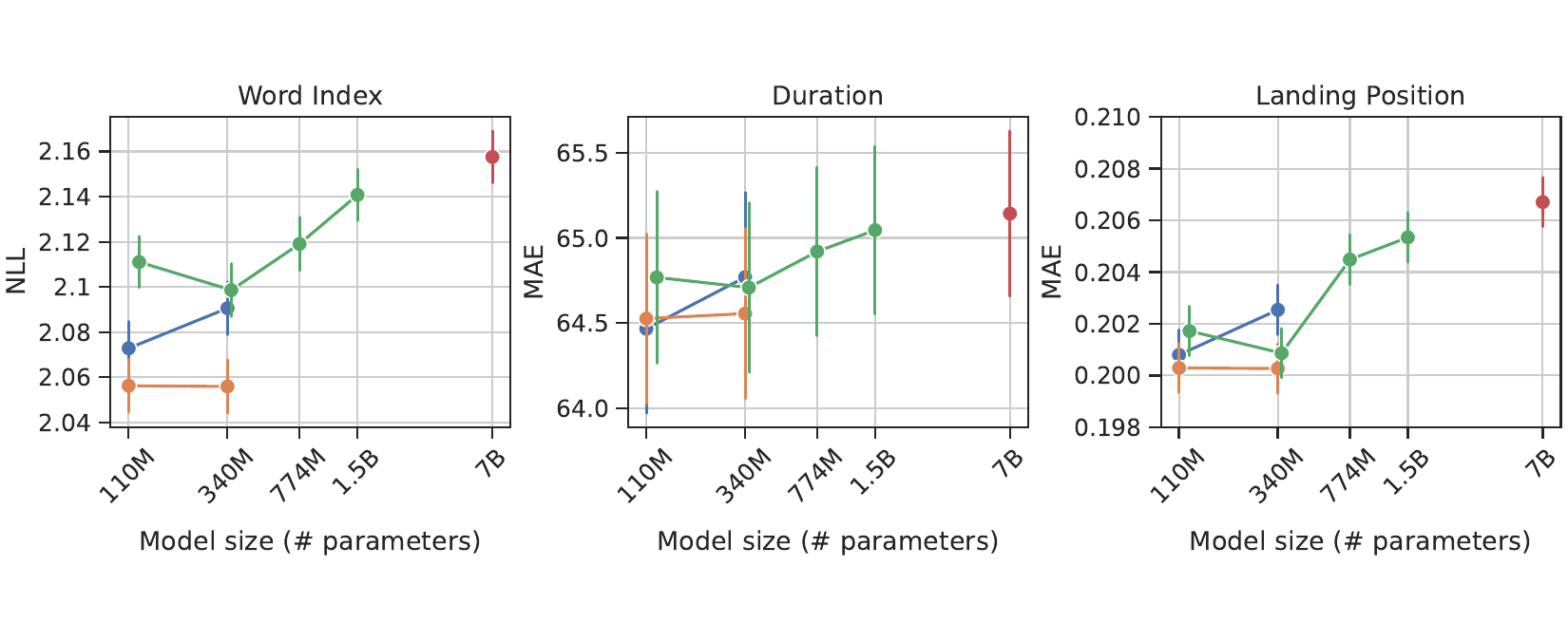}
  }
   
    \caption{Comparison of different LM families and their variants by size for scanpath prediction.}
    \label{fig: large_LM}
\end{center}
  \end{figure}
  
Scaling laws for neural LMs suggest that both test loss and downstream task performance improve monotonically with increasing model size~\cite{kaplan2020scaling}. To investigate whether larger-scale LMs provide linguistic features that improve scanpath predictions, we evaluate the performance of Eyettention~\Romannum{2} with different pre-trained Transformer-based LMs in the Word-Sequence Encoder. This includes two variants of both BERT~\cite{devlin2018bert} and RoBERTa~\cite{liu2019roberta}, four variants of GPT-2~\cite{radford2019language}, and LLaMA-2~\cite{touvron2023llama}, with the variants differing solely in model size. The size of these models ranges from 110 million to 7 billion parameters. We report results for the New Sentence\,/\,New Reader split in Figure~\ref{fig: large_LM}.

The main finding is that the smaller variants of LMs within each family, such as BERT 110 million and GPT-2 340 million, demonstrate better performance in predicting human scanpaths compared to their larger counterparts.
This suggests that increasing model size does not necessarily improve the ability to model human reading behavior.  
Additionally, when comparing GPT-2, BERT, and RoBERTa models with similar capacities (110 and 340 million), GPT-2 generally underperforms. This discrepancy may arise from the architectural differences, as BERT and RoBERTa utilize a bidirectional architecture that facilitates the learning of contextual features from both directions, while GPT relies on a unidirectional design. 

\subsubsection{Effect of Different Levels of LM Representations on Scanpath Prediction}

\label{sec_diff_LMrep}

\begin{figure}[t!]

 \subfloat[BSC dataset.\label{subfig:diff_LM_layer_BSC}]{%
  \includegraphics[width=\textwidth,keepaspectratio, trim={0cm 0.4cm 0cm 0.4cm}, clip]{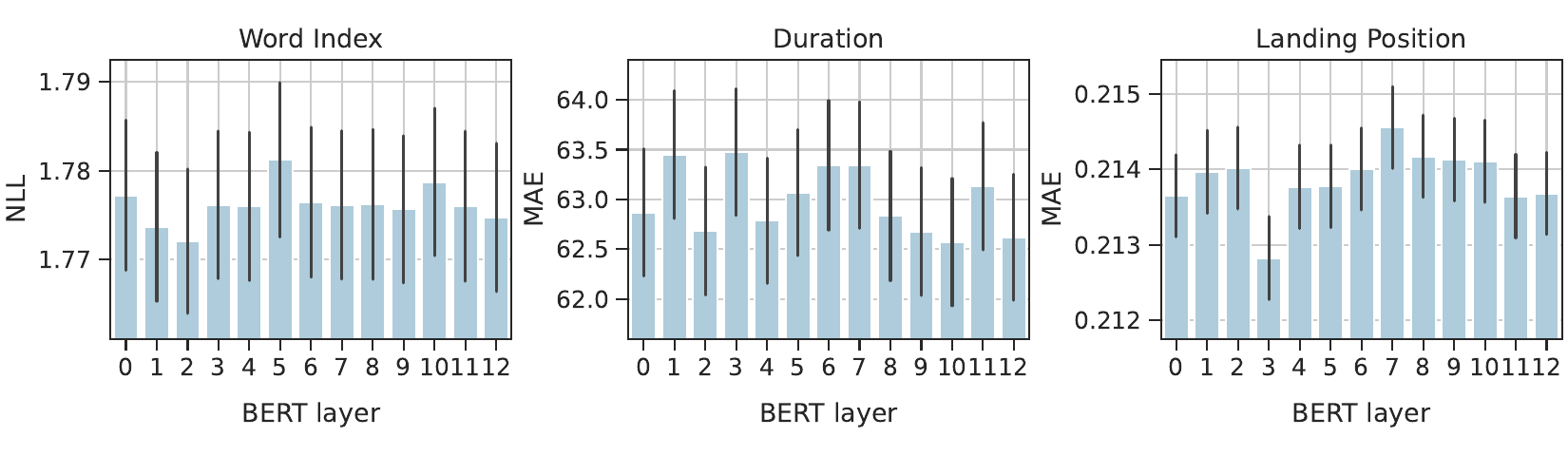}
    }

    \subfloat[ CELER L1 dataset.\label{subfig:crossdataset_wordindex}]{%
  \includegraphics[width=\textwidth,keepaspectratio, trim={0cm 0.4cm 0cm 0.4cm}, clip]{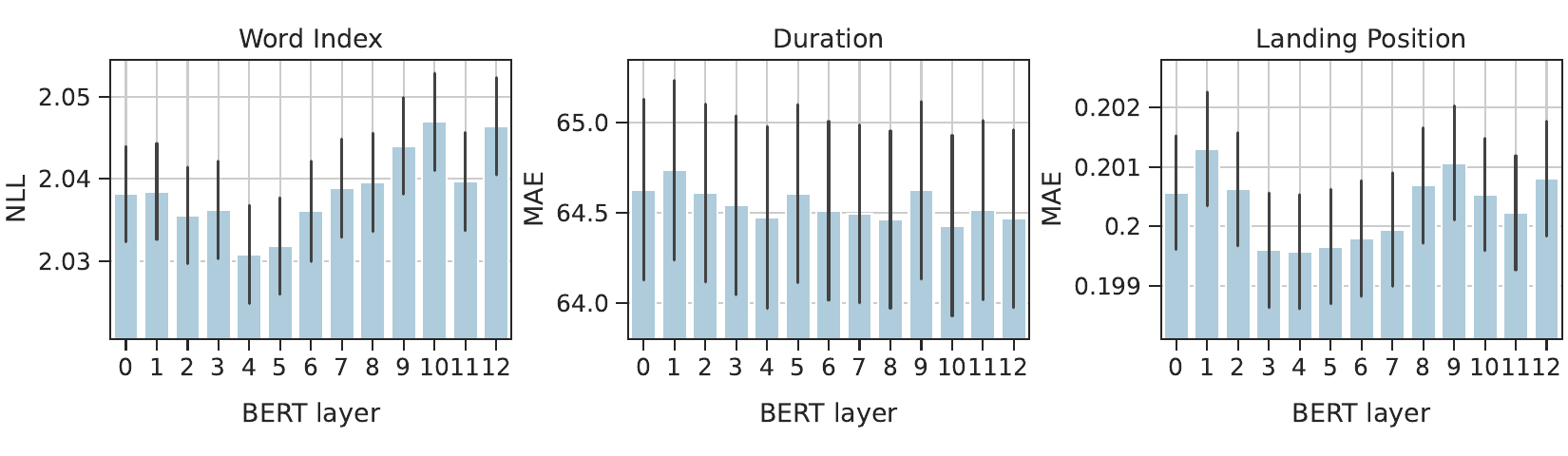}
    }

    \caption{Impact of different BERT layer representations on scanpath prediction.}
    \label{fig: sec_diff_LMemb}

  \end{figure}
Pre-trained LM embeddings have shown promise in providing linguistic features that support human scanpath prediction during reading. Different layers in these models capture distinct types of linguistic information: BERT’s intermediate layers encode a rich hierarchy of linguistic information, with surface features at the bottom,
syntactic features in the middle, and semantic features at the top~\cite{jawahar2019does}. We therefore examine how representations from different BERT layers contribute to scanpath prediction.
Figure~\ref{fig: sec_diff_LMemb} shows the variation in model performance across three fixation attributes for the BSC and CELER datasets. For word index prediction, the lower 2nd layer performs best on BSC, while the 4th layer is most effective on CELER.  Duration prediction, however, benefits most from upper layers (10th and 12th) on both datasets. Landing position prediction consistently favors the 3rd layer. These results suggest that phrase-level and surface-level information from lower layers are essential for predicting spatial aspects of scanpaths (word index and landing position), whereas high-level semantic features from upper layers are more influential for temporal aspects (fixation duration).

\section{General Discussion}
\label{sec_discussion}
\bmhead*{Bidirectional context in word representations.}
One limitation of the model’s autoregressive design is that the word embeddings produced by the Word-Sequence Encoder rely on a bidirectional LM and therefore incorporate both left and right context. A more human-like approach would compute each word embedding based only on the words that have already been fixated. However, even this is not necessarily the most cognitively plausible option, as it does not account for shallow processing or for degradation of word representations in memory due to interference or decay. Ongoing work in computational psycholinguistics is investigating which context size and weighting in word embeddings best reflect human processing; see, e.g., \cite{kuribayashi-etal-2022-context, mccurdy-hahn-2024-lossy, hennertcan}. 
In any case, computing cognitively more plausible embeddings would require a different context for every fixation---depending on whether it is a first-pass or later-pass fixation, and including partially masked contexts in cases of skips---which would introduce substantial computational overhead. For simplicity, we therefore use a standard bidirectional model for computing the embeddings. Importantly, the LM in the Word-Sequence Encoder can be replaced with any other type of LM, including a purely autoregressive one, if a different context-conditioning behavior is desired.

\bmhead*{Transfer learning in scanpath prediction.} 
In cross-dataset evaluations (\textsection~\ref{sec_crossdata_eval}), we explore a pre-training and fine-tuning strategy that demonstrates notable advantages, especially when the target dataset is small. Without pre-training, limited target data alone yields less satisfactory results compared to training on a large amount of data, as shown in Figure~\ref{fig: cross_dataset}, highlighting the challenge of generalizing effectively from small eye-tracking datasets. However, we show that pre-training the model on existing eye-tracking-while-reading data from the same language---despite being collected in different labs with different experimental setups and stimulus properties---enables the model to learn general eye movement patterns. As a result, only a small amount of fine-tuning data recorded under the target setup is needed for adaptation, allowing the model to achieve performance comparable to training from scratch on a large target dataset. Beyond cross-dataset transfer, this strategy has broader applications. For instance, a model pre-trained on natural reading scanpath data could provide a robust starting point for fine-tuning in task-specific scanpath generation, especially with limited target data.

\bmhead*{Larger LMs are less predictive of human reading behavior.}

An important finding presented in Section~\ref{sec_larger_LMs}, reveals that the smallest transformer-based LMs, namely BERT and RoBERTa 110 million, generate linguistic features that are superior predictors of both the temporal and spatial aspects of eye movements compared to larger LMs.  This observation is consistent with prior research showing that larger transformer-based LMs yield surprisal estimates that are less predictive of human reading times~\cite{oh2022comparison,oh2023does,de2023scaling}. Our work extends these findings by offering a more comprehensive investigation into how the rich linguistic features (presumably indirectly also encoding predictability estimates) provided by different LMs influence various aspects of eye movement prediction, including both durational (reading times) and spatial (fixation locations) aspects. This contributes to a deeper understanding of the correlation between different LM sizes and the prediction of human reading behavior. The insight that ``larger is not better" when modeling human reading behavior implies a need for caution when using pre-trained language models to study human language processing.

\bmhead*{Low-level vs high-level linguistic features in scanpath prediction.} 
While earlier cognitive models of oculomotor control in reading such as SWIFT and EZ-Reader focus on low-level text properties, such as word frequency and word length, as key factors influencing scanpath modeling, more recent approaches integrate higher level factors such as syntactic processing~\cite{veldre2020towards,rabe2024seam}. While these cognitive models can be considered computational implementations of theories of reading, machine-learning models are purely data-driven in a theory-agnostic way. They increasingly use LMs that encode rich linguistic features, showcasing their effectiveness in predicting eye movements during reading. In Section~\ref{sec_diff_LMrep}, we investigate the effectiveness of features extracted from different layers of the BERT model. Lower layers demonstrate greater predictive power for fixation position prediction by effectively capturing phrase-level information and surface features. For example, readers are more likely to skip frequent, shorter words while fixating on longer or syntactically complex phrases as they attempt to parse sentences. On the other hand, fixation duration prediction depends more heavily on higher-level semantic features from upper layers. Readers tend to fixate longer on words or phrases that demand greater cognitive effort, especially in cases of semantic ambiguity or complexity. Moreover, semantic integration at the sentence or paragraph level can prolong fixation duration as the reader tries to fully comprehend the material. These findings suggest that both lower-level and higher-level features are crucial for accurately modeling eye movement behavior, with their contributions varying depending on the specific aspect of eye movement being predicted.

\bmhead*{Trade-off between predicting average vs diverse scanpaths.}

In Section~\ref{sec_decode_strategy}, we investigate different decoding strategies for scanpath generation. Our findings reveal that adjusting the temperature parameter $t$ or nucleus threshold $p$ enables a flexible trade-off between generating more probable, ``average'' scanpaths and producing more diverse scanpaths. 
This adaptability is highly advantageous for different applications. For instance, in psycholinguistic experiments, researchers might prioritize simulating diverse scanpaths to represent variability across readers. Conversely, in NLP tasks, generating average eye movement patterns could be more useful for interpreting neural language models or for using the eye gaze as feedback in post-training approaches (e.g., when deriving a reward signal from a simulated reader's eye movement pattern for Reinforcement Learning with Human Feedback (RLHF) or related methods. 
Additionally, we observe that inter-reader similarity scores are lower than the similarity between human data and our model-generated scanpaths. By calibrating the sampling or nucleus parameters, it is possible to align generated scanpath variability with human similarity scores. However, caution is needed to avoid generating incoherent scanpaths if parameters are set too high or too low.

\bmhead*{Effects of reduced scanpath variability on power estimation}
\label{power_analysis_limitations}
A potential limitation of the proposed power estimation approach (Section~\ref{sec_power_analysis}) concerns the variance of model-generated reading measures. Eyettention~\Romannum{2} tends to produce scanpaths with lower variability than observed in human data, which may lead to overly optimistic power estimates, particularly for predictors such as lexical frequency and surprisal. Since statistical power is highly sensitive to the variance components, even accurate estimates of mean effects can yield inflated power when variability is underestimated. Several strategies could mitigate this limitation in future work. One approach is to calibrate the variance of simulated reading measures using empirical estimates from real eye-tracking datasets, rather than relying solely on model estimates. Furthermore, decoding strategies (Section~\ref{sec_decode_strategy})
that explicitly encourage variability may help address variance underestimation and thereby improve the realism of power predictions.

\bmhead*{Scalability} We develop a lightweight model that can be efficiently trained on limited GPU resources, making it accessible to a broader range of research communities. Its design also minimizes the risk of overfitting to typically small eye movement datasets. The model has the potential to be scaled up, for example by replacing the LSTM layers with self-attention blocks. While such an enhanced model could achieve new state-of-the-art performance, it would require substantial computational resources and more extensive eye-tracking data for training.

\section{Conclusion}
\label{sec_conclusion}
In this work, we introduce Eyettention~\Romannum{2}, a
lightweight scanpath generation model capable of predicting complete fixation attributes---word index, within-word landing position, and duration---in a sequential scanpath given a text stimulus. Additionally, we develop Eyettention~\Romannum{2}$_\text{reader}$, an extension designed to simulate reader-specific scanpaths. Our experiments demonstrate that Eyettention~\Romannum{2} achieves state-of-the-art performance and exhibits human-like reading behavior that replicate key statistical and behavioral properties of real human eye movements, making it a versatile tool for diverse applications, such as piloting the materials of psycholinguistic experiments, conducting statistical power analyses that directly take into account the experimental materials of interest, gaze-augmented language modeling to enhance performance on NLP downstream tasks, or pre-training complex machine learning models for applications such as assessments of reading competence and document readability.




\backmatter





\bmhead{Acknowledgements}

This work was partially funded by the German Federal Ministry of Education and Research under grant 01$\vert$ S20043 (AEye) and the Swiss National Science Foundation (SNSF) under grant IZCOZ0\_220330/1 (EyeNLG). Shuwen Deng was furthermore funded by the Gertrud-Feiertag-Stipendium from the University of Potsdam. Cui Ding was funded by the SNSF under grant 100015L 212276/1 (MeRID).










\newpage
\begin{appendices}

\section{Eyettention~\cite{deng2023eyettention} Model Configuration}
\label{sec:Eyettention_configure}
Table~\ref{tab: Eyettention_configure} presents the model configuration for the previous version of Eyettention~\cite{deng2023eyettention}, which we have kept consistent in our updated model. For the newly integrated modules that predict landing position and fixation duration, see Table~\ref{tab: hp} for the optimal configuration found during hyperparameter tuning.

\begin{table*}[t!]
\caption{Eyettention~\cite{deng2023eyettention} Model Configuration}
\label{tab: Eyettention_configure}
\begin{center}
    \begin{tabular}{l|l}
    \toprule
    Parameter        & Value           \\\hline
    number of LSTM / BiLSTM layers    & 8 / 8\\
    LSTM / BiLSTM units        & 128 / 64\\
    number of dense layers  & 4\\
    number of dense units  & {512, 256, 256, 256}\\
    embedding dropout rate   &0.4\\
    LSTM / BiLSTM dropout rate &0.2\\
    dense dropout rate  &0.2 \\
    attention window size    & 1\\
    \bottomrule
    \end{tabular}
\end{center}
\end{table*}




\end{appendices}

\bibliography{sn-bibliography}


\begin{thebibliography}{100}
\ifx \bisbn   \undefined \def \bisbn  #1{ISBN #1}\fi
\ifx \binits  \undefined \def \binits#1{#1}\fi
\ifx \bauthor  \undefined \def \bauthor#1{#1}\fi
\ifx \batitle  \undefined \def \batitle#1{#1}\fi
\ifx \bjtitle  \undefined \def \bjtitle#1{#1}\fi
\ifx \bvolume  \undefined \def \bvolume#1{\textbf{#1}}\fi
\ifx \byear  \undefined \def \byear#1{#1}\fi
\ifx \bissue  \undefined \def \bissue#1{#1}\fi
\ifx \bfpage  \undefined \def \bfpage#1{#1}\fi
\ifx \blpage  \undefined \def \blpage #1{#1}\fi
\ifx \burl  \undefined \def \burl#1{\textsf{#1}}\fi
\ifx \doiurl  \undefined \def \doiurl#1{\url{https://doi.org/#1}}\fi
\ifx \betal  \undefined \def \betal{\textit{et al.}}\fi
\ifx \binstitute  \undefined \def \binstitute#1{#1}\fi
\ifx \binstitutionaled  \undefined \def \binstitutionaled#1{#1}\fi
\ifx \bctitle  \undefined \def \bctitle#1{#1}\fi
\ifx \beditor  \undefined \def \beditor#1{#1}\fi
\ifx \bpublisher  \undefined \def \bpublisher#1{#1}\fi
\ifx \bbtitle  \undefined \def \bbtitle#1{#1}\fi
\ifx \bedition  \undefined \def \bedition#1{#1}\fi
\ifx \bseriesno  \undefined \def \bseriesno#1{#1}\fi
\ifx \blocation  \undefined \def \blocation#1{#1}\fi
\ifx \bsertitle  \undefined \def \bsertitle#1{#1}\fi
\ifx \bsnm \undefined \def \bsnm#1{#1}\fi
\ifx \bsuffix \undefined \def \bsuffix#1{#1}\fi
\ifx \bparticle \undefined \def \bparticle#1{#1}\fi
\ifx \barticle \undefined \def \barticle#1{#1}\fi
\bibcommenthead
\ifx \bconfdate \undefined \def \bconfdate #1{#1}\fi
\ifx \botherref \undefined \def \botherref #1{#1}\fi
\ifx \url \undefined \def \url#1{\textsf{#1}}\fi
\ifx \bchapter \undefined \def \bchapter#1{#1}\fi
\ifx \bbook \undefined \def \bbook#1{#1}\fi
\ifx \bcomment \undefined \def \bcomment#1{#1}\fi
\ifx \oauthor \undefined \def \oauthor#1{#1}\fi
\ifx \citeauthoryear \undefined \def \citeauthoryear#1{#1}\fi
\ifx \endbibitem  \undefined \def \endbibitem {}\fi
\ifx \bconflocation  \undefined \def \bconflocation#1{#1}\fi
\ifx \arxivurl  \undefined \def \arxivurl#1{\textsf{#1}}\fi
\csname PreBibitemsHook\endcsname

\bibitem[\protect\citeauthoryear{Rayner}{1998}]{Rayner1998}
\begin{barticle}
\bauthor{\bsnm{Rayner}, \binits{K.}}:
\batitle{Eye movements in reading and information processing: 20 years of research}.
\bjtitle{Psychological Bulletin}
\bvolume{124}(\bissue{3}),
\bfpage{372}--\blpage{422}
(\byear{1998})
\end{barticle}
\endbibitem

\bibitem[\protect\citeauthoryear{Rayner}{2009}]{rayner2009}
\begin{barticle}
\bauthor{\bsnm{Rayner}, \binits{K.}}:
\batitle{The 35th {S}ir {F}rederick {B}artlett {L}ecture: Eye movements and attention in reading, scene perception, and visual search}.
\bjtitle{Quarterly Journal of Experimental Psychology}
\bvolume{62}(\bissue{8}),
\bfpage{1457}--\blpage{1506}
(\byear{2009})
\end{barticle}
\endbibitem

\bibitem[\protect\citeauthoryear{Reichle et~al.}{2003}]{reichle2003ezreader}
\begin{barticle}
\bauthor{\bsnm{Reichle}, \binits{E.}},
\bauthor{\bsnm{Rayner}, \binits{K.}},
\bauthor{\bsnm{A}, \binits{P.}}:
\batitle{The {E}-{Z} reader model of eye-movement control in reading: comparisons to other models}.
\bjtitle{The Behavioral and Brain Sciences}
\bvolume{26},
\bfpage{445}--\blpage{526}
(\byear{2003})
\end{barticle}
\endbibitem

\bibitem[\protect\citeauthoryear{Engbert et~al.}{2002}]{engbert2002dynamical}
\begin{barticle}
\bauthor{\bsnm{Engbert}, \binits{R.}},
\bauthor{\bsnm{Longtin}, \binits{A.}},
\bauthor{\bsnm{Kliegl}, \binits{R.}}:
\batitle{A dynamical model of saccade generation in reading based on spatially distributed lexical processing}.
\bjtitle{Vision Research}
\bvolume{42}(\bissue{5}),
\bfpage{621}--\blpage{636}
(\byear{2002})
\end{barticle}
\endbibitem

\bibitem[\protect\citeauthoryear{Engelmann et~al.}{2013}]{engelmann2013framework}
\begin{barticle}
\bauthor{\bsnm{Engelmann}, \binits{F.}},
\bauthor{\bsnm{Vasishth}, \binits{S.}},
\bauthor{\bsnm{Engbert}, \binits{R.}},
\bauthor{\bsnm{Kliegl}, \binits{R.}}:
\batitle{A framework for modeling the interaction of syntactic processing and eye movement control}.
\bjtitle{Topics in Cognitive Science}
\bvolume{5}(\bissue{3}),
\bfpage{452}--\blpage{474}
(\byear{2013})
\end{barticle}
\endbibitem

\bibitem[\protect\citeauthoryear{Barrett et~al.}{2016}]{barrett-etal-2016-weakly}
\begin{bchapter}
\bauthor{\bsnm{Barrett}, \binits{M.}},
\bauthor{\bsnm{Bingel}, \binits{J.}},
\bauthor{\bsnm{Keller}, \binits{F.}},
\bauthor{\bsnm{S{\o}gaard}, \binits{A.}}:
\bctitle{Weakly supervised part-of-speech tagging using eye-tracking data}.
In: \bbtitle{Proceedings of the 54th Annual Meeting of the Association for Computational Linguistics {(ACL)}},
\bconflocation{Berlin, Germany},
pp. \bfpage{579}--\blpage{584}
(\byear{2016})
\end{bchapter}
\endbibitem

\bibitem[\protect\citeauthoryear{Mishra et~al.}{2016}]{mishra-etal-2016-leveraging}
\begin{bchapter}
\bauthor{\bsnm{Mishra}, \binits{A.}},
\bauthor{\bsnm{Kanojia}, \binits{D.}},
\bauthor{\bsnm{Nagar}, \binits{S.}},
\bauthor{\bsnm{Dey}, \binits{K.}},
\bauthor{\bsnm{Bhattacharyya}, \binits{P.}}:
\bctitle{Leveraging cognitive features for sentiment analysis}.
In: \bbtitle{Proceedings of the 20th {SIGNLL} Conference on Computational Natural Language Learning},
\bconflocation{Berlin, Germany},
pp. \bfpage{156}--\blpage{166}
(\byear{2016})
\end{bchapter}
\endbibitem

\bibitem[\protect\citeauthoryear{Hollenstein and Zhang}{2019}]{hollenstein-zhang-2019-entity}
\begin{bchapter}
\bauthor{\bsnm{Hollenstein}, \binits{N.}},
\bauthor{\bsnm{Zhang}, \binits{C.}}:
\bctitle{Entity recognition at first sight: {I}mproving {NER} with eye movement information}.
In: \bbtitle{Proceedings of North {A}merican Chapter of the Association for Computational Linguistics: Human Language Technologies (NAACL-HLT)},
\bconflocation{Minneapolis, Minnesota},
pp. \bfpage{1}--\blpage{10}
(\byear{2019})
\end{bchapter}
\endbibitem

\bibitem[\protect\citeauthoryear{Barrett et~al.}{2018}]{barrett-etal-2018-sequence}
\begin{bchapter}
\bauthor{\bsnm{Barrett}, \binits{M.}},
\bauthor{\bsnm{Bingel}, \binits{J.}},
\bauthor{\bsnm{Hollenstein}, \binits{N.}},
\bauthor{\bsnm{Rei}, \binits{M.}},
\bauthor{\bsnm{S{\o}gaard}, \binits{A.}}:
\bctitle{Sequence classification with human attention}.
In: \bbtitle{Proceedings of the 22nd Conference on Computational Natural Language Learning (CoNLL)},
\bconflocation{Brussels, Belgium},
pp. \bfpage{302}--\blpage{312}
(\byear{2018})
\end{bchapter}
\endbibitem

\bibitem[\protect\citeauthoryear{Sood et~al.}{2020}]{Sood2020ImprovingAttention}
\begin{bchapter}
\bauthor{\bsnm{Sood}, \binits{E.}},
\bauthor{\bsnm{Tannert}, \binits{S.}},
\bauthor{\bsnm{M{\"u}ller}, \binits{P.}},
\bauthor{\bsnm{Bulling}, \binits{A.}}:
\bctitle{Improving natural language processing tasks with human gaze-guided neural attention}.
In: \bbtitle{Proceedings of the Conference on Neural Information Processing Systems},
vol. \bseriesno{33}.
\bconflocation{Online},
pp. \bfpage{6327}--\blpage{6341}
(\byear{2020})
\end{bchapter}
\endbibitem

\bibitem[\protect\citeauthoryear{Sood et~al.}{2023}]{sood2023multimodal}
\begin{bchapter}
\bauthor{\bsnm{Sood}, \binits{E.}},
\bauthor{\bsnm{K{\"o}gel}, \binits{F.}},
\bauthor{\bsnm{M{\"u}ller}, \binits{P.}},
\bauthor{\bsnm{Thomas}, \binits{D.}},
\bauthor{\bsnm{B{\^a}ce}, \binits{M.}},
\bauthor{\bsnm{Bulling}, \binits{A.}}:
\bctitle{Multimodal integration of human-like attention in visual question answering}.
In: \bbtitle{Proceedings of the IEEE/CVF Conference on Computer Vision and Pattern Recognition Workshops (CVPRW)},
\bconflocation{Vancouver, BC, Canada},
pp. \bfpage{2648}--\blpage{2658}
(\byear{2023})
\end{bchapter}
\endbibitem

\bibitem[\protect\citeauthoryear{Beinborn and Hollenstein}{2023}]{beinborn2024cognitive}
\begin{bbook}
\bauthor{\bsnm{Beinborn}, \binits{L.}},
\bauthor{\bsnm{Hollenstein}, \binits{N.}}:
\bbtitle{Cognitive Plausibility in Natural Language Processing}.
\bpublisher{Springer},
\blocation{Cham}
(\byear{2023})
\end{bbook}
\endbibitem

\bibitem[\protect\citeauthoryear{Sood et~al.}{2020}]{sood2020interpreting}
\begin{bchapter}
\bauthor{\bsnm{Sood}, \binits{E.}},
\bauthor{\bsnm{Tannert}, \binits{S.}},
\bauthor{\bsnm{Frassinelli}, \binits{D.}},
\bauthor{\bsnm{Bulling}, \binits{A.}},
\bauthor{\bsnm{Vu}, \binits{N.T.}}:
\bctitle{Interpreting attention models with human visual attention in machine reading comprehension}.
In: \bbtitle{Proceedings of the 24th Conference on Computational Natural Language Learning},
\bconflocation{Online},
pp. \bfpage{12}--\blpage{25}
(\byear{2020})
\end{bchapter}
\endbibitem

\bibitem[\protect\citeauthoryear{Hollenstein et~al.}{2021}]{hollenstein2021multilingual}
\begin{bchapter}
\bauthor{\bsnm{Hollenstein}, \binits{N.}},
\bauthor{\bsnm{Pirovano}, \binits{F.}},
\bauthor{\bsnm{Zhang}, \binits{C.}},
\bauthor{\bsnm{J{\"a}ger}, \binits{L.}},
\bauthor{\bsnm{Beinborn}, \binits{L.}}:
\bctitle{Multilingual language models predict human reading behavior}.
In: \bbtitle{Proceedings of the 2021 Conference of the North American Chapter of the Association for Computational Linguistics: Human Language Technologies},
\bconflocation{Online},
pp. \bfpage{106}--\blpage{123}
(\byear{2021})
\end{bchapter}
\endbibitem

\bibitem[\protect\citeauthoryear{Hollenstein et~al.}{2022}]{hollenstein2022patterns}
\begin{bchapter}
\bauthor{\bsnm{Hollenstein}, \binits{N.}},
\bauthor{\bsnm{Gonzalez-Dios}, \binits{I.}},
\bauthor{\bsnm{Beinborn}, \binits{L.}},
\bauthor{\bsnm{Jäger}, \binits{L.A.}}:
\bctitle{Patterns of text readability in human and predicted eye movements}.
In: \bbtitle{Proceedings of the Workshop on the Cognitive Aspects of the Lexicon, AACL},
\bconflocation{Online}
(\byear{2022})
\end{bchapter}
\endbibitem

\bibitem[\protect\citeauthoryear{Merkx and Frank}{2021}]{merkx-frank-2021-human}
\begin{bchapter}
\bauthor{\bsnm{Merkx}, \binits{D.}},
\bauthor{\bsnm{Frank}, \binits{S.L.}}:
\bctitle{Human sentence processing: recurrence or attention?}
In: \bbtitle{Proceedings of the Workshop on Cognitive Modeling and Computational Linguistics},
\bconflocation{Online},
pp. \bfpage{12}--\blpage{22}
(\byear{2021})
\end{bchapter}
\endbibitem

\bibitem[\protect\citeauthoryear{Deng et~al.}{2022}]{deng2022detection}
\begin{bchapter}
\bauthor{\bsnm{Deng}, \binits{S.}},
\bauthor{\bsnm{Prasse}, \binits{P.}},
\bauthor{\bsnm{Reich}, \binits{D.R.}},
\bauthor{\bsnm{Dziemian}, \binits{S.}},
\bauthor{\bsnm{Stegenwallner-Schütz}, \binits{M.}},
\bauthor{\bsnm{Krakowczyk}, \binits{D.}},
\bauthor{\bsnm{Makowski}, \binits{S.}},
\bauthor{\bsnm{Langer}, \binits{N.}},
\bauthor{\bsnm{Scheffer}, \binits{T.}},
\bauthor{\bsnm{Jäger}, \binits{L.A.}}:
\bctitle{Detection of {ADHD} based on eye movements during natural viewing}.
In: \bbtitle{Proceedings of the European Conference on Machine Learning and Knowledge Discovery in Databases},
pp. \bfpage{403}--\blpage{418}.
\bpublisher{Springer},
\blocation{Grenoble, France}
(\byear{2022})
\end{bchapter}
\endbibitem

\bibitem[\protect\citeauthoryear{Raatikainen et~al.}{2021}]{Raatikainen2021DetectionData}
\begin{barticle}
\bauthor{\bsnm{Raatikainen}, \binits{P.}},
\bauthor{\bsnm{Hautala}, \binits{J.}},
\bauthor{\bsnm{Loberg}, \binits{O.}},
\bauthor{\bsnm{K{\"{a}}rkk{\"{a}}inen}, \binits{T.}},
\bauthor{\bsnm{Lepp{\"{a}}nen}, \binits{P.}},
\bauthor{\bsnm{Nieminen}, \binits{P.}}:
\batitle{{Detection of developmental dyslexia with machine learning using eye movement data}}.
\bjtitle{Array}
\bvolume{12},
\bfpage{100087}
(\byear{2021})
\end{barticle}
\endbibitem

\bibitem[\protect\citeauthoryear{Haller et~al.}{2022}]{haller2022eye-tracking}
\begin{bchapter}
\bauthor{\bsnm{Haller}, \binits{P.}},
\bauthor{\bsnm{S{\"a}uberli}, \binits{A.}},
\bauthor{\bsnm{Kiener}, \binits{S.}},
\bauthor{\bsnm{Pan}, \binits{J.}},
\bauthor{\bsnm{Yan}, \binits{M.}},
\bauthor{\bsnm{J{\"a}ger}, \binits{L.}}:
\bctitle{Eye-tracking based classification of {M}andarin {C}hinese readers with and without dyslexia using neural sequence models}.
In: \bbtitle{Proceedings of the Workshop on Text Simplification, Accessibility, and Readability (TSAR-2022)},
\bconflocation{Abu Dhabi, United Arab Emirates (Virtual)},
pp. \bfpage{111}--\blpage{118}
(\byear{2022})
\end{bchapter}
\endbibitem

\bibitem[\protect\citeauthoryear{Reich et~al.}{2022}]{reich2022inferring}
\begin{bchapter}
\bauthor{\bsnm{Reich}, \binits{D.R.}},
\bauthor{\bsnm{Prasse}, \binits{P.}},
\bauthor{\bsnm{Tschirner}, \binits{C.}},
\bauthor{\bsnm{Haller}, \binits{P.}},
\bauthor{\bsnm{Goldhammer}, \binits{F.}},
\bauthor{\bsnm{J\"{a}ger}, \binits{L.A.}}:
\bctitle{Inferring native and non-native human reading comprehension and subjective text difficulty from scanpaths in reading}.
In: \bbtitle{Proceedings of the 2022 Symposium on Eye Tracking Research and Applications}.
\bsertitle{ETRA '22},
vol. \bseriesno{23}.
\bconflocation{Seattle, WA, USA}
(\byear{2022})
\end{bchapter}
\endbibitem

\bibitem[\protect\citeauthoryear{Ahn et~al.}{2020}]{Ahn2020TowardsBehavior}
\begin{bchapter}
\bauthor{\bsnm{Ahn}, \binits{S.}},
\bauthor{\bsnm{Kelton}, \binits{C.}},
\bauthor{\bsnm{Balasubramanian}, \binits{A.}},
\bauthor{\bsnm{Zelinsky}, \binits{G.}}:
\bctitle{Towards predicting reading comprehension from gaze behavior}.
In: \bbtitle{Proceedings of the 2020 Symposium on Eye Tracking Research and Applications},
\bconflocation{Stuttgart, Germany},
pp. \bfpage{1}--\blpage{5}
(\byear{2020})
\end{bchapter}
\endbibitem

\bibitem[\protect\citeauthoryear{Berzak et~al.}{2018}]{Berzak2018Assessing}
\begin{bchapter}
\bauthor{\bsnm{Berzak}, \binits{Y.}},
\bauthor{\bsnm{Katz}, \binits{B.}},
\bauthor{\bsnm{Levy}, \binits{R.}}:
\bctitle{Assessing language proficiency from eye movements in reading}.
In: \bbtitle{Proceedings of the 17th Annual Conference of the North American Chapter of the Association for Computational Linguistics: Human Language Technologies},
\bconflocation{New Orleans, Louisiana},
pp. \bfpage{1986}--\blpage{1996}
(\byear{2018})
\end{bchapter}
\endbibitem

\bibitem[\protect\citeauthoryear{Shubi et~al.}{2025}]{shubieyebench}
\begin{bchapter}
\bauthor{\bsnm{Shubi}, \binits{O.}},
\bauthor{\bsnm{Reich}, \binits{D.R.}},
\bauthor{\bsnm{Gruteke~Klein}, \binits{K.}},
\bauthor{\bsnm{Angel}, \binits{Y.}},
\bauthor{\bsnm{Prasse}, \binits{P.}},
\bauthor{\bsnm{J{\"a}ger}, \binits{L.A.}},
\bauthor{\bsnm{Berzak}, \binits{Y.}}:
\bctitle{{EyeBench}: {P}redictive modeling from eye movements in reading}.
In: \bbtitle{Proceedings of the 39th Conference on Neural Information Processing Systems}
(\byear{2025})
\end{bchapter}
\endbibitem

\bibitem[\protect\citeauthoryear{Khurana et~al.}{2023}]{khurana-etal-2023-synthesizing}
\begin{bchapter}
\bauthor{\bsnm{Khurana}, \binits{V.}},
\bauthor{\bsnm{Kumar}, \binits{Y.}},
\bauthor{\bsnm{Hollenstein}, \binits{N.}},
\bauthor{\bsnm{Kumar}, \binits{R.}},
\bauthor{\bsnm{Krishnamurthy}, \binits{B.}}:
\bctitle{Synthesizing human gaze feedback for improved {NLP} performance}.
In: \bbtitle{Proceedings of the 17th Conference of the European Chapter of the Association for Computational Linguistics},
\bconflocation{Dubrovnik, Croatia},
pp. \bfpage{1895}--\blpage{1908}
(\byear{2023})
\end{bchapter}
\endbibitem

\bibitem[\protect\citeauthoryear{Deng et~al.}{2023}]{deng-etal-2023-pre}
\begin{bchapter}
\bauthor{\bsnm{Deng}, \binits{S.}},
\bauthor{\bsnm{Prasse}, \binits{P.}},
\bauthor{\bsnm{Reich}, \binits{D.}},
\bauthor{\bsnm{Scheffer}, \binits{T.}},
\bauthor{\bsnm{J{\"a}ger}, \binits{L.}}:
\bctitle{Pre-trained language models augmented with synthetic scanpaths for natural language understanding}.
In: \bbtitle{Proceedings of the 2023 Conference on Empirical Methods in Natural Language Processing},
\bconflocation{Singapore},
pp. \bfpage{6500}--\blpage{6507}
(\byear{2023})
\end{bchapter}
\endbibitem

\bibitem[\protect\citeauthoryear{Deng et~al.}{2024}]{deng-etal-2024-fine}
\begin{bchapter}
\bauthor{\bsnm{Deng}, \binits{S.}},
\bauthor{\bsnm{Prasse}, \binits{P.}},
\bauthor{\bsnm{Reich}, \binits{D.}},
\bauthor{\bsnm{Scheffer}, \binits{T.}},
\bauthor{\bsnm{J{\"a}ger}, \binits{L.}}:
\bctitle{Fine-tuning pre-trained language models with gaze supervision}.
In: \bbtitle{Proceedings of the 62nd Annual Meeting of the Association for Computational Linguistics (Volume 2: Short Papers)},
\bconflocation{Bangkok, Thailand},
pp. \bfpage{217}--\blpage{224}
(\byear{2024})
\end{bchapter}
\endbibitem

\bibitem[\protect\citeauthoryear{Reich et~al.}{2024}]{reich-etal-2024-reading}
\begin{bchapter}
\bauthor{\bsnm{Reich}, \binits{D.R.}},
\bauthor{\bsnm{Deng}, \binits{S.}},
\bauthor{\bsnm{Bj{\"o}rnsd{\'o}ttir}, \binits{M.}},
\bauthor{\bsnm{J{\"a}ger}, \binits{L.}},
\bauthor{\bsnm{Hollenstein}, \binits{N.}}:
\bctitle{Reading does not equal reading: Comparing, simulating and exploiting reading behavior across populations}.
In: \bbtitle{Proceedings of the 2024 Joint International Conference on Computational Linguistics, Language Resources and Evaluation (LREC-COLING)},
\bconflocation{Torino, Italia},
pp. \bfpage{13586}--\blpage{13594}
(\byear{2024})
\end{bchapter}
\endbibitem

\bibitem[\protect\citeauthoryear{Amrhein et~al.}{2017}]{amrhein2017earth}
\begin{barticle}
\bauthor{\bsnm{Amrhein}, \binits{V.}},
\bauthor{\bsnm{Korner-Nievergelt}, \binits{F.}},
\bauthor{\bsnm{Roth}, \binits{T.}}:
\batitle{The earth is flat (p> 0.05): significance thresholds and the crisis of unreplicable research}.
\bjtitle{PeerJ}
\bvolume{5},
\bfpage{3544}
(\byear{2017})
\end{barticle}
\endbibitem

\bibitem[\protect\citeauthoryear{Collaboration}{2015}]{open2015estimating}
\begin{barticle}
\bauthor{\bsnm{Collaboration}, \binits{O.S.}}:
\batitle{Estimating the reproducibility of psychological science}.
\bjtitle{Science}
\bvolume{349}(\bissue{6251}),
\bfpage{4716}
(\byear{2015})
\end{barticle}
\endbibitem

\bibitem[\protect\citeauthoryear{J{\"a}ger et~al.}{2020}]{jager2020interference}
\begin{barticle}
\bauthor{\bsnm{J{\"a}ger}, \binits{L.A.}},
\bauthor{\bsnm{Mertzen}, \binits{D.}},
\bauthor{\bsnm{Van~Dyke}, \binits{J.A.}},
\bauthor{\bsnm{Vasishth}, \binits{S.}}:
\batitle{Interference patterns in subject-verb agreement and reflexives revisited: A large-sample study}.
\bjtitle{Journal of Memory and Language}
\bvolume{111},
\bfpage{104063}
(\byear{2020})
\end{barticle}
\endbibitem

\bibitem[\protect\citeauthoryear{Vasishth et~al.}{2018}]{vasishthReplicability}
\begin{barticle}
\bauthor{\bsnm{Vasishth}, \binits{S.}},
\bauthor{\bsnm{Mertzen}, \binits{D.}},
\bauthor{\bsnm{Jäger}, \binits{L.A.}},
\bauthor{\bsnm{Gelman}, \binits{A.}}:
\batitle{The statistical significance filter leads to overoptimistic expectations of replicability}.
\bjtitle{Journal of Memory and Language}
\bvolume{103},
\bfpage{151}--\blpage{175}
(\byear{2018})
\end{barticle}
\endbibitem

\bibitem[\protect\citeauthoryear{Prasse et~al.}{2024}]{prasse2024improving}
\begin{barticle}
\bauthor{\bsnm{Prasse}, \binits{P.}},
\bauthor{\bsnm{Reich}, \binits{D.R.}},
\bauthor{\bsnm{Makowski}, \binits{S.}},
\bauthor{\bsnm{Scheffer}, \binits{T.}},
\bauthor{\bsnm{J{\"a}ger}, \binits{L.A.}}:
\batitle{Improving cognitive-state analysis from eye gaze with synthetic eye-movement data}.
\bjtitle{Computers \& Graphics}
\bvolume{119},
\bfpage{103901}
(\byear{2024})
\end{barticle}
\endbibitem

\bibitem[\protect\citeauthoryear{Deng et~al.}{2023}]{deng2023eyettention}
\begin{barticle}
\bauthor{\bsnm{Deng}, \binits{S.}},
\bauthor{\bsnm{Reich}, \binits{D.R.}},
\bauthor{\bsnm{Prasse}, \binits{P.}},
\bauthor{\bsnm{Haller}, \binits{P.}},
\bauthor{\bsnm{Scheffer}, \binits{T.}},
\bauthor{\bsnm{J{\"a}ger}, \binits{L.A.}}:
\batitle{Eyettention: An attention-based dual-sequence model for predicting human scanpaths during reading}.
\bjtitle{Proceedings of the ACM on Human-Computer Interaction}
\bvolume{7}(\bissue{ETRA}),
\bfpage{1}--\blpage{24}
(\byear{2023})
\end{barticle}
\endbibitem

\bibitem[\protect\citeauthoryear{Bolliger et~al.}{2023}]{bolliger-etal-2023-scandl}
\begin{bchapter}
\bauthor{\bsnm{Bolliger}, \binits{L.}},
\bauthor{\bsnm{Reich}, \binits{D.}},
\bauthor{\bsnm{Haller}, \binits{P.}},
\bauthor{\bsnm{Jakobi}, \binits{D.}},
\bauthor{\bsnm{Prasse}, \binits{P.}},
\bauthor{\bsnm{J{\"a}ger}, \binits{L.}}:
\bctitle{{S}can{DL}: A diffusion model for generating synthetic scanpaths on texts}.
In: \beditor{\bsnm{Bouamor}, \binits{H.}},
\beditor{\bsnm{Pino}, \binits{J.}},
\beditor{\bsnm{Bali}, \binits{K.}} (eds.)
\bbtitle{Proceedings of the 2023 Conference on Empirical Methods in Natural Language Processing},
\bconflocation{Singapore},
pp. \bfpage{15513}--\blpage{15538}
(\byear{2023})
\end{bchapter}
\endbibitem

\bibitem[\protect\citeauthoryear{Hahn and Keller}{2016}]{hahn2016modeling}
\begin{bchapter}
\bauthor{\bsnm{Hahn}, \binits{M.}},
\bauthor{\bsnm{Keller}, \binits{F.}}:
\bctitle{Modeling human reading with neural attention}.
In: \bbtitle{Proceedings of the 2016 Conference on Empirical Methods in Natural Language Processing},
\bconflocation{Austin, Texas},
pp. \bfpage{85}--\blpage{95}
(\byear{2016})
\end{bchapter}
\endbibitem

\bibitem[\protect\citeauthoryear{Wang et~al.}{2019}]{wang2019new}
\begin{botherref}
\oauthor{\bsnm{Wang}, \binits{X.}},
\oauthor{\bsnm{Zhao}, \binits{X.}},
\oauthor{\bsnm{Ren}, \binits{J.}}:
A new type of eye movement model based on recurrent neural networks for simulating the gaze behavior of human reading.
Complexity
\textbf{2019}
(2019)
\end{botherref}
\endbibitem

\bibitem[\protect\citeauthoryear{Hollenstein et~al.}{2021a}]{hollenstein-etal-2021-multilingual}
\begin{bchapter}
\bauthor{\bsnm{Hollenstein}, \binits{N.}},
\bauthor{\bsnm{Pirovano}, \binits{F.}},
\bauthor{\bsnm{Zhang}, \binits{C.}},
\bauthor{\bsnm{J{\"a}ger}, \binits{L.}},
\bauthor{\bsnm{Beinborn}, \binits{L.}}:
\bctitle{Multilingual language models predict human reading behavior}.
In: \bbtitle{Proceedings of the 2021 Conference of the North American Chapter of the Association for Computational Linguistics: Human Language Technologies},
\bconflocation{Online},
pp. \bfpage{106}--\blpage{123}
(\byear{2021})
\end{bchapter}
\endbibitem

\bibitem[\protect\citeauthoryear{Hollenstein et~al.}{2021b}]{hollenstein-etal-2021-cmcl}
\begin{bchapter}
\bauthor{\bsnm{Hollenstein}, \binits{N.}},
\bauthor{\bsnm{Chersoni}, \binits{E.}},
\bauthor{\bsnm{Jacobs}, \binits{C.L.}},
\bauthor{\bsnm{Oseki}, \binits{Y.}},
\bauthor{\bsnm{Pr{\'e}vot}, \binits{L.}},
\bauthor{\bsnm{Santus}, \binits{E.}}:
\bctitle{{CMCL} 2021 shared task on eye-tracking prediction}.
In: \bbtitle{Proceedings of the Workshop on Cognitive Modeling and Computational Linguistics},
\bconflocation{Online},
pp. \bfpage{72}--\blpage{78}
(\byear{2021})
\end{bchapter}
\endbibitem

\bibitem[\protect\citeauthoryear{Hollenstein et~al.}{2022}]{hollenstein-etal-2022-cmcl}
\begin{bchapter}
\bauthor{\bsnm{Hollenstein}, \binits{N.}},
\bauthor{\bsnm{Chersoni}, \binits{E.}},
\bauthor{\bsnm{Jacobs}, \binits{C.}},
\bauthor{\bsnm{Oseki}, \binits{Y.}},
\bauthor{\bsnm{Pr{\'e}vot}, \binits{L.}},
\bauthor{\bsnm{Santus}, \binits{E.}}:
\bctitle{{CMCL} 2022 shared task on multilingual and crosslingual prediction of human reading behavior}.
In: \bbtitle{Proceedings of the Workshop on Cognitive Modeling and Computational Linguistics},
\bconflocation{Dublin, Ireland},
pp. \bfpage{121}--\blpage{129}
(\byear{2022})
\end{bchapter}
\endbibitem

\bibitem[\protect\citeauthoryear{Reichle et~al.}{1998}]{reichle1998toward}
\begin{barticle}
\bauthor{\bsnm{Reichle}, \binits{E.D.}},
\bauthor{\bsnm{Pollatsek}, \binits{A.}},
\bauthor{\bsnm{Fisher}, \binits{D.L.}},
\bauthor{\bsnm{Rayner}, \binits{K.}}:
\batitle{Toward a model of eye movement control in reading.}
\bjtitle{Psychological review}
\bvolume{105}(\bissue{1}),
\bfpage{125}
(\byear{1998})
\end{barticle}
\endbibitem

\bibitem[\protect\citeauthoryear{Reichle et~al.}{1999}]{reichle1999eye}
\begin{barticle}
\bauthor{\bsnm{Reichle}, \binits{E.D.}},
\bauthor{\bsnm{Rayner}, \binits{K.}},
\bauthor{\bsnm{Pollatsek}, \binits{A.}}:
\batitle{Eye movement control in reading: Accounting for initial fixation locations and refixations within the {E-Z} {R}eader model}.
\bjtitle{Vision Research}
\bvolume{39}(\bissue{26}),
\bfpage{4403}--\blpage{4411}
(\byear{1999})
\end{barticle}
\endbibitem

\bibitem[\protect\citeauthoryear{Engbert et~al.}{2005}]{engbert2005swift}
\begin{barticle}
\bauthor{\bsnm{Engbert}, \binits{R.}},
\bauthor{\bsnm{Nuthmann}, \binits{A.}},
\bauthor{\bsnm{Richter}, \binits{E.M.}},
\bauthor{\bsnm{Kliegl}, \binits{R.}}:
\batitle{S{WIFT}: a dynamical model of saccade generation during reading.}
\bjtitle{Psychological Review}
\bvolume{112}(\bissue{4}),
\bfpage{777}
(\byear{2005})
\end{barticle}
\endbibitem

\bibitem[\protect\citeauthoryear{Rayner and McConkie}{1976}]{rayner1976guides}
\begin{barticle}
\bauthor{\bsnm{Rayner}, \binits{K.}},
\bauthor{\bsnm{McConkie}, \binits{G.W.}}:
\batitle{What guides a reader's eye movements?}
\bjtitle{Vision Research}
\bvolume{16}(\bissue{8}),
\bfpage{829}--\blpage{837}
(\byear{1976})
\end{barticle}
\endbibitem

\bibitem[\protect\citeauthoryear{Posner et~al.}{1980}]{posner1980attention}
\begin{barticle}
\bauthor{\bsnm{Posner}, \binits{M.I.}},
\bauthor{\bsnm{Snyder}, \binits{C.R.}},
\bauthor{\bsnm{Davidson}, \binits{B.J.}}:
\batitle{Attention and the detection of signals.}
\bjtitle{Journal of Experimental Psychology: General}
\bvolume{109}(\bissue{2}),
\bfpage{160}--\blpage{174}
(\byear{1980})
\end{barticle}
\endbibitem

\bibitem[\protect\citeauthoryear{Bouma and De~Voogd}{1974}]{bouma1974control}
\begin{barticle}
\bauthor{\bsnm{Bouma}, \binits{H.}},
\bauthor{\bsnm{De~Voogd}, \binits{A.}}:
\batitle{On the control of eye saccades in reading}.
\bjtitle{Vision Research}
\bvolume{14}(\bissue{4}),
\bfpage{273}--\blpage{284}
(\byear{1974})
\end{barticle}
\endbibitem

\bibitem[\protect\citeauthoryear{Morrison}{1984}]{morrison1984manipulation}
\begin{barticle}
\bauthor{\bsnm{Morrison}, \binits{R.E.}}:
\batitle{Manipulation of stimulus onset delay in reading: evidence for parallel programming of saccades.}
\bjtitle{Journal of Experimental Psychology: Human Perception and Performance}
\bvolume{10}(\bissue{5}),
\bfpage{667}--\blpage{682}
(\byear{1984})
\end{barticle}
\endbibitem

\bibitem[\protect\citeauthoryear{Engbert and Kliegl}{2001}]{engbert2001mathematical}
\begin{barticle}
\bauthor{\bsnm{Engbert}, \binits{R.}},
\bauthor{\bsnm{Kliegl}, \binits{R.}}:
\batitle{Mathematical models of eye movements in reading: A possible role for autonomous saccades}.
\bjtitle{Biological Cybernetics}
\bvolume{85}(\bissue{2}),
\bfpage{77}--\blpage{87}
(\byear{2001})
\end{barticle}
\endbibitem

\bibitem[\protect\citeauthoryear{Salvucci}{2001}]{salvucci2001integrated}
\begin{barticle}
\bauthor{\bsnm{Salvucci}, \binits{D.D.}}:
\batitle{An integrated model of eye movements and visual encoding}.
\bjtitle{Cognitive Systems Research}
\bvolume{1}(\bissue{4}),
\bfpage{201}--\blpage{220}
(\byear{2001})
\end{barticle}
\endbibitem

\bibitem[\protect\citeauthoryear{Anderson et~al.}{2004}]{anderson2004integrated}
\begin{barticle}
\bauthor{\bsnm{Anderson}, \binits{J.R.}},
\bauthor{\bsnm{Bothell}, \binits{D.}},
\bauthor{\bsnm{Byrne}, \binits{M.D.}},
\bauthor{\bsnm{Douglass}, \binits{S.}},
\bauthor{\bsnm{Lebiere}, \binits{C.}},
\bauthor{\bsnm{Qin}, \binits{Y.}}:
\batitle{An integrated theory of the mind.}
\bjtitle{Psychological Review}
\bvolume{111}(\bissue{4}),
\bfpage{1036}
(\byear{2004})
\end{barticle}
\endbibitem

\bibitem[\protect\citeauthoryear{Veldre et~al.}{2020}]{veldre2020towards}
\begin{bchapter}
\bauthor{\bsnm{Veldre}, \binits{A.}},
\bauthor{\bsnm{Yu}, \binits{L.}},
\bauthor{\bsnm{Andrews}, \binits{S.}},
\bauthor{\bsnm{Reichle}, \binits{E.D.}}:
\bctitle{Towards a complete model of reading: Simulating lexical decision, word naming, and sentence reading with {{\"U}}ber-reader}.
In: \bbtitle{Proceedings of the Annual Meeting of the Cognitive Science Society},
vol. \bseriesno{42}
(\byear{2020})
\end{bchapter}
\endbibitem

\bibitem[\protect\citeauthoryear{Rabe et~al.}{2024}]{rabe2024seam}
\begin{barticle}
\bauthor{\bsnm{Rabe}, \binits{M.M.}},
\bauthor{\bsnm{Paape}, \binits{D.}},
\bauthor{\bsnm{Mertzen}, \binits{D.}},
\bauthor{\bsnm{Vasishth}, \binits{S.}},
\bauthor{\bsnm{Engbert}, \binits{R.}}:
\batitle{Seam: An integrated activation-coupled model of sentence processing and eye movements in reading}.
\bjtitle{Journal of Memory and Language}
\bvolume{135},
\bfpage{104496}
(\byear{2024})
\end{barticle}
\endbibitem

\bibitem[\protect\citeauthoryear{Lewis and Vasishth}{2005}]{lewis2005activation}
\begin{barticle}
\bauthor{\bsnm{Lewis}, \binits{R.L.}},
\bauthor{\bsnm{Vasishth}, \binits{S.}}:
\batitle{An activation-based model of sentence processing as skilled memory retrieval}.
\bjtitle{Cognitive Science}
\bvolume{29}(\bissue{3}),
\bfpage{375}--\blpage{419}
(\byear{2005})
\end{barticle}
\endbibitem

\bibitem[\protect\citeauthoryear{Mancheva et~al.}{2015}]{mancheva04072015}
\begin{barticle}
\bauthor{\bsnm{Mancheva}, \binits{L.}},
\bauthor{\bsnm{Reichle}, \binits{E.}},
\bauthor{\bsnm{Lemaire}, \binits{B.}},
\bauthor{\bsnm{Valdois}, \binits{S.}},
\bauthor{\bsnm{Ecalle}, \binits{J.}},
\bauthor{\bsnm{Guérin-Dugué}, \binits{A.}}:
\batitle{An analysis of reading skill development using {E-Z} reader}.
\bjtitle{Journal of Cognitive Psychology}
\bvolume{27}(\bissue{5}),
\bfpage{657}--\blpage{676}
(\byear{2015})
\end{barticle}
\endbibitem

\bibitem[\protect\citeauthoryear{Reichle et~al.}{2013}]{reichle2013using}
\begin{barticle}
\bauthor{\bsnm{Reichle}, \binits{E.D.}},
\bauthor{\bsnm{Liversedge}, \binits{S.P.}},
\bauthor{\bsnm{Drieghe}, \binits{D.}},
\bauthor{\bsnm{Blythe}, \binits{H.I.}},
\bauthor{\bsnm{Joseph}, \binits{H.S.}},
\bauthor{\bsnm{White}, \binits{S.J.}},
\bauthor{\bsnm{Rayner}, \binits{K.}}:
\batitle{Using {E-Z} reader to examine the concurrent development of eye-movement control and reading skill}.
\bjtitle{Developmental Review}
\bvolume{33}(\bissue{2}),
\bfpage{110}--\blpage{149}
(\byear{2013})
\end{barticle}
\endbibitem

\bibitem[\protect\citeauthoryear{Nilsson and Nivre}{2009}]{nilsson2009learning}
\begin{bchapter}
\bauthor{\bsnm{Nilsson}, \binits{M.}},
\bauthor{\bsnm{Nivre}, \binits{J.}}:
\bctitle{Learning where to look: Modeling eye movements in reading}.
In: \bbtitle{Proceedings of the Thirteenth Conference on Computational Natural Language Learning},
\bconflocation{Boulder, Colorado},
pp. \bfpage{93}--\blpage{101}
(\byear{2009})
\end{bchapter}
\endbibitem

\bibitem[\protect\citeauthoryear{Nilsson and Nivre}{2010}]{nilsson2010towards}
\begin{bchapter}
\bauthor{\bsnm{Nilsson}, \binits{M.}},
\bauthor{\bsnm{Nivre}, \binits{J.}}:
\bctitle{Towards a data-driven model of eye movement control in reading}.
In: \bbtitle{Proceedings of the 2010 Workshop on Cognitive Modeling and Computational Linguistics, ACL},
\bconflocation{Uppsala, Sweden},
pp. \bfpage{63}--\blpage{71}
(\byear{2010})
\end{bchapter}
\endbibitem

\bibitem[\protect\citeauthoryear{Nilsson and Nivre}{2011}]{nilsson2011entropy}
\begin{bchapter}
\bauthor{\bsnm{Nilsson}, \binits{M.}},
\bauthor{\bsnm{Nivre}, \binits{J.}}:
\bctitle{Entropy-driven evaluation of models of eye movement control in reading}.
In: \bbtitle{Proceedings of the 8th International NLPCS Workshop},
\bconflocation{Copenhagen, Denmark},
pp. \bfpage{201}--\blpage{212}
(\byear{2011})
\end{bchapter}
\endbibitem

\bibitem[\protect\citeauthoryear{Matthies and S{\o}gaard}{2013}]{matthies-sogaard-2013-blinkers}
\begin{bchapter}
\bauthor{\bsnm{Matthies}, \binits{F.}},
\bauthor{\bsnm{S{\o}gaard}, \binits{A.}}:
\bctitle{With blinkers on: Robust prediction of eye movements across readers}.
In: \bbtitle{Proceedings of the 2013 Conference on Empirical Methods in Natural Language Processing},
\bconflocation{Seattle, Washington, USA},
pp. \bfpage{803}--\blpage{807}
(\byear{2013})
\end{bchapter}
\endbibitem

\bibitem[\protect\citeauthoryear{Hara et~al.}{2012}]{hara-etal-2012-predicting}
\begin{bchapter}
\bauthor{\bsnm{Hara}, \binits{T.}},
\bauthor{\bsnm{Mochihashi}, \binits{D.}},
\bauthor{\bsnm{Kano}, \binits{Y.}},
\bauthor{\bsnm{Aizawa}, \binits{A.}}:
\bctitle{Predicting word fixations in text with a {CRF} model for capturing general reading strategies among readers}.
In: \bbtitle{Proceedings of the First Workshop on Eye-tracking and Natural Language Processing},
\bconflocation{Mumbai, India},
pp. \bfpage{55}--\blpage{70}
(\byear{2012})
\end{bchapter}
\endbibitem

\bibitem[\protect\citeauthoryear{D'Agostino et~al.}{2025}]{dmoves}
\begin{bchapter}
\bauthor{\bsnm{D'Agostino}, \binits{F.}},
\bauthor{\bsnm{Schwetlick}, \binits{L.}},
\bauthor{\bsnm{Bethge}, \binits{M.}},
\bauthor{\bsnm{Kuemmerer}, \binits{M.}}:
\bctitle{What moves the eyes: Doubling mechanistic model performance using deep networks to discover and test cognitive hypotheses}.
In: \bbtitle{Proceedings of the 39th Annual Conference on Neural Information Processing Systems}
(\byear{2025})
\end{bchapter}
\endbibitem

\bibitem[\protect\citeauthoryear{Seelig et~al.}{2020}]{seelig2020bayesian}
\begin{barticle}
\bauthor{\bsnm{Seelig}, \binits{S.A.}},
\bauthor{\bsnm{Rabe}, \binits{M.M.}},
\bauthor{\bsnm{Malem-Shinitski}, \binits{N.}},
\bauthor{\bsnm{Risse}, \binits{S.}},
\bauthor{\bsnm{Reich}, \binits{S.}},
\bauthor{\bsnm{Engbert}, \binits{R.}}:
\batitle{Bayesian parameter estimation for the {SWIFT} model of eye-movement control during reading}.
\bjtitle{Journal of Mathematical Psychology}
\bvolume{95},
\bfpage{102313}
(\byear{2020})
\end{barticle}
\endbibitem

\bibitem[\protect\citeauthoryear{K{\"u}mmerer et~al.}{2022}]{kummerer2022deepgaze}
\begin{barticle}
\bauthor{\bsnm{K{\"u}mmerer}, \binits{M.}},
\bauthor{\bsnm{Bethge}, \binits{M.}},
\bauthor{\bsnm{Wallis}, \binits{T.S.}}:
\batitle{Deepgaze iii: Modeling free-viewing human scanpaths with deep learning}.
\bjtitle{Journal of Vision}
\bvolume{22}(\bissue{5}),
\bfpage{7}--\blpage{7}
(\byear{2022})
\end{barticle}
\endbibitem

\bibitem[\protect\citeauthoryear{Devlin et~al.}{2019}]{devlin2018bert}
\begin{bchapter}
\bauthor{\bsnm{Devlin}, \binits{J.}},
\bauthor{\bsnm{Chang}, \binits{M.W.}},
\bauthor{\bsnm{Lee}, \binits{K.}},
\bauthor{\bsnm{Toutanova}, \binits{K.}}:
\bctitle{{BERT: Pre-training of deep bidirectional transformers for language understanding}}.
In: \bbtitle{Proceedings of the Conference of the North American Chapter of the Association for Computational Linguistics: Human Language Technologies},
vol. \bseriesno{1}.
\bconflocation{Minneapolis, MN, USA},
pp. \bfpage{4171}--\blpage{4186}
(\byear{2019})
\end{bchapter}
\endbibitem

\bibitem[\protect\citeauthoryear{Liu et~al.}{2019}]{liu2019roberta}
\begin{botherref}
\oauthor{\bsnm{Liu}, \binits{Y.}},
\oauthor{\bsnm{Ott}, \binits{M.}},
\oauthor{\bsnm{Goyal}, \binits{N.}},
\oauthor{\bsnm{Du}, \binits{J.}},
\oauthor{\bsnm{Joshi}, \binits{M.}},
\oauthor{\bsnm{Chen}, \binits{D.}},
\oauthor{\bsnm{Levy}, \binits{O.}},
\oauthor{\bsnm{Lewis}, \binits{M.}},
\oauthor{\bsnm{Zettlemoyer}, \binits{L.}},
\oauthor{\bsnm{Stoyanov}, \binits{V.}}:
Roberta: A robustly optimized {BERT} pretraining approach.
arXiv preprint arXiv:1907.11692
(2019)
\end{botherref}
\endbibitem

\bibitem[\protect\citeauthoryear{Hochreiter and Schmidhuber}{1997}]{hochreiter1997long}
\begin{barticle}
\bauthor{\bsnm{Hochreiter}, \binits{S.}},
\bauthor{\bsnm{Schmidhuber}, \binits{J.}}:
\batitle{Long short-term memory}.
\bjtitle{Neural Computation}
\bvolume{9}(\bissue{8}),
\bfpage{1735}--\blpage{1780}
(\byear{1997})
\end{barticle}
\endbibitem

\bibitem[\protect\citeauthoryear{Graves and Schmidhuber}{2005}]{graves2005framewise}
\begin{barticle}
\bauthor{\bsnm{Graves}, \binits{A.}},
\bauthor{\bsnm{Schmidhuber}, \binits{J.}}:
\batitle{Framewise phoneme classification with bidirectional lstm and other neural network architectures}.
\bjtitle{Neural Networks}
\bvolume{18}(\bissue{5-6}),
\bfpage{602}--\blpage{610}
(\byear{2005})
\end{barticle}
\endbibitem

\bibitem[\protect\citeauthoryear{Vaswani et~al.}{2017}]{vaswani2017attention}
\begin{bchapter}
\bauthor{\bsnm{Vaswani}, \binits{A.}},
\bauthor{\bsnm{Shazeer}, \binits{N.}},
\bauthor{\bsnm{Parmar}, \binits{N.}},
\bauthor{\bsnm{Uszkoreit}, \binits{J.}},
\bauthor{\bsnm{Jones}, \binits{L.}},
\bauthor{\bsnm{Gomez}, \binits{A.N.}},
\bauthor{\bsnm{Kaiser}, \binits{{\L}.}},
\bauthor{\bsnm{Polosukhin}, \binits{I.}}:
\bctitle{Attention is all you need}.
In: \bbtitle{Proceedings of the 31th Conference on Neural Information Processing Systems},
vol. \bseriesno{30}.
\bconflocation{Long Beach, CA, USA}
(\byear{2017})
\end{bchapter}
\endbibitem

\bibitem[\protect\citeauthoryear{Bargary et~al.}{2017}]{bargary2017individual}
\begin{barticle}
\bauthor{\bsnm{Bargary}, \binits{G.}},
\bauthor{\bsnm{Bosten}, \binits{J.M.}},
\bauthor{\bsnm{Goodbourn}, \binits{P.T.}},
\bauthor{\bsnm{Lawrance-Owen}, \binits{A.J.}},
\bauthor{\bsnm{Hogg}, \binits{R.E.}},
\bauthor{\bsnm{Mollon}, \binits{J.D.}}:
\batitle{Individual differences in human eye movements: An oculomotor signature?}
\bjtitle{Vision research}
\bvolume{141},
\bfpage{157}--\blpage{169}
(\byear{2017})
\end{barticle}
\endbibitem

\bibitem[\protect\citeauthoryear{J\"ager et~al.}{2020}]{JaegerECML2019}
\begin{bchapter}
\bauthor{\bsnm{J\"ager}, \binits{L.A.}},
\bauthor{\bsnm{Makowski}, \binits{S.}},
\bauthor{\bsnm{Prasse}, \binits{P.}},
\bauthor{\bsnm{Sascha}, \binits{L.}},
\bauthor{\bsnm{Seidler}, \binits{M.}},
\bauthor{\bsnm{Scheffer}, \binits{T.}}:
\bctitle{{Deep Eyedentification}: {B}iometric identification using micro-movements of the eye}.
In: \beditor{\bsnm{Brefeld}},
\beditor{\bsnm{Fromont}},
\beditor{\bsnm{Knobbe}},
\beditor{\bsnm{Hotho}},
\beditor{\bsnm{Maathuis}},
\beditor{\bsnm{Robardet}} (eds.)
\bbtitle{Machine Learning and Knowledge Discovery in Databases. ECML PKDD 2019}.
\bsertitle{Lecture Notes in Computer Science},
vol. \bseriesno{11907},
pp. \bfpage{299}--\blpage{314}.
\bpublisher{Springer},
\blocation{Cham, Switzerland}
(\byear{2020})
\end{bchapter}
\endbibitem

\bibitem[\protect\citeauthoryear{Makowski et~al.}{2021}]{makowski2021deepeyedentificationlive}
\begin{barticle}
\bauthor{\bsnm{Makowski}, \binits{S.}},
\bauthor{\bsnm{Prasse}, \binits{P.}},
\bauthor{\bsnm{Reich}, \binits{D.R.}},
\bauthor{\bsnm{Krakowczyk}, \binits{D.}},
\bauthor{\bsnm{J{\"a}ger}, \binits{L.A.}},
\bauthor{\bsnm{Scheffer}, \binits{T.}}:
\batitle{Deepeyedentificationlive: Oculomotoric biometric identification and presentation-attack detection using deep neural networks}.
\bjtitle{IEEE Transactions on Biometrics, Behavior, and Identity Science}
\bvolume{3}(\bissue{4}),
\bfpage{506}--\blpage{518}
(\byear{2021})
\end{barticle}
\endbibitem

\bibitem[\protect\citeauthoryear{Williams and Zipser}{1989}]{williams1989learning}
\begin{barticle}
\bauthor{\bsnm{Williams}, \binits{R.J.}},
\bauthor{\bsnm{Zipser}, \binits{D.}}:
\batitle{A learning algorithm for continually running fully recurrent neural networks}.
\bjtitle{Neural computation}
\bvolume{1}(\bissue{2}),
\bfpage{270}--\blpage{280}
(\byear{1989})
\end{barticle}
\endbibitem

\bibitem[\protect\citeauthoryear{Pan et~al.}{2021}]{pan2021bsc}
\begin{botherref}
\oauthor{\bsnm{Pan}, \binits{J.}},
\oauthor{\bsnm{Yan}, \binits{M.}},
\oauthor{\bsnm{Richter}, \binits{E.M.}},
\oauthor{\bsnm{Shu}, \binits{H.}},
\oauthor{\bsnm{Kliegl}, \binits{R.}}:
The {B}eijing {S}entence {C}orpus: A {C}hinese sentence corpus with eye movement data and predictability norms.
Behavior Research Methods
\textbf{2021}
(2021)
\end{botherref}
\endbibitem

\bibitem[\protect\citeauthoryear{Berzak et~al.}{2022}]{berzak2022celer}
\begin{botherref}
\oauthor{\bsnm{Berzak}, \binits{Y.}},
\oauthor{\bsnm{Nakamura}, \binits{C.}},
\oauthor{\bsnm{Smith}, \binits{A.}},
\oauthor{\bsnm{Weng}, \binits{E.}},
\oauthor{\bsnm{Katz}, \binits{B.}},
\oauthor{\bsnm{Flynn}, \binits{S.}},
\oauthor{\bsnm{Levy}, \binits{R.}}:
{CELER: A 365-participant corpus of eye movements in L1 and L2 English reading}.
Open Mind,
1--10
(2022)
\end{botherref}
\endbibitem

\bibitem[\protect\citeauthoryear{Hollenstein et~al.}{2018}]{hollenstein2018zuco}
\begin{barticle}
\bauthor{\bsnm{Hollenstein}, \binits{N.}},
\bauthor{\bsnm{Rotsztejn}, \binits{J.}},
\bauthor{\bsnm{Troendle}, \binits{M.}},
\bauthor{\bsnm{Pedroni}, \binits{A.}},
\bauthor{\bsnm{Zhang}, \binits{C.}},
\bauthor{\bsnm{Langer}, \binits{N.}}:
\batitle{Zuco, a simultaneous {EEG} and eye-tracking resource for natural sentence reading}.
\bjtitle{Scientific Data}
\bvolume{5},
\bfpage{180291}
(\byear{2018})
\end{barticle}
\endbibitem

\bibitem[\protect\citeauthoryear{Hollenstein et~al.}{2020}]{hollenstein2019zuco2}
\begin{bchapter}
\bauthor{\bsnm{Hollenstein}, \binits{N.}},
\bauthor{\bsnm{Troendle}, \binits{M.}},
\bauthor{\bsnm{Zhang}, \binits{C.}},
\bauthor{\bsnm{Langer}, \binits{N.}}:
\bctitle{{Z}u{C}o 2.0: A dataset of physiological recordings during natural reading and annotation}.
In: \bbtitle{Proceedings of the Twelfth Language Resources and Evaluation Conference},
\bconflocation{Marseille, France},
pp. \bfpage{138}--\blpage{146}
(\byear{2020})
\end{bchapter}
\endbibitem

\bibitem[\protect\citeauthoryear{Bestgen}{2021}]{bestgen-2021-last}
\begin{bchapter}
\bauthor{\bsnm{Bestgen}, \binits{Y.}}:
\bctitle{{LAST} at {CMCL} 2021 shared task: Predicting gaze data during reading with a gradient boosting decision tree approach}.
In: \bbtitle{Proceedings of the Workshop on Cognitive Modeling and Computational Linguistics},
pp. \bfpage{90}--\blpage{96}.
\bpublisher{Association for Computational Linguistics},
\blocation{Online}
(\byear{2021})
\end{bchapter}
\endbibitem

\bibitem[\protect\citeauthoryear{Rayner et~al.}{2007}]{rayner2007chineseezreader}
\begin{barticle}
\bauthor{\bsnm{Rayner}, \binits{K.}},
\bauthor{\bsnm{Li}, \binits{X.}},
\bauthor{\bsnm{Pollatsek}, \binits{A.}}:
\batitle{Extending the {E-Z} reader model of eye movement control to chinese readers}.
\bjtitle{Cognitive Science}
\bvolume{31}(\bissue{6}),
\bfpage{1021}--\blpage{1033}
(\byear{2007})
\end{barticle}
\endbibitem

\bibitem[\protect\citeauthoryear{Rabe et~al.}{2021}]{rabe2021bayes}
\begin{barticle}
\bauthor{\bsnm{Rabe}, \binits{M.M.}},
\bauthor{\bsnm{Chandra}, \binits{J.}},
\bauthor{\bsnm{Kr\"{u}gel}, \binits{A.}},
\bauthor{\bsnm{Seelig}, \binits{S.A.}},
\bauthor{\bsnm{Vasishth}, \binits{S.}},
\bauthor{\bsnm{Engbert}, \binits{R.}}:
\batitle{A bayesian approach to dynamical modeling of eye-movement control in reading of normal, mirrored, and scrambled texts.}
\bjtitle{Psychol Rev}
\bvolume{128},
\bfpage{803}--\blpage{823}
(\byear{2021})
\end{barticle}
\endbibitem

\bibitem[\protect\citeauthoryear{Jarodzka et~al.}{2010}]{jarodzka2010vector}
\begin{bchapter}
\bauthor{\bsnm{Jarodzka}, \binits{H.}},
\bauthor{\bsnm{Holmqvist}, \binits{K.}},
\bauthor{\bsnm{Nystr{\"o}m}, \binits{M.}}:
\bctitle{A vector-based, multidimensional scanpath similarity measure}.
In: \bbtitle{Proceedings of the 2010 Symposium on Eye-Tracking Research and Applications}.
\bsertitle{ETRA '10},
pp. \bfpage{211}--\blpage{218},
\bconflocation{Austin, Texas}
(\byear{2010})
\end{bchapter}
\endbibitem

\bibitem[\protect\citeauthoryear{Wagner et~al.}{2019}]{wagner2019multimatch}
\begin{barticle}
\bauthor{\bsnm{Wagner}, \binits{A.S.}},
\bauthor{\bsnm{Halchenko}, \binits{Y.O.}},
\bauthor{\bsnm{Hanke}, \binits{M.}}:
\batitle{multimatch-gaze: The multimatch algorithm for gaze path comparison in python}.
\bjtitle{Journal of Open Source Software}
\bvolume{4}(\bissue{40}),
\bfpage{1525}
(\byear{2019})
\end{barticle}
\endbibitem

\bibitem[\protect\citeauthoryear{Kingma and Ba}{2015}]{kingma2014adam}
\begin{bchapter}
\bauthor{\bsnm{Kingma}, \binits{D.P.}},
\bauthor{\bsnm{Ba}, \binits{J.}}:
\bctitle{Adam: {A} method for stochastic optimization}.
In: \bbtitle{Proceedings of the 3rd International Conference on Learning Representations ({ICLR})}
(\byear{2015})
\end{bchapter}
\endbibitem

\bibitem[\protect\citeauthoryear{Paszke et~al.}{2019}]{pytorch2019paszke}
\begin{bchapter}
\bauthor{\bsnm{Paszke}, \binits{A.}},
\bauthor{\bsnm{Gross}, \binits{S.}},
\bauthor{\bsnm{Massa}, \binits{F.}},
\bauthor{\bsnm{Lerer}, \binits{A.}},
\bauthor{\bsnm{Bradbury}, \binits{J.}},
\bauthor{\bsnm{Chanan}, \binits{G.}},
\bauthor{\bsnm{Killeen}, \binits{T.}},
\bauthor{\bsnm{Lin}, \binits{Z.}},
\bauthor{\bsnm{Gimelshein}, \binits{N.}},
\bauthor{\bsnm{Antiga}, \binits{L.}},
\bauthor{\bsnm{Desmaison}, \binits{A.}},
\bauthor{\bsnm{Kopf}, \binits{A.}},
\bauthor{\bsnm{Yang}, \binits{E.}},
\bauthor{\bsnm{DeVito}, \binits{Z.}},
\bauthor{\bsnm{Raison}, \binits{M.}},
\bauthor{\bsnm{Tejani}, \binits{A.}},
\bauthor{\bsnm{Chilamkurthy}, \binits{S.}},
\bauthor{\bsnm{Steiner}, \binits{B.}},
\bauthor{\bsnm{Fang}, \binits{L.}},
\bauthor{\bsnm{Bai}, \binits{J.}},
\bauthor{\bsnm{Chintala}, \binits{S.}}:
\bctitle{Pytorch: An imperative style, high-performance deep learning library}.
In: \bbtitle{Proceedings of the 33rd International Conference on Neural Information Processing Systems},
pp. \bfpage{8024}--\blpage{8035}
(\byear{2019})
\end{bchapter}
\endbibitem

\bibitem[\protect\citeauthoryear{Wilcox et~al.}{2024}]{wilcox2024mouse}
\begin{barticle}
\bauthor{\bsnm{Wilcox}, \binits{E.G.}},
\bauthor{\bsnm{Ding}, \binits{C.}},
\bauthor{\bsnm{Sachan}, \binits{M.}},
\bauthor{\bsnm{J{\"a}ger}, \binits{L.A.}}:
\batitle{Mouse tracking for reading ({MoTR}): A new naturalistic incremental processing measurement tool}.
\bjtitle{Journal of Memory and Language}
\bvolume{138},
\bfpage{104534}
(\byear{2024})
\end{barticle}
\endbibitem

\bibitem[\protect\citeauthoryear{Jakobi et~al.}{2024}]{jakobi2024potec}
\begin{botherref}
\oauthor{\bsnm{Jakobi}, \binits{D.N.}},
\oauthor{\bsnm{Kern}, \binits{T.}},
\oauthor{\bsnm{Reich}, \binits{D.R.}},
\oauthor{\bsnm{Haller}, \binits{P.}},
\oauthor{\bsnm{J{\"a}ger}, \binits{L.A.}}:
Potec: A {G}erman naturalistic eye-tracking-while-reading corpus.
arXiv preprint arXiv:2403.00506
(2024)
\end{botherref}
\endbibitem

\bibitem[\protect\citeauthoryear{Bates et~al.}{2005}]{bates2005fitting}
\begin{barticle}
\bauthor{\bsnm{Bates}, \binits{D.}}, \betal:
\batitle{Fitting linear mixed models in {R}}.
\bjtitle{R news}
\bvolume{5}(\bissue{1}),
\bfpage{27}--\blpage{30}
(\byear{2005})
\end{barticle}
\endbibitem

\bibitem[\protect\citeauthoryear{{R Core Team}}{2024}]{rteam}
\begin{bbook}
\bauthor{\bsnm{{R Core Team}}}:
\bbtitle{R: A Language and Environment for Statistical Computing}.
\bpublisher{R Foundation for Statistical Computing},
\blocation{Vienna, Austria}
(\byear{2024}).
\bcomment{R Foundation for Statistical Computing}.
\burl{https://www.R-project.org/}
\end{bbook}
\endbibitem

\bibitem[\protect\citeauthoryear{Bürkner}{2021}]{brms2021}
\begin{barticle}
\bauthor{\bsnm{Bürkner}, \binits{P.-C.}}:
\batitle{Bayesian item response modeling in {R} with {brms} and {Stan}}.
\bjtitle{Journal of Statistical Software}
\bvolume{100}(\bissue{5}),
\bfpage{1}--\blpage{54}
(\byear{2021})
\end{barticle}
\endbibitem

\bibitem[\protect\citeauthoryear{Nicenboim et~al.}{2025}]{nicenboim2021introduction}
\begin{bbook}
\bauthor{\bsnm{Nicenboim}, \binits{B.}},
\bauthor{\bsnm{Schad}, \binits{D.J.}},
\bauthor{\bsnm{Vasishth}, \binits{S.}}:
\bbtitle{Introduction to Bayesian Data Analysis for Cognitive Science},
\bedition{1st} edn.
\bpublisher{CRC Press},
\blocation{Boca Raton, FL}
(\byear{2025})
\end{bbook}
\endbibitem

\bibitem[\protect\citeauthoryear{Ackley et~al.}{1985}]{ackley1985learning}
\begin{barticle}
\bauthor{\bsnm{Ackley}, \binits{D.H.}},
\bauthor{\bsnm{Hinton}, \binits{G.E.}},
\bauthor{\bsnm{Sejnowski}, \binits{T.J.}}:
\batitle{A learning algorithm for boltzmann machines}.
\bjtitle{Cognitive science}
\bvolume{9}(\bissue{1}),
\bfpage{147}--\blpage{169}
(\byear{1985})
\end{barticle}
\endbibitem

\bibitem[\protect\citeauthoryear{Holtzman et~al.}{2020}]{holtzmancurious}
\begin{bchapter}
\bauthor{\bsnm{Holtzman}, \binits{A.}},
\bauthor{\bsnm{Buys}, \binits{J.}},
\bauthor{\bsnm{Du}, \binits{L.}},
\bauthor{\bsnm{Forbes}, \binits{M.}},
\bauthor{\bsnm{Choi}, \binits{Y.}}:
\bctitle{The curious case of neural text degeneration}.
In: \bbtitle{Proceedings of the 8th International Conference on Learning Representations},
\bconflocation{Online}
(\byear{2020})
\end{bchapter}
\endbibitem

\bibitem[\protect\citeauthoryear{Kaplan et~al.}{2020}]{kaplan2020scaling}
\begin{botherref}
\oauthor{\bsnm{Kaplan}, \binits{J.}},
\oauthor{\bsnm{McCandlish}, \binits{S.}},
\oauthor{\bsnm{Henighan}, \binits{T.}},
\oauthor{\bsnm{Brown}, \binits{T.B.}},
\oauthor{\bsnm{Chess}, \binits{B.}},
\oauthor{\bsnm{Child}, \binits{R.}},
\oauthor{\bsnm{Gray}, \binits{S.}},
\oauthor{\bsnm{Radford}, \binits{A.}},
\oauthor{\bsnm{Wu}, \binits{J.}},
\oauthor{\bsnm{Amodei}, \binits{D.}}:
Scaling laws for neural language models.
arXiv preprint arXiv:2001.08361
(2020)
\end{botherref}
\endbibitem

\bibitem[\protect\citeauthoryear{Radford et~al.}{2019}]{radford2019language}
\begin{botherref}
\oauthor{\bsnm{Radford}, \binits{A.}},
\oauthor{\bsnm{Wu}, \binits{J.}},
\oauthor{\bsnm{Child}, \binits{R.}},
\oauthor{\bsnm{Luan}, \binits{D.}},
\oauthor{\bsnm{Amodei}, \binits{D.}},
\oauthor{\bsnm{Sutskever}, \binits{I.}}:
Language models are unsupervised multitask learners
(2019)
\end{botherref}
\endbibitem

\bibitem[\protect\citeauthoryear{Touvron et~al.}{2023}]{touvron2023llama}
\begin{botherref}
\oauthor{\bsnm{Touvron}, \binits{H.}},
\oauthor{\bsnm{Martin}, \binits{L.}},
\oauthor{\bsnm{Stone}, \binits{K.}},
\oauthor{\bsnm{Albert}, \binits{P.}},
\oauthor{\bsnm{Almahairi}, \binits{A.}},
\oauthor{\bsnm{Babaei}, \binits{Y.}},
\oauthor{\bsnm{Bashlykov}, \binits{N.}},
\oauthor{\bsnm{Batra}, \binits{S.}},
\oauthor{\bsnm{Bhargava}, \binits{P.}},
\oauthor{\bsnm{Bhosale}, \binits{S.}}, et al.:
Llama 2: Open foundation and fine-tuned chat models.
arXiv preprint arXiv:2307.09288
(2023)
\end{botherref}
\endbibitem

\bibitem[\protect\citeauthoryear{Jawahar et~al.}{2019}]{jawahar2019does}
\begin{bchapter}
\bauthor{\bsnm{Jawahar}, \binits{G.}},
\bauthor{\bsnm{Sagot}, \binits{B.}},
\bauthor{\bsnm{Seddah}, \binits{D.}}:
\bctitle{What does {BERT} learn about the structure of language?}
In: \bbtitle{Proceedings of the 57th Annual Meeting of the Association for Computational Linguistics},
pp. \bfpage{3651}--\blpage{3657}
(\byear{2019})
\end{bchapter}
\endbibitem

\bibitem[\protect\citeauthoryear{Kuribayashi et~al.}{2022}]{kuribayashi-etal-2022-context}
\begin{bchapter}
\bauthor{\bsnm{Kuribayashi}, \binits{T.}},
\bauthor{\bsnm{Oseki}, \binits{Y.}},
\bauthor{\bsnm{Brassard}, \binits{A.}},
\bauthor{\bsnm{Inui}, \binits{K.}}:
\bctitle{Context limitations make neural language models more human-like}.
In: \bbtitle{Proceedings of the 2022 {C}onference on {E}mpirical {M}ethods in {N}atural {L}anguage {P}rocessing},
\bconflocation{Abu Dhabi, United Arab Emirates},
pp. \bfpage{10421}--\blpage{10436}
(\byear{2022})
\end{bchapter}
\endbibitem

\bibitem[\protect\citeauthoryear{McCurdy and Hahn}{2024}]{mccurdy-hahn-2024-lossy}
\begin{bchapter}
\bauthor{\bsnm{McCurdy}, \binits{K.}},
\bauthor{\bsnm{Hahn}, \binits{M.}}:
\bctitle{Lossy context surprisal predicts task-dependent patterns in relative clause processing}.
In: \bbtitle{Proceedings of the 28th {C}onference on {C}omputational {N}atural {L}anguage {L}earning},
\bconflocation{Miami, FL, USA},
pp. \bfpage{36}--\blpage{45}
(\byear{2024})
\end{bchapter}
\endbibitem

\bibitem[\protect\citeauthoryear{Hennert et~al.}{2025}]{hennertcan}
\begin{botherref}
\oauthor{\bsnm{Hennert}, \binits{J.}},
\oauthor{\bsnm{Paape}, \binits{D.}},
\oauthor{\bsnm{Yadav}, \binits{H.}},
\oauthor{\bsnm{Vasishth}, \binits{S.}}:
Can lossy-context surprisal capture both locality and anti-locality effects? {A} model evaluation using {R}ussian, {H}indi, and {P}ersian data.
OSF
(2025)
\end{botherref}
\endbibitem

\bibitem[\protect\citeauthoryear{Oh et~al.}{2022}]{oh2022comparison}
\begin{barticle}
\bauthor{\bsnm{Oh}, \binits{B.-D.}},
\bauthor{\bsnm{Clark}, \binits{C.}},
\bauthor{\bsnm{Schuler}, \binits{W.}}:
\batitle{Comparison of structural parsers and neural language models as surprisal estimators}.
\bjtitle{Frontiers in Artificial Intelligence}
\bvolume{5},
\bfpage{777963}
(\byear{2022})
\end{barticle}
\endbibitem

\bibitem[\protect\citeauthoryear{Oh and Schuler}{2023}]{oh2023does}
\begin{barticle}
\bauthor{\bsnm{Oh}, \binits{B.-D.}},
\bauthor{\bsnm{Schuler}, \binits{W.}}:
\batitle{Why does surprisal from larger transformer-based language models provide a poorer fit to human reading times?}
\bjtitle{Transactions of the Association for Computational Linguistics}
\bvolume{11},
\bfpage{336}--\blpage{350}
(\byear{2023})
\end{barticle}
\endbibitem

\bibitem[\protect\citeauthoryear{de~Varda and Marelli}{2023}]{de2023scaling}
\begin{bchapter}
\bauthor{\bsnm{Varda}, \binits{A.}},
\bauthor{\bsnm{Marelli}, \binits{M.}}:
\bctitle{Scaling in cognitive modelling: A multilingual approach to human reading times}.
In: \bbtitle{Proceedings of the 61st Annual Meeting of the Association for Computational Linguistics (ACL)},
\bconflocation{Toronto, Canada},
pp. \bfpage{139}--\blpage{149}
(\byear{2023})
\end{bchapter}
\endbibitem

\end{thebibliography}

\end{document}